\documentclass{article}
\usepackage{jfrExamplee}
\usepackage{graphicx}
\usepackage{apalike}
\usepackage{setspace}
\usepackage[ruled,linesnumbered]{algorithm2e}
\usepackage{amsmath}
\usepackage{subfigure}
\usepackage{arydshln}
\usepackage{float}
 \usepackage{mathtools}
 \usepackage{nccmath}
\newcommand{\norm}[1]{\left\lVert#1\right\rVert}
\usepackage[section]{placeins}
\usepackage{xcolor}
\usepackage{url}
\usepackage[section]{placeins}

\title{Visual Model-predictive Localization for Computationally Efficient Autonomous Racing of a 72-gram Drone}

\author{
Shuo Li\thanks{ ∗Corresponding author} \\
Micro Air Vehicle Lab\\
Delft University of Technology\\
Delft,The Netherlands, 2629HS 1 \\
\texttt{s.li-4@tudelft.nl} \\
\And
Erik van der Horst \\
Micro Air Vehicle Lab \\
Delft University of Technology\\
Delft,The Netherlands, 2629HS 1 \\
\texttt{e.vanderhorst@tudelft.nl} \\
\And
Philipp Duernay \\
\\
Delft University of Technology\\
Delft,The Netherlands, 2629HS 1 \\
\texttt{p.durnay@student.tudelft.nl} \\
\AND
Christophe De Wagter \\
Micro Air Vehicle Lab\\
Delft University of Technology\\
Delft,The Netherlands, 2629HS 1 \\
\texttt{C.deWagter@tudelft.nl}
\And
Guido C.H.E. de Croon \\
Micro Air Vehicle Lab\\
Delft University of Technology\\
Delft,The Netherlands, 2629HS 1 \\
\texttt{g.c.h.e.decroon@tudelft.nl}
}

\begin{document}

\maketitle

\begin{abstract}
Drone racing is becoming a popular e-sport all over the world, and beating the best human drone race pilots has quickly become a new major challenge for artificial intelligence and robotics.
In this paper, we propose a strategy for autonomous drone racing which is computationally more efficient than navigation methods like visual inertial odometry and simultaneous localization and mapping. This fast light-weight vision-based navigation algorithm estimates the position of the drone by fusing race gate detections with model dynamics predictions. Theoretical analysis and simulation results show the clear advantage compared to Kalman filtering when dealing with the relatively low frequency visual updates and occasional large outliers that occur in fast drone racing. Flight tests are performed on a tiny racing quadrotor named ``Trashcan'', which was equipped with a Jevois smart-camera for a total of $72g$. The test track consists of $3$ laps around a 4-gate racing track. The gates are spaced $4$ meters apart and can be displaced from their supposed position. An average speed of $2m/s$ is achieved while the maximum speed is $2.6m/s$. \footnote{The video of the experiment is available at: \\ \url{https://www.youtube.com/playlist?list=PL_KSX9GOn2P8H0QvUZtLZSYggzQ2DFHCi}} To the best of our knowledge, this flying platform is currently the smallest autonomous racing drone in the world and is $6$ times lighter than the existing lightest autonomous racing drone setup ($420g$), while still being one of the fastest autonomous racing drones.  
\end{abstract}

\section{Introduction}
\begin{figure} [hbt!]
    \centering
    \subfigure[IROS 2016 drone race track]{
\includegraphics[width=.3\textwidth]{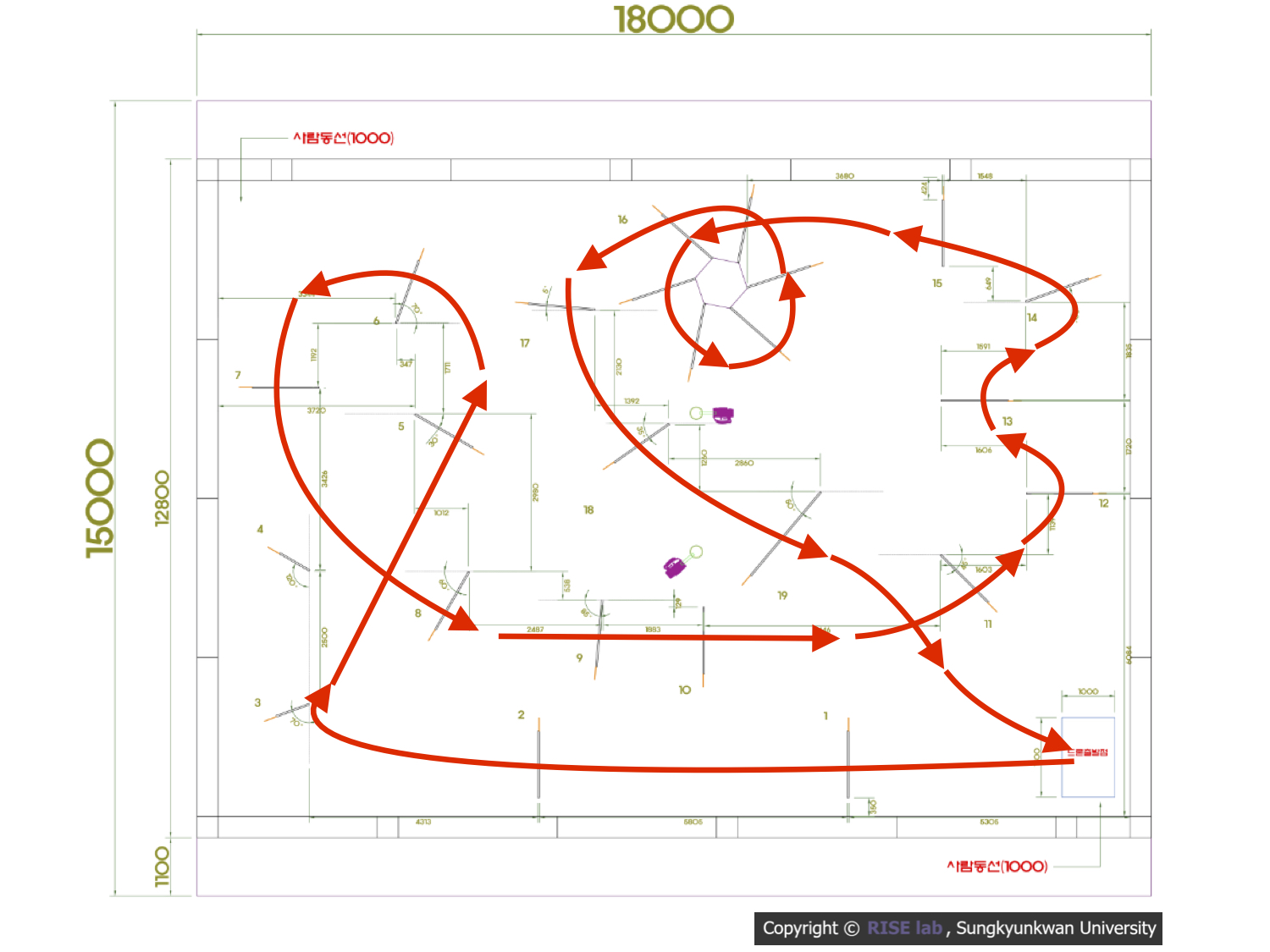}
}
\subfigure[IROS 2017 drone race track]{
\includegraphics[width=.3\textwidth]{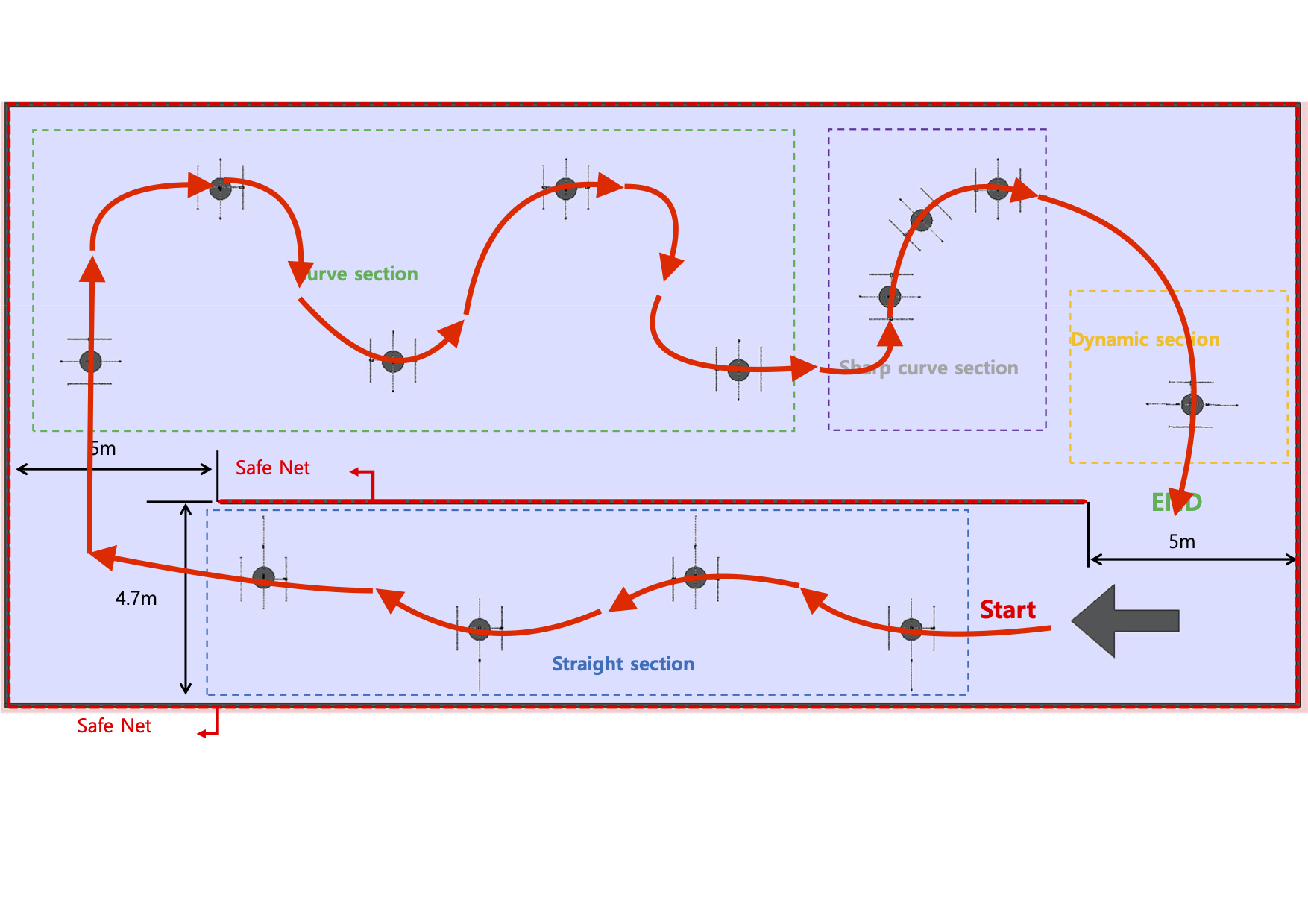}
}
\subfigure[IROS 2018 drone race track]{
\includegraphics[width=.3\textwidth]{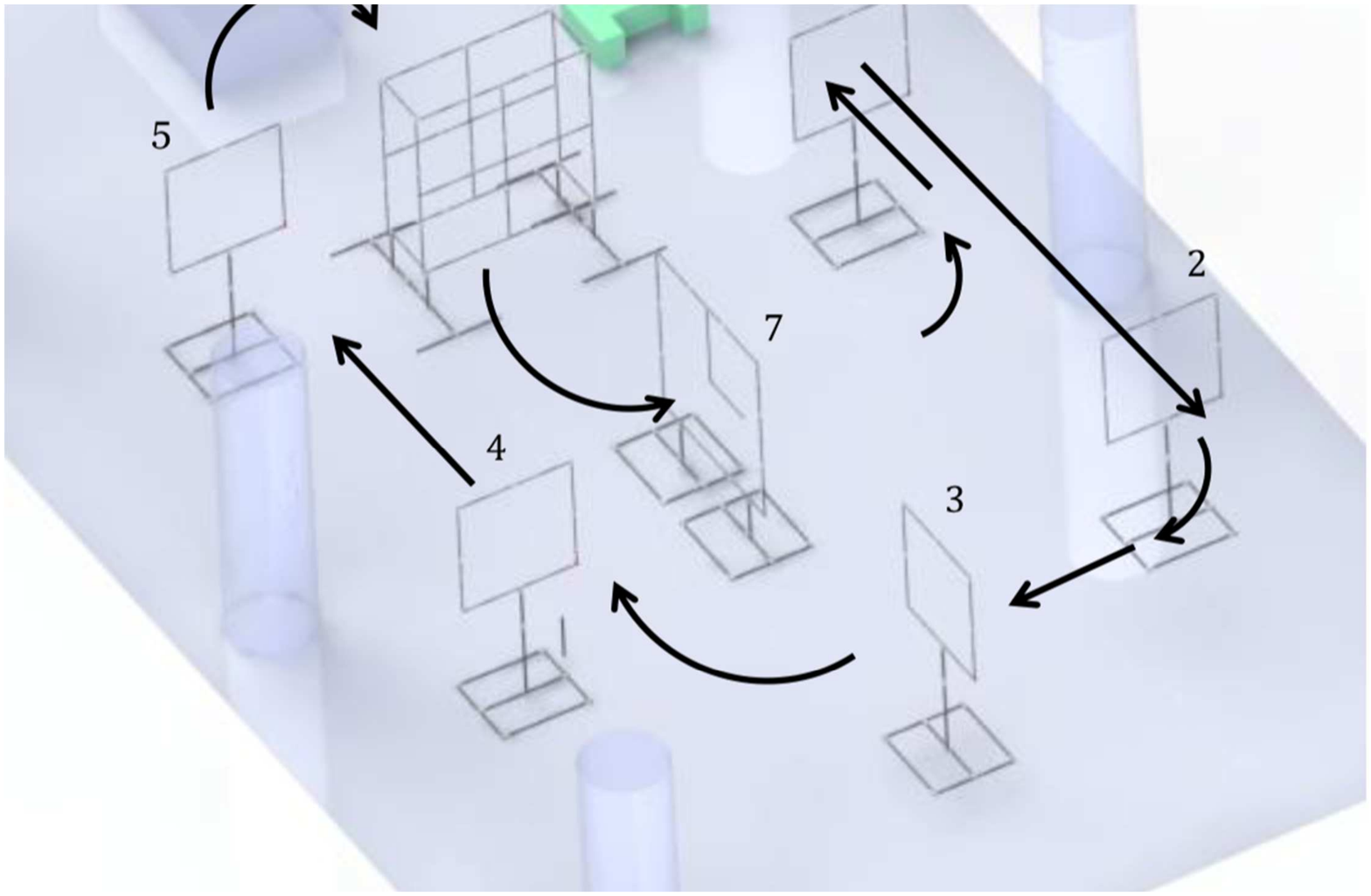}
}
    \caption{The IROS autonomous drone race track over the years 2016 - 2018 (a-c). The rules have always been the same. Flight is to be fully autonomous, so there can be no human intervention. The drone that passes through most subsequent gates in the track wins the race. When the number of passed gates is the same, or the track is fully completed, the fastest drone wins the race.}
    \label{fig:tracks}
\end{figure}

Drones, especially quadrotors, are transformed by enthusiasts in spectacular racing platforms. After years of development, drone racing has become a major e-sports, where the racers fly their drones in a preset course at high speed. It was reported that an experienced first person view (FPV) racer can achieve speeds up to $190 km/h$ when sufficient space is available. The quadrotor itself uses an inertial measurement unit (IMU) to determine its attitude and rotation rates, allowing it to execute the human's steering commands. The human mostly looks at the images and provides the appropriate steering commands to fly through the track as fast as possible. The advance in research areas such as computer vision, artificial intelligence and control raises the question: would drones not be able to fly faster than human pilots if they flew completely by themselves? Until now, this is an open question. In 2016, the world's first autonomous drone race was held at IROS 2016 \cite{moon2017iros}, which became an annual event trying to answer this question (Figure \ref{fig:tracks}). 

We focus on developing computationally efficient algorithms and extremely light weight autonomous racing drones that have the same or even better performance than currently existing larger drones. We believe that these drones may be able to fly faster, as the gates will be relatively larger for them. Moreover, a cheap, light-weight solution to drone racing would allow many people to use autonomous drones for training their racing skills. When the autonomous racing drone becomes small enough, people may even practice with such drones in their own home.   

Autonomous drone racing is indebted to earlier work on agile flight. Initially, quadrotors made agile maneuvers with the help of external motion capture systems \cite{mellinger2011minimum,mellinger2012trajectory}. The most impressive feats involved passing at high speeds through gaps and circles. More recently, various researchers have focused on bringing the necessary state estimation for these maneuvers onboard. Loianno et al. plan an optimal trajectory through a narrow gap with difficult angles while using Visual Inertial Odometry (VIO) for navigation \cite{loianno2017estimation}. The average maximum speed of their drone can achieve $4.5m/s$. However, the position of the gap is known accurately a priori, so no gap detection module is included in their research. Falanga et al. have their research on flying a drone through a gap aggressively by detecting the gap with fully onboard resources \cite{falanga2017aggressive}. They fuse the pose estimation from the detected gap and onboard sensors to estimate the state. In their experiment, the platform with a forward-facing fish-eye camera can fly through the gap with $3 m/s$.  Sanket et al. develop a solution for a drone to fly through arbitrarily shaped gaps without building an explicit 3D model of a scene, using only a monocular camera \cite{sanket2018gapflyt}. 

Drone racing represents a larger, even more challenging problem than performing short agile flight maneuvers. The reasons for this are that: (1) all sensing and computing has to happen on board, (2) passing one gate is not enough. Drone races can contain complex trajectories through many gates, requiring good estimation and (optimal) control also on the longer term, and (3) depending on the race, gate positions can change, other obstacles than gates can be present, and the environment is much less controlled than an indoor motion tracking arena. 

One category of strategies for autonomous drone racing is to have an accurate map of the track, where the gates have to be in the same place. One of the participants of the IROS 2017 autonomous drone race, the Robotics and Perception Group, reached gate $8$ in $35s$. In their approach, waypoints were set using the pre-defined map and VIO was used for navigation. A depth sensor was used for aligning the track reference system with the odometry reference system. NASA's JPL lab report in their research results that their drone can finish their race track in a similar amount of time as a professional pilot. In their research, a visual-inertial localization and mapping system is used for navigation and an aggressive trajectory connecting waypoints is generated to finish the track \cite{morrell2018differential}. Gao et al. come up with a teach-and-repeat solution for drone racing \cite{gao2019optimal}.  In the teaching phase, the surrounding environment is reconstructed and a flight corridor is found. Then, the trajectory can be optimized within the corridor and be tracked during the repeating phase. In their research, VIO is employed for pose estimation and the speed can reach $3m/s$. However, this approach is sensitive to changing environments. When the position of the gate is changed, the drone has to learn the environment again. 

The other category of strategies for autonomous drone race employs coarser maps and is more oriented on gate detection. This category is more robust to displacements of gates. The winner of IROS 2016 autonomous drone race, Unmanned Systems Research Group, uses a stereo camera for detecting the gates \cite{jung2018direct}. When the gate is detected, a waypoint will be placed in the center of the gate and a velocity command is generated to steer the drone to be aligned with the gate. The winner of the IROS 2017 autonomous drone race, the INAOE team, uses metric monocular SLAM for navigation. In their approach, the relative waypoints are set and the detection of the gates is used to correct the drift of the drone \cite{moon2019challenges}. Li et al. combine gate detection with onboard IMU readings and a simplified drag model for navigation \cite{li2018autonomous}. With their approach, a Parrot Bebop 1 ($420g$) can use its native onboard camera and processor to fly through $15$ gates with $1.5m/s$ along a narrow track in a basement full of exhibits. Kaufmann et al. use a trained CNN to map the input images to the desired waypoint and the desired speed to approach it \cite{kaufmann2018deep}. With the generated waypoint, a trajectory through the gate can be determined and executed while VIO is used for navigation. The winner of the IROS 2018 autonomous drone race, the Robotics and Perception Group, finished the track with $2 m/s$ \cite{kaufmann2018beauty}. During the flight, the relative position of the gates and a corresponding uncertainty measure are predicted by a Convolutional Neural Network (CNN). With the estimated position of the gate, the waypoints are generated, and a model predictive controller (MPC) is used to control the drone to fly through the waypoints while VIO is used for navigation.

From the research mentioned above, it can be seen that many of the strategies for autonomous drone racing are based on generic, but computationally relatively expensive navigation methods such as VIO or SLAM. These methods require heavier and more expensive processors and sensors, which leads to heavier and more expensive drone platforms. Forgoing these methods could lead to a considerable gain in computational effort, but raises the challenge of still obtaining fast and robust flight. 

\begin{figure} [hbt!]
    \centering
    \includegraphics[scale = 0.6,trim={0cm 7cm 0cm 8cm},clip]{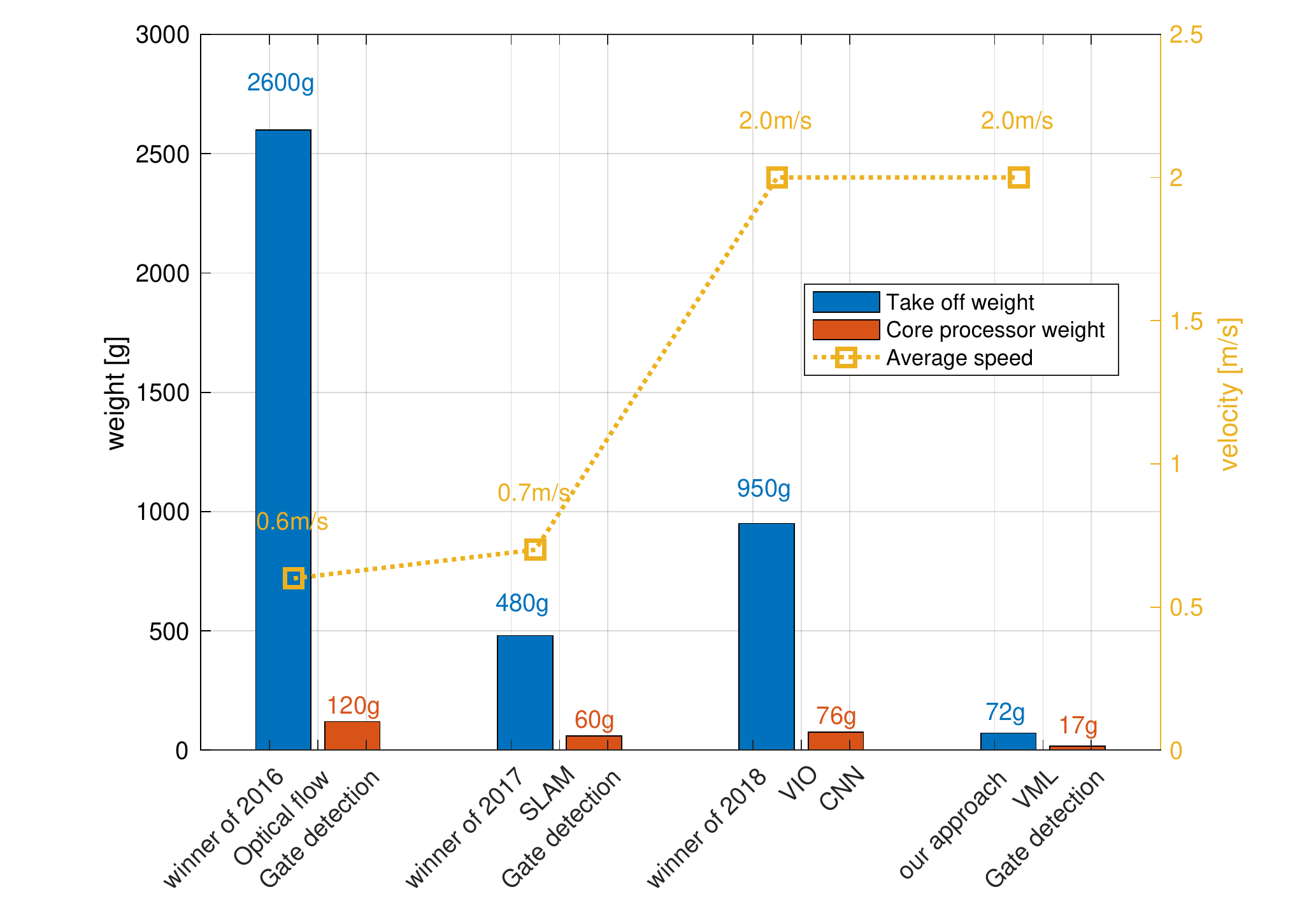}
    \caption{The weight and the speed of the approach proposed in this article and the winners' of IROS autonomous drone race. All weights are either directly from the articles or estimated from online specs of the used processors.}
    \label{fig:weight and speed comparision}
\end{figure}

In this paper, we present a solution to this challenge. In particular, we propose a Visual Model-predictive Localization (VML) approach to autonomous drone racing. The approach does not use generic vision methods such as VIO and SLAM and is still robust to gate changes, while reaching speeds competitive to the currently fastest autonomous racing drones. The main idea is to rely as much as possible on a predictive model of the drone dynamics, while correcting the model and localizing the drone visually based on the detected gates and their supposed positions in the global map. To demonstrate the efficiency of our approach, we implement the proposed algorithms on a cheap, commercially available smart-camera called ``Jevois'' and mount it on the ``Trashcan'' racing drone. The modified Trashcan weighs only $72g$ and is able to fly the race track with high speed (up to $2.6m/s$). The vision-based navigation and high-level controller run on the Jevois camera while the low-level controller provided by the open source Paparazzi autopilot \cite{gati2013open,hattenberger2014using} runs on the Trashcan. To the best of our knowledge, the presented drone is the smallest and one of the fastest autonomous racing drone in the world. Figure \ref{fig:weight and speed comparision} shows the weight and the speed of our drone in comparison to the drones of the winners of the IROS autonomous drone races. 

\section{Problem Formulation and System Description}
\subsection{Problem Formulation}
In this work, we will develop a hardware and a software system that the flying platform can fly through a drone race track fully autonomously with high speed using only onboard resources. The racing track setup can be changed and the system should be adaptive to this change autonomously. 

For visual navigation, instead of using SLAM or VIO, we directly use a computationally efficient vision algorithm for the detection of the racing gate to provide the position information. However, implementing such a vision algorithm on low-grade vision and processing hardware results in low frequency, noisy detections with occasional outliers. Thus, a filter should be employed to still provide high frequency and accurate state estimation. In Section~\ref{lab:MHE}, we first briefly introduce the 'Snake Gate Detection' method and a pose estimation method used to provide position measurements. Then, we propose and analyze the novel visual model-predictive localization technique that estimates the drone's states within a time window. It fuses the low-frequency onboard gate detections and high-frequency onboard sensor readings to estimate the position and the velocity of the drone. The control strategy to steer the drone through the racing track is discussed. The simulation result in Section~\ref{lab:simulation result} shows the comparison between the proposed filter and the Kalman filter in different scenarios with outliers and delay. In Section~\ref{sec:Experiment Result}, we will introduce the flying experiment of the drone flying through a racing track with gate displacement, different altitude and moving gate during the flight. In Section \ref{sec:Discussion}, the generalization and the limitation of the proposed method are discussed. Section~\ref{lab:conclusion} concludes the article.

\subsection{System Overview}
\begin{figure} [hbt!]
    \centering
\includegraphics[scale=0.06,trim={0cm 0cm 0cm 0cm},clip]{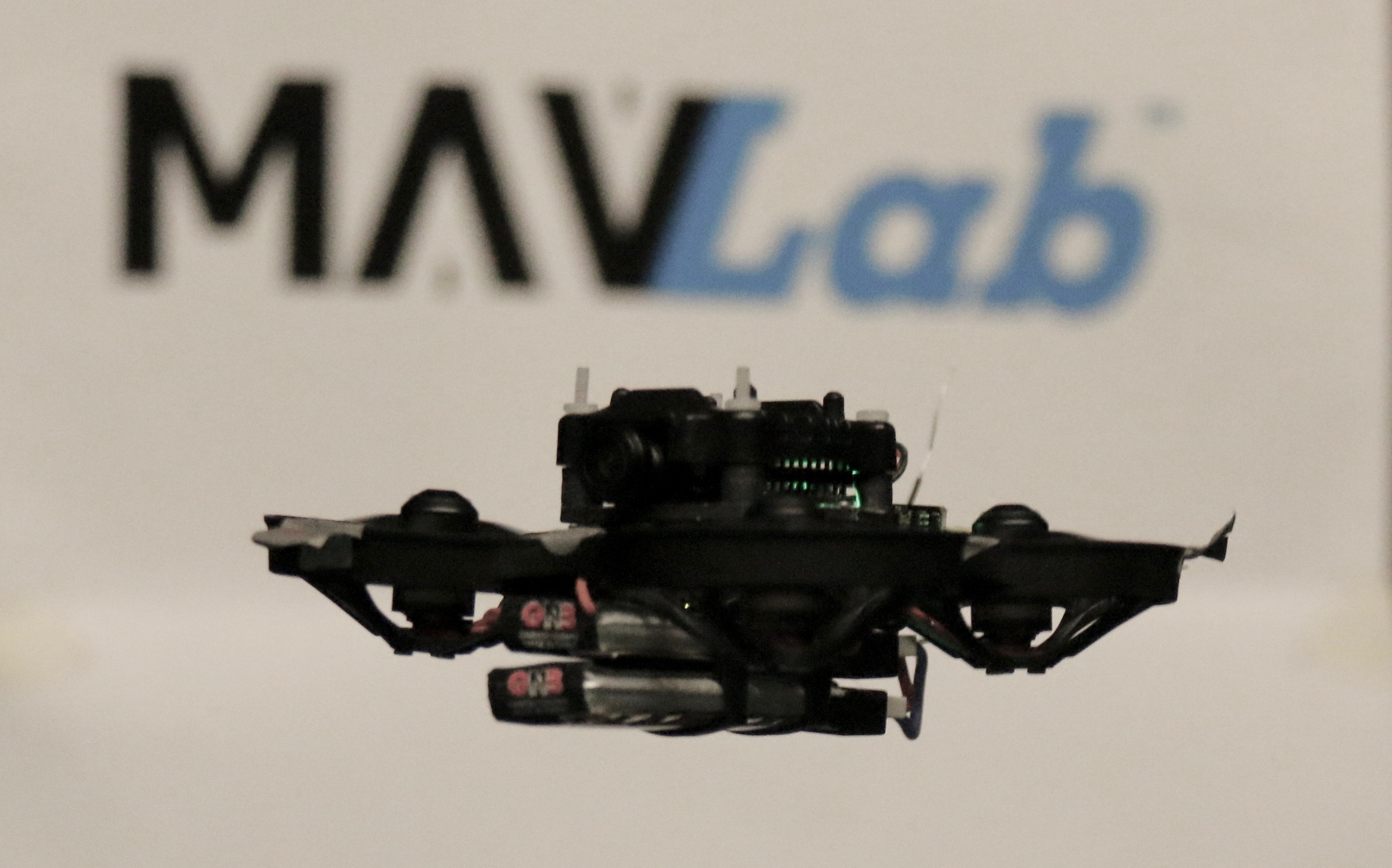}
    \caption{The flying platform. The Jevois is mounted on the Trashcan. The Trashcan provides power to the Jevois and they communicate with each other by the MAVLink protocol. The weight of the whole platform is only $72g$.}
    \label{fig:trashcan_jevois}
\end{figure}

To illustrate the efficiency of our approach, we use a small racing drone called Trashcan (Figure \ref{fig:trashcan_jevois}). This racing drone is designed for FPV racing with the Betaflight flight controller software. In our case, to fly this Trashcan autonomously, we replaced Betaflight by the Paparazzi open source autopilot for its flexibility of adding custom code, stable communication with the ground for testing code and active maintenance from the research community. In this article, the Paparazzi software only aims to provide a low level controller. The main loop frequency is $2k$Hz. We employ a basic complementary filter for attitude estimation and the attitude control loop is a cascade control including a rate loop and an attitude loop. For each loop, a P-controller is used. The details of Trashcan's hardware can be found in Table \ref{tab:specifications of Trashcan} 

\begin{table}[H]
\caption{The specifications of Trashcan's hardware}
\centering
\begin{tabular}{|c|c|}
\hline
\centering
Weight & $48g$ (with the original camera) \\ \hline
Size & $98mm\times98mm\times36mm$ \\ \hline
Motor & TC0803 KV15000    \\ \hline
MCU & STM32F4 ($100$MHZ)   \\ \hline
Receiver & FrSky D16  \\ \hline
\end{tabular}
\label{tab:specifications of Trashcan}
\end{table}

\begin{figure} [hbt!]
    \centering
    \subfigure[The two threads structure running on Jevois. For the gate detection thread, the frequency of gate detection ranges from $10$HZ to $30$HZ while the frequency of control and filtering thread is $512$HZ]{\includegraphics[scale=0.45,trim={0cm 0cm 0cm 0cm},clip]{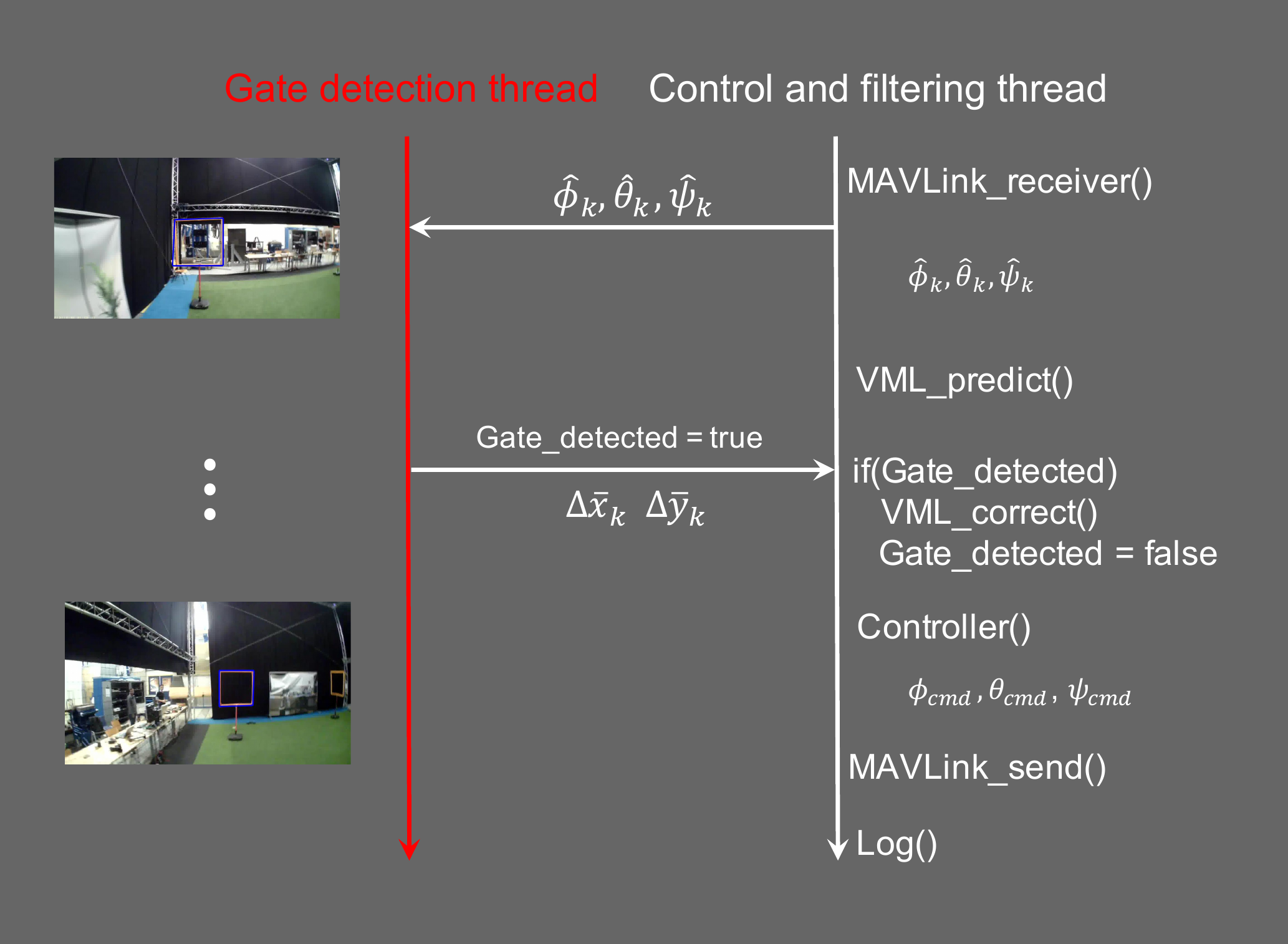}} \hspace{1cm} \\
     \subfigure[The software architecture of the UAV platform. The vision detection, filtering and control are all running on Jevois. Paparazzi provides the low level controller to stabilize the drone]{\includegraphics[scale=0.5,trim={0cm 0cm 0cm 0cm},clip]{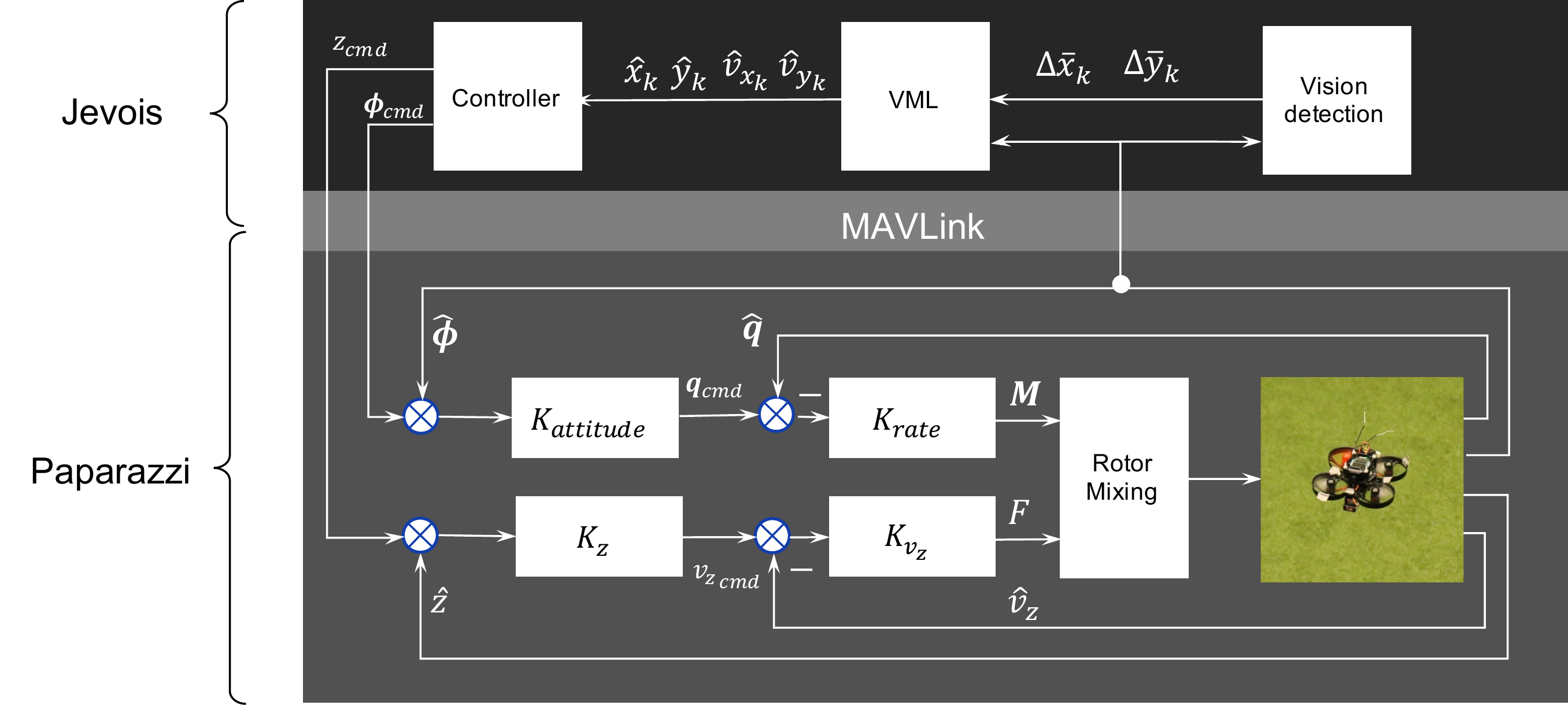}}
    \caption{The architectures of the software on Jevois and the software of the whole flying platform}
    \label{fig:two threads}
\end{figure}

For the high level vision, flight planning and control tasks, we use a light-weight smart camera ($17g$) called Jevois, which is equipped with a quad core ARM Cortex A7 processor and a dual core Mali-400 GPU. In our experiment, there are two threads running on the Jevois, one of which is for vision detection and the other one is for filtering and control (Figure \ref{fig:two threads}(a)). In our case, the frequency of detecting gates ranges from $10$HZ to $30$HZ and the frequency of filtering and control is set to $512$HZ.  The Gate detection thread processes the images in sequence. When it detects the gate it will send a signal telling the other thread a gate is detected. The control and filtering thread keeps predicting the states and calculating control command in high frequency. It uses a novel filtering method, explained in Section \ref{lab:MHE}, for estimating the state based on the IMU and the gate detections.

The communication between the Jevois and Trashcan is based on the MAVLink protocol with a baud rate of $115200$. Trashcan sends the AHRS estimation with a frequency of $512$HZ. And the Jevois sends the attitude and altitude commands to Trashcan with a frequency of $200$HZ. The software architecture of the flying platform can be found in Figure \ref{fig:two threads}(b).


In Figure \ref{fig:two threads}(b), the Gate detection and Pose estimation module first detects the gate and estimates the relative position between the drone and the gate. Next, the relative position will be sent to the Gate assignment module to be transferred to global position. With the global position measurements and the onboard AHRS reading, the proposed VML filter fuses them together to have accurate position and velocity estimation. Then, the Flight plan and high level controller will calculate the desired attitude commands to steer the drone through the whole track. These attitude commands will be sent to the drone via MAVLink protocol. On the Trashcan drone, Paparazzi provides the low level controller to stabilize the drone.  

\section{Robust Visual Model-predictive Localization (VML) and Control}
\label{lab:MHE}
State estimation is an essential part of drones' autonomous navigation. For outdoor flight, fusing a GPS signal with onboard inertial sensors is a common way to estimate the pose of the drone \cite{santana2015outdoor}. However, for indoor flight, a GPS signal is no longer available. Thus, off-board cameras \cite{lupashin2014platform}, Ultra Wide Band Range beacons \cite{mueller2015fusing} or onboard cameras \cite{mcguire2017efficient} can be used to provide the position or velocity measurements for the drone. The accuracy and time-delay of these types of infrastructure setups differ from each other. Hence, the different sensing setups have an effect on what type of filtering is best for each situation. The most commonly used state estimation technique in robotics is the Kalman filter and its variants, such as the Extended Kalman filter \cite{weiss2012versatile,santamaria2018autonomous,gross2012flight}. However, the racing scenario has properties that make it challenging for a Kalman filter. Position measurements from gate detections often are subject to outliers, have non-Gaussian noise, and can arrive at a low frequency. This makes the typical Kalman filter approach unsuitable because it is sensitive to outliers, is optimal only for Gaussian noise, and can converge slowly when few measurements arrive. In this section, we will propose a visual model-predictive localization technique which is robust to low-frequency measurements with significant numbers of outliers. Subsequently, we will also present the control strategy for the autonomous drone race.

\subsection{Gate assignment}
\begin{figure} [hbt!]
    \centering
    \subfigure[Snake gate detection. From one point on the gate $P_0$, the Snake gate detection method first searches up and down, then left and right to find all the four corners of the gate]{
\includegraphics[scale = 0.3,trim={0cm 0 0cm 0},clip]{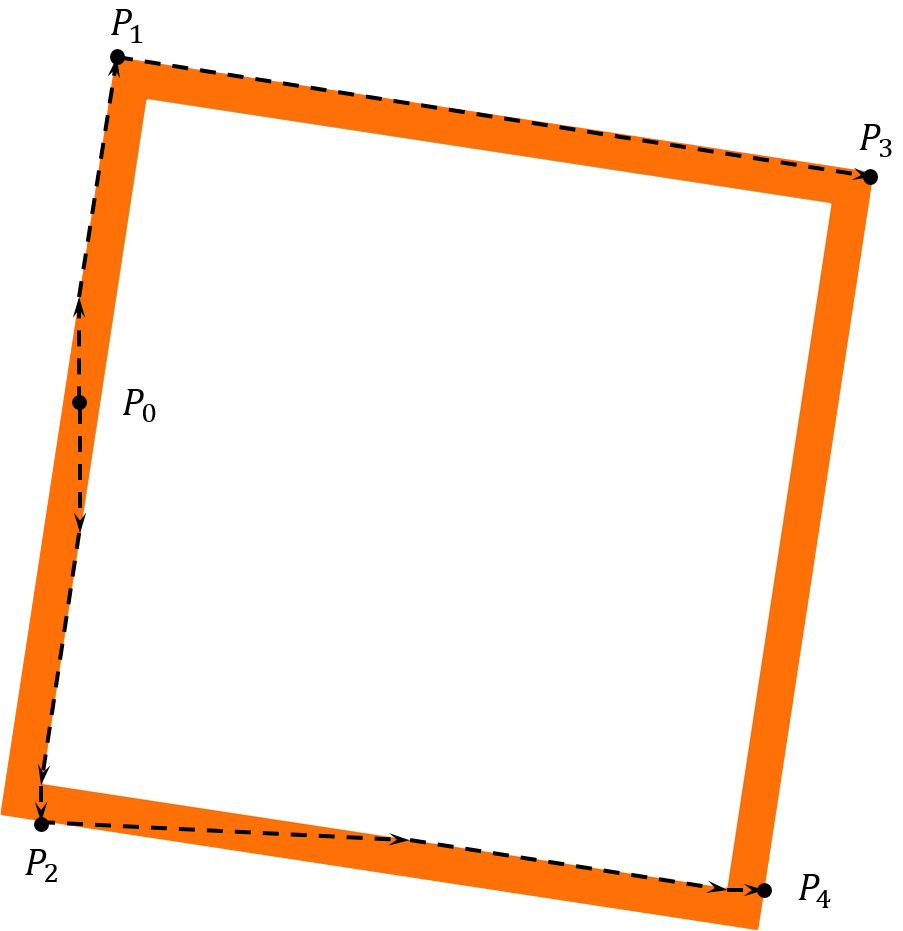}} \hspace{2cm}
\subfigure[When the four points of the gate are found, The relative position between the drone and the gate is calculated with the points' position, the camera's intrinsic parameters and the current attitude estimation]{
\includegraphics[width=.40\textwidth,trim={0cm 0 0cm 0},clip]{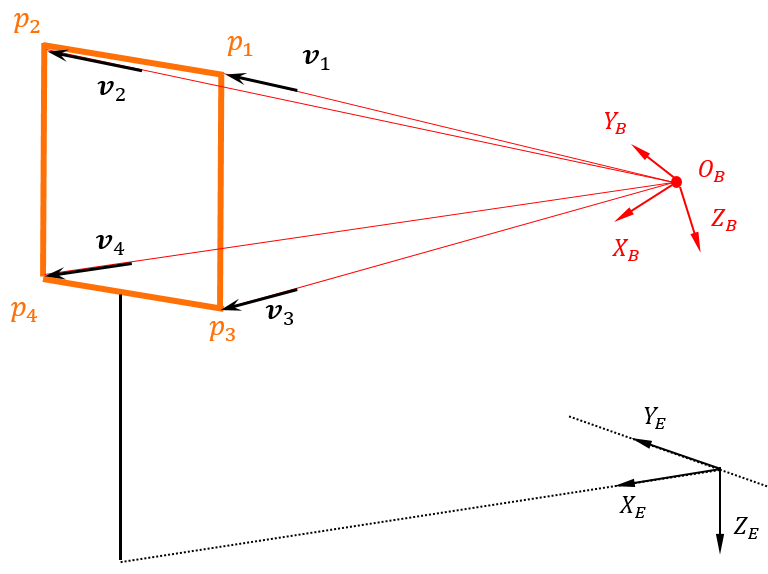}
}
    \caption{The Snake gate detection method and pose estimation method \cite{li2018autonomous}}
    \label{fig:gate detection}
\end{figure}

In this article, we use the ``snake gate detection'' and pose estimation technique as in Li et al. \cite{li2018autonomous}. The basic idea of snake gate detection is searching for continuing pixels with the target color to find the four corners of the gate. Subsequently, a perspective $n$-point (PnP) problem is solved, using the position of the four corners in the image plane, the camera's intrinsic parameters, and the attitude estimation to solve the relative position between the drone and the $i^{th}$ gate at time $k$, $\Delta \bar{\mathbf{x}}_k^i = [\Delta \bar{x}_k^i, \Delta \bar{y}_k^i]$. Figure \ref{fig:gate detection} shows this procedure, which is explained more in detail in \cite{li2018autonomous}. In most cases, when the light is even and the camera's auto exposure works properly, the gate in the image is continuous and the Snake gate detection algorithm can detect the gate correctly. However, after an aggressive turn, such as a turn to a window, the camera cannot adapt to the new light condition immediately. In this case, Snake gate detection usually cannot detect the gate. Another failure case is that due to the uneven light condition or the similar color in the background, Snake gate detection may get interfered with. These situations make the searching stop in the middle of the bar or stop at the background pixels. Although we have some mechanism to prevent these false positive detections, there is still a small chance that a false positive happens. The negative effect is that outliers may appear which leads to a challenge for the filter and the controller.   

\begin{figure} [hbt!]
    \centering
\includegraphics[scale=0.6,trim={0cm 0cm 0cm 0cm},clip]{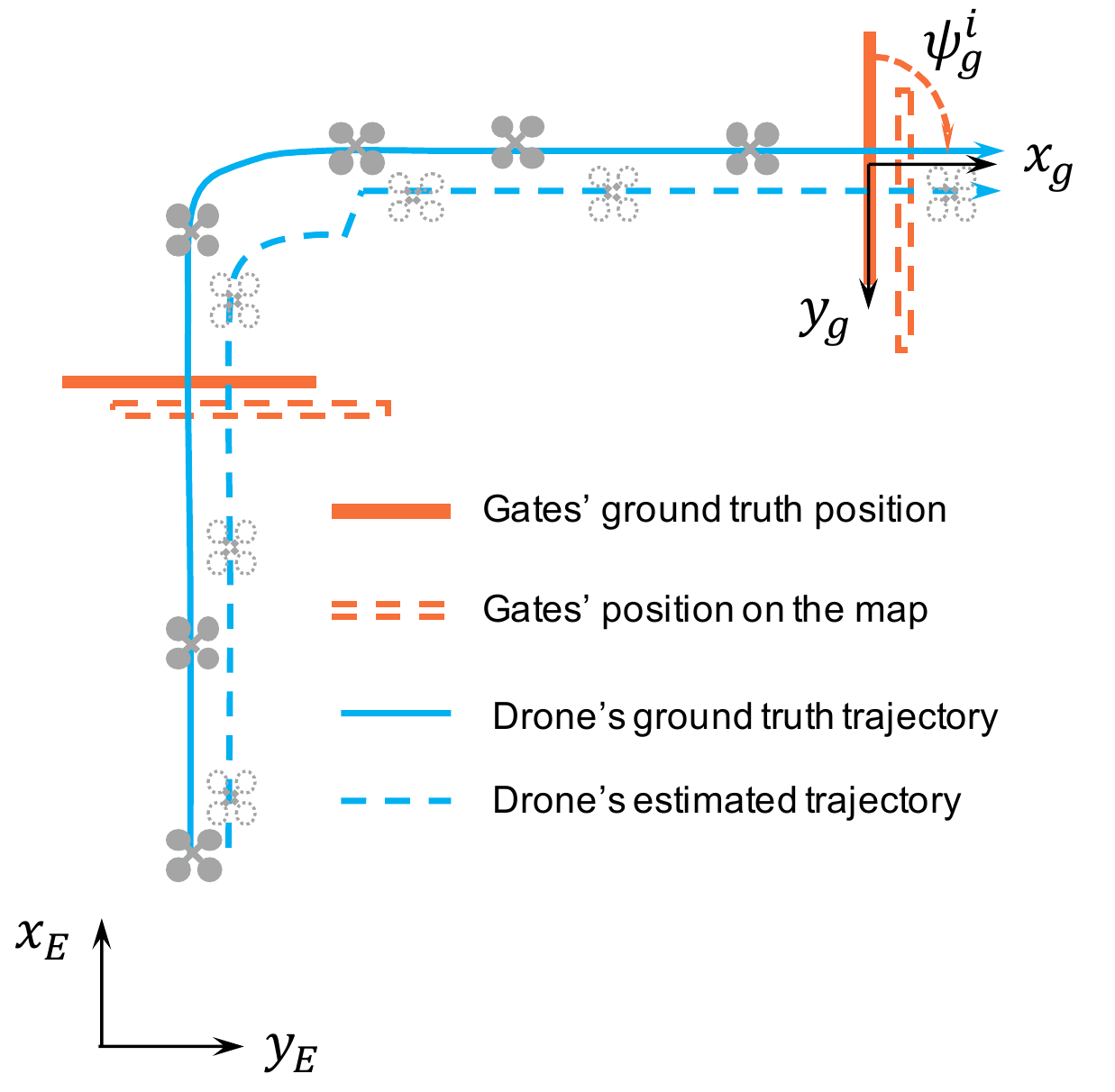}
    \caption{The gates are displaced. The drone uses the gate's position on the map to navigate. After passing through the first gate, it will use the second gate's position on the map for navigation. After seeing the second gate, the position of the drone will be corrected.}
    \label{fig:gate mismatch}
\end{figure}

Since for any race a coarse map of the gates is given a priori (cf. Figure \ref{fig:tracks}), the position and the heading of gate $i$, $\mathbf{x}_g^i=[x_g^i,y_g^i,\psi_g^i]$ can be known roughly (Figure \ref{fig:gate mismatch}). We use the gates' positions to transfer the relative position $\Delta \mathbf{\bar{x}}_k^i$ measured by camera to a global position $\mathbf{\bar{x}}_k = [\bar{x}_k, \bar{y}_k]$ by equation \ref{equ:local2global}. In equation \ref{equ:local2global}, $x_g^i$, $y_g^i$ and $\psi_g^i$ are the position of the gate $i$ which are known from the map. 


\begin{align}
    \begin{bmatrix}
    \bar{x}_k \\ \bar{y}_k
    \end{bmatrix} = \begin{bmatrix}
    x_g^i \\ y_g^i
    \end{bmatrix}+\begin{bmatrix}
    \cos{\psi_g^i} & \sin{\psi_g^i} \\ -\sin{\psi_g^i} & \cos{\psi_g^i}
    \end{bmatrix}
    \begin{bmatrix}
    \Delta \bar{x}_k^i \\ \Delta \bar{y}_k^i
    \end{bmatrix}
    \label{equ:local2global}
\end{align}

Here, we assume that the position of the gate is fixed. Any error experienced in the observations is then assumed to be due to estimation drift on the part of the drone. Namely, without generic VIO, it is difficult to make the difference between drone drift and gate displacements. If the displacements of the gates are moderate, this approach will work: after passing a displaced gate, the drone will see the next gate, and correct its position again. We only need a very rough map with the supposed global positions of the gates (Figure \ref{fig:gate mismatch}). Gate displacements only become problematic if after passing gate $i$ the gate $i+1$ would not be visible when following the path from the expected positions of gate $i$ to gate $i+1$.


At the IROS drone race, gates are identical, so for our position to be estimated well, we need to assign a detection to the right gate. For this, we rely on our current estimated global position $\hat{\mathbf{x}}_{k} = [\hat{x}_k,\hat{y}_k]$. When a gate is detected, we go through all the gates on the map using equation \ref{equ:local2global} to calculate the predicted position $\bar{\mathbf{x}}_k^i=[\bar{x}_k^i,\bar{y}_k^i]$. Then, we calculate the distance between the predicted drone's position ${\bar{\mathbf{x}}}_k^i$ and its estimated position ${\hat{\mathbf{x}}}_{k}$ at time $t_k$ by 
\begin{align}
\Delta d_k^i = \norm{\bar{\mathbf{x}}_k^i - \hat{\mathbf{x}}_{k}}_2
\end{align}

After going through all the gates, the gate with the predicted position closest to the estimated drone position is considered as the detected gate. At time $t_k$, the measurement position is determined by
\begin{align}
\begin{split}
&j = \operatorname*{argmin}_i\Delta d_k^i \\
&\bar{\mathbf{x}}_k = \bar{\mathbf{x}}_k^j
\end{split}
\end{align}


\begin{figure} [hbt!]
    \centering
    \subfigure[It iterates through all gates, evaluating where the drone would be if it was observing those gates. The position closest to the current global position is chosen as the right observation.]{
\includegraphics[width=.45\textwidth,trim={5cm 0cm 5cm 0},clip]{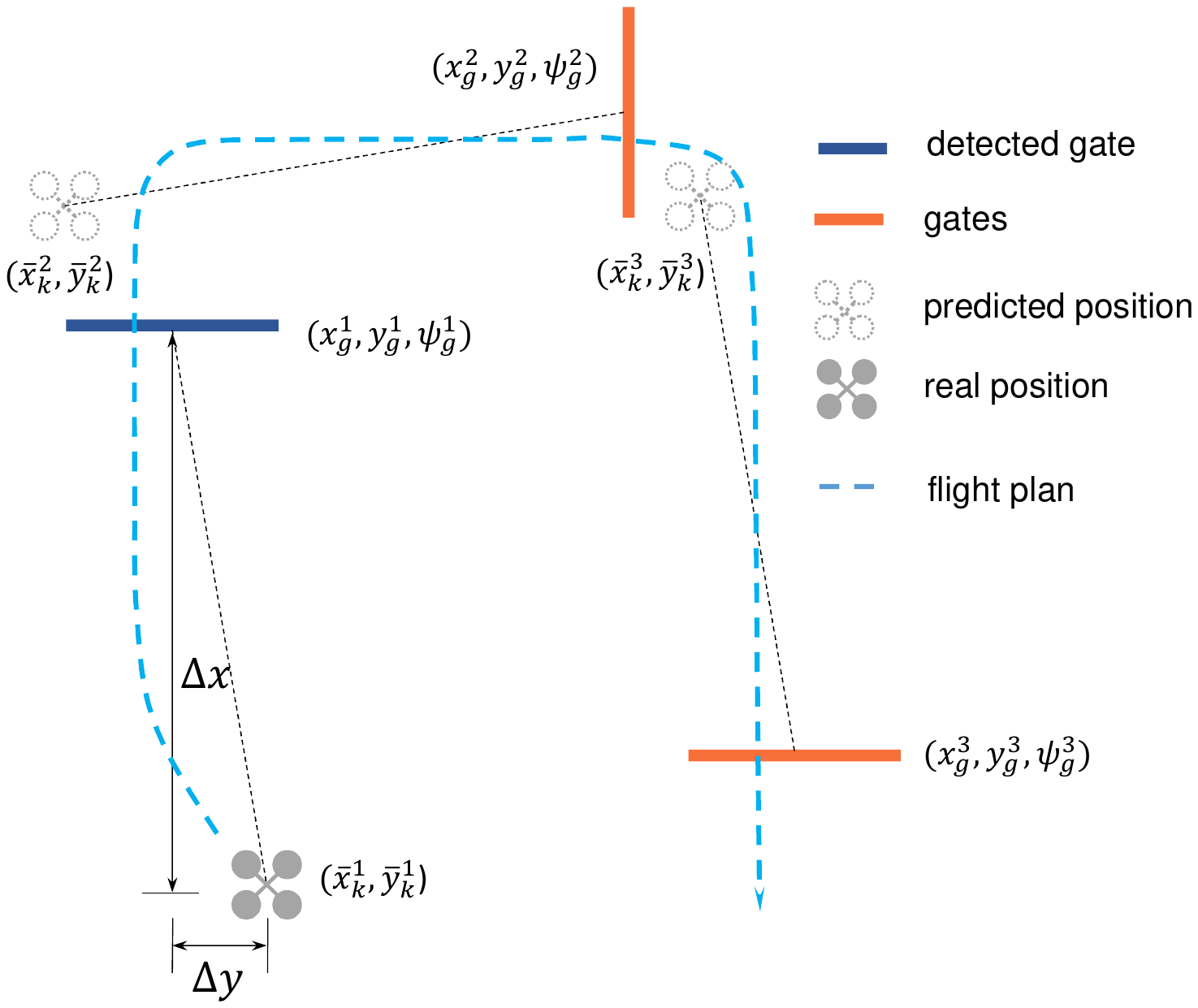}} \hspace{0.5cm}
\subfigure[The drone detects other gate instead of the one to be flew through. This still helps state estimation, as the observed gate indeed gives an estimate closest to the current estimated global position. ]{
\includegraphics[width=.45\textwidth,trim={5cm 0 5cm 0},clip]{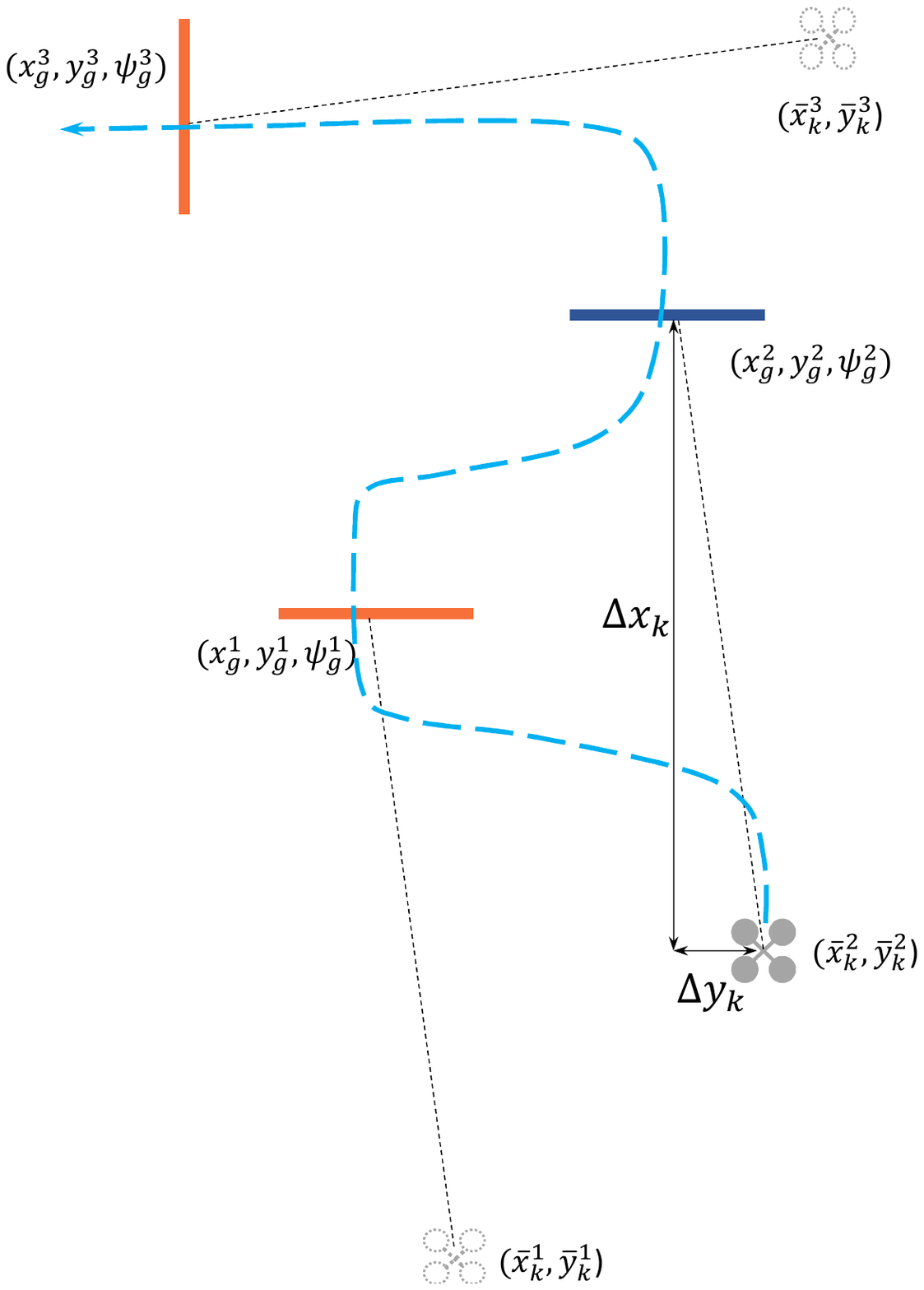}
}
    \caption{In most cases the drone will detect the next gate in the race track. However, the proposed gate assignment strategy also allows to exploit detections of other gates.}
    \label{fig:gate_assignment}
\end{figure}

The gate assignment technique can help us obtain as much information on the drone's position as possible when a gate is detected. Namely, it can also use detections of other gates than the next gate, and allows to use multiple gate detections at the same time in order to improve the estimation. Still, this procedure will always output a global coordinate for any detection. Hence, false positive or inaccurate detections can occur and have to be dealt with by the state estimation filter.

\subsection{Visual Model-predictive Localization (VML)}
The racing drone envisaged in this article has a forward-looking camera and an Inertial Measurement Unit (IMU). As explained in the previous section, the camera is used for localization in the environment, with the help of gate detections. Using a typical, cheap CMOS camera will result in relatively slow position updates from the gate detection, with occasional outliers. The IMU can provide high-frequency, and quite accurate attitude estimation by means of an Attitude and Heading Reference System (AHRS).
The accelerations can also be used in predicting the change in translational velocities of the drone. 
In traditional \emph{inertial} approaches, the accelerations would be integrated. However, for smaller drones the accelerometer readings become increasingly noisy, due to less possible damping of the autopilot. Integrating accelerometers is `acceleration stable', meaning that a bias in the accelerometers that is not accounted for can lead to unbounded velocity estimates. Another option is to use the accelerometers to measure the drag on the frame, which - assuming no wind - can be easily mapped to the drone's translational velocity (cf. \cite{li2018autonomous}). Such a setup is `velocity stable', meaning that an accelerometer offset of drag model error would lead to a proportional velocity offset, which is bounded. On really small vehicles like the one we will use in the experiments, the accelerometers are even too noisy for reliably measuring the drag. Hence, the proposed approach uses a prediction model that only relies on the attitude estimated by the AHRS which is an indirect way of using the accelerometer. It uses the attitude and a constant altitude assumption to predict the forward acceleration, and subsequently velocity of the drone. The model is corrected from time to time by means of the visual localization. Although the IMU is used for estimating attitude, it is not used as an inertial measurement for updating translational velocities. This leads to the name of the method; Visual Model-predictive Localization (VML), which will be explained in detail in this subsection.



\subsubsection{Prediction Error Model}
\begin{figure} [hbt!]
    \centering
    \includegraphics[scale=0.45]{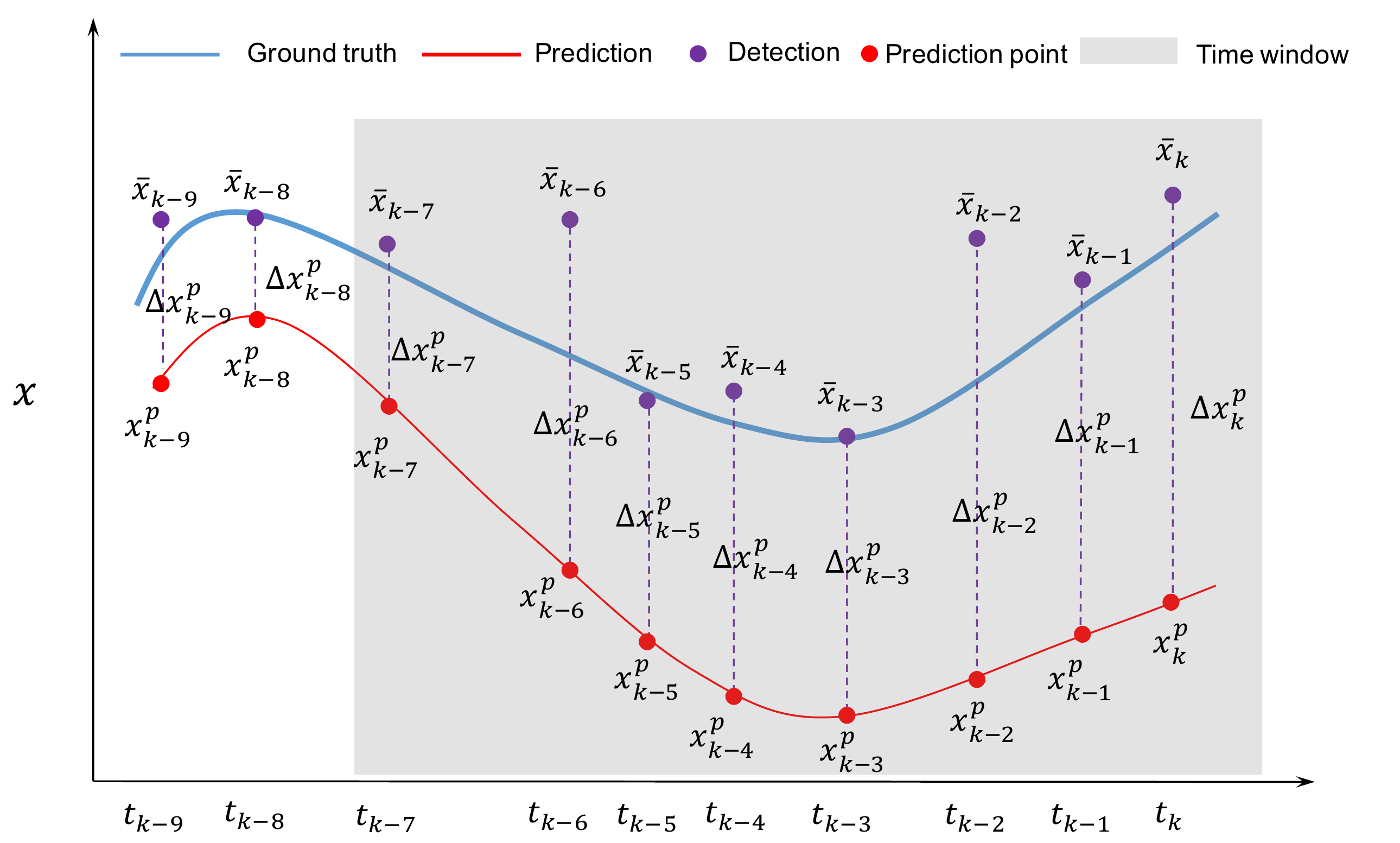}
    \caption{Illustrative sketch of the time window $t\in [t_{k-q},t_k]$. At the beginning of this time window, the difference between the ground truth and the prediction is $\Delta x_{k-q}$ and $\Delta v_{k-q}$. The prediction can be done with high frequency AHRS estimates. The vision algorithm outputs low frequency unbiased measurements. The prediction curve deviates more and more from the ground truth curve over time because of the AHRS bias and model inaccuracy.}
    \label{fig:MHE}
\end{figure}

As mentioned above, the attitude estimated from the AHRS is used in the prediction of the drone's velocity and position. However, due to the AHRS bias and the model inaccuracy, the prediction will diverge from the ground truth over time. Fortunately, we have visual gate detections to provide position information. This \emph{vision-based localization} will not integrate the error over time but it has a low frequency.  Figure \ref{fig:MHE} is a sketch of what the onboard predictions and the vision measurements look like. The red curve is the prediction result diverging from the ground truth curve because of AHRS biases. The magenta dots are the low frequency detections which distribute around the ground truth. The error between the prediction and measurements can be modeled as a linear function of time which will be explained later in this section. When the error model is estimated correctly, it can be used to compensate for the divergence of the prediction to obtain accurate state estimation.

\begin{figure} [hbt!]
    \centering
    \includegraphics[scale=0.6,trim={0cm 6cm 0cm 5cm},clip]{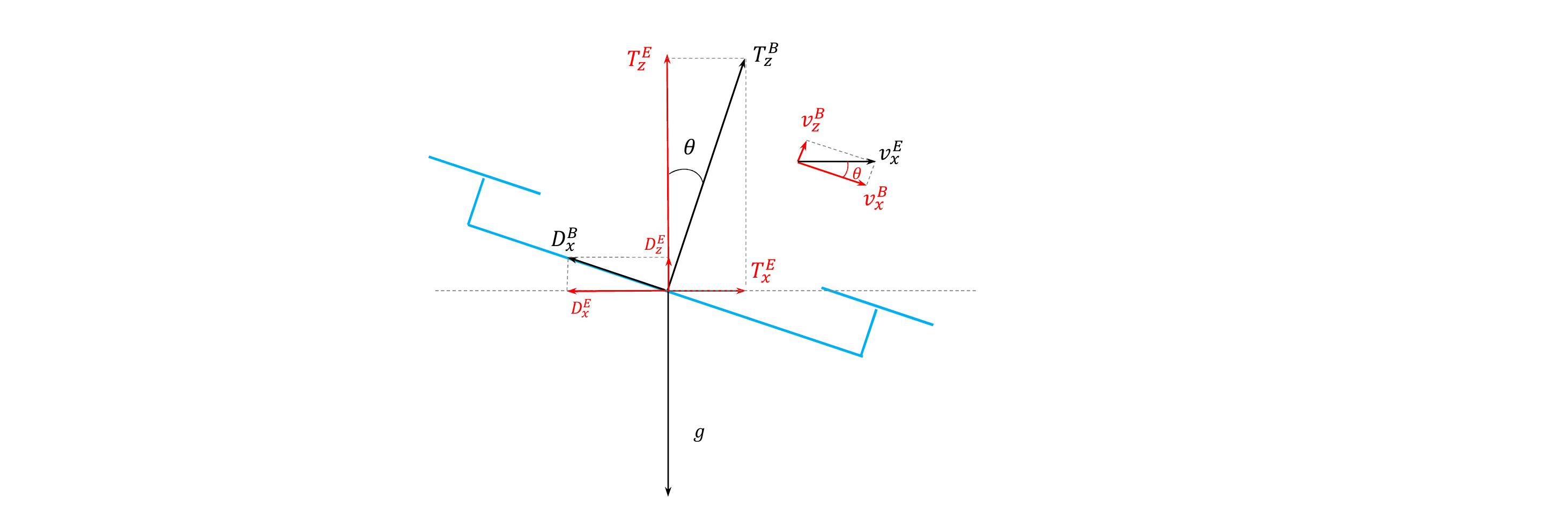}
    \caption{Free body diagram of the drone. $v_*^*(t)$ is the velocity of the drone. The superscript $E$ denotes north-east-down (NED) earth frame while $B$ denotes body frame. $T_*^*$ is the acceleration caused by thrust and $D_*^*$ is the acceleration caused by the drag, which is a linear function of the body velocity. $g$ is the gravity factor and $c$ is the drag factor which is positive. $\theta(t)$ is the pitch angle of the drone. It should be noted that since we use NED frame, $\theta < 0$ when the drone pitches down.}
    \label{fig:Free body diagram}
\end{figure}

Assuming that there is no wind, and knowing the attitude, we can predict the acceleration in the $x$ and $y$ axis. Figure \ref{fig:Free body diagram} shows the forces the drone experiences. $T_*^*$ denotes the acceleration caused by the thrust of the drone. It provides the forward acceleration together with the pitch angle $\theta$. $D_*^*$ denotes the acceleration caused by the drag which is simplified as a linear function of body velocity. \cite{faessler2017differential}

\begin{align}
\begin{cases}
    D^B_x &= c_xv^B_x \\ D^B_y &= c_yv^B_y
\end{cases}
\end{align}

where $c_*$ is the drag coefficient.

According to Newton's second law in $xoz$ plane,

\begin{align}
\begin{bmatrix}
a_x(t) \\ a_z(t)
\end{bmatrix}=
\begin{bmatrix}
0 \\ g
\end{bmatrix}+\Re_B^E(\theta)
\begin{bmatrix}
0 \\ T_z^B(t)
\end{bmatrix}+\Re_B^E(\theta)\mathbf{D}\Re_E^B(\theta)
\begin{bmatrix}
v^E_x(t) \\ v^E_z(t)
\end{bmatrix}
\label{equ:Newton second law}
\end{align}

Expand equation \ref{equ:Newton second law}, we have

\begin{align}
\begin{cases}
a_x(t) = \sin{\theta(t)}T^B_z(t)-v^E_x(t)c \\ a_z(t) = \cos{\theta(t)}T^B_z(t)+g-v^E_z(t)c
\end{cases}
\end{align}

where $\Re_E^B(\theta)$ is the rotation matrix and $\mathbf{D}=\begin{bmatrix}
-c & 0 \\ 0 & -c \end{bmatrix}$ is the drag coefficient matrix. If the altitude is kept the same as in the IROS drone race, we have

\begin{align}
\begin{cases}
T^B_z(t) = \frac{-g}{\cos\theta(t)} \\ a_x(t) = -g\tan\theta(t) - v^E_x(t)c
\end{cases}
\end{align}

Since the model in the $y$ axis has the same form as in the $x$ axis, the dynamic model of the quadrotor can be simplified as 

\begin{align}
\begin{cases}
\dot{x}(t) &= v^E_x(t) \\
\dot{y}(t) &= v^E_y(t) \\
\dot{v}^E_x(t) &= -g\tan{\theta}(t) - v^E_x(t)c \\
\dot{v}^E_y(t) &= g\tan{\phi}(t) - v^E_y(t)c 
\end{cases}
\label{equ:dynamics model}
\end{align}

where $x(t)$ and $y(t)$ are the position of the drone, and $\phi$ is the roll angle of the drone. In equation \ref{equ:dynamics model}, the movement in $x$ and $y$ axis is decoupled. Thus we only analyze the movement in the $x$ axis. The result can be directly generalized to the $y$ axis. The nominal model of the drone in $x$ axis can be written by 
 
\begin{align}
    \dot{\mathbf{x}}^n(t) = \mathbf{A}\mathbf{x}^n(t) +  \mathbf{B} u^n(t)
    \label{equ:nominal model}
\end{align}

where $\mathbf{x}^n(t) = \begin{bmatrix} x^n(t) \\ v_x^n(t) \end{bmatrix}$,$\mathbf{A} = \begin{bmatrix}
    0 & 1 \\ 0 & -c \end{bmatrix}$, $\mathbf{B} = \begin{bmatrix}0 \\ -g\end{bmatrix}$ and $u^n = \tan(\theta)$. The superscript $n$ denotes the nominal model. Similarly, with the assumption that the drag factor is accurate, the prediction model can be written as 

\begin{align}
    \dot{\mathbf{x}}^p(t) = \mathbf{A}\mathbf{x}^p(t) +  \mathbf{B} u^p(t)
    \label{equ:prediction model}
\end{align}

where $\mathbf{x}^p(t) = \begin{bmatrix} x^p(t) \\ v_x^p(t) \end{bmatrix}$ and $u^p = \tan(\theta + \theta_b)$. $\theta_b$ is the AHRS bias and is assumed to be a constant in short time. Consider a time window $t\in [t_{k-q},t_k]$, the states of nominal model at time $t_k$ are 
\begin{align}
    \mathbf{x}^n_k = (\mathbf{I}+\mathbf{A}T_s)^{q}\mathbf{x}^n_{k-q}+\sum_{i=1}^{q}(\mathbf{I}+\mathbf{A}T_s)^{i-1}\mathbf{B}T_su^n_{k-i}
\end{align}

where $T_s$ is the sampling time. The predicted states of model \ref{equ:prediction model} are
\begin{align}
    \mathbf{x}^p_k = (\mathbf{I}+\mathbf{A}T_s)^{q}\mathbf{x}^p_k+\sum_{i=1}^{q}(\mathbf{I}+\mathbf{A}T_s)^{i-1}\mathbf{B}T_su^p_{k-i}
\end{align}

Thus, the error between the predicted model and nominal model can be written as 

\begin{align}
    \Delta \mathbf{x}^p_k = (\mathbf{I}+\mathbf{A}T_s)^{q}\left[\mathbf{x}^p_{k-q}-\mathbf{x}^n_{k-q} \right]+
    \sum_{i=1}^{q}(\mathbf{I}+\mathbf{A}T_s)^{i-1} T_s\mathbf{B}u_b
    \label{equ:prediction error}
\end{align}

where $u_b = (u^p_{k-i}-u^n_{k-i})$ is the input bias which can be considered as a constant in a short time. In equation \ref{equ:prediction error},

\begin{align}
    (\mathbf{I}+\mathbf{A}T_s)^{i}=\begin{bmatrix}
    1 & T_s\sum\limits_{j=1}^{i}(1-cT_s)^{j-1} \\
    0 & (1-cT_s)^{i}
    \end{bmatrix}
\end{align}

Since the sampling time $T_s$ is small, ($T_s=0.002s$ in our case), we can assume

\begin{align}
    (\mathbf{I}+\mathbf{A}T_s)^{i} \approx \begin{bmatrix}
    1 & iT_s \\ 0 & 1
    \end{bmatrix}
\end{align}

Hence, equation \ref{equ:prediction error} can be approximated by

\begin{align}
     \Delta \mathbf{x}^p_k &=(\mathbf{I}+\mathbf{A}T_s)^{q}\left[\mathbf{x}^p_{k-q}-\mathbf{x}^n_{k-q} \right]+\sum_{i=1}^{q}\begin{bmatrix}
    1 & iT_s \\ 0 & 1
    \end{bmatrix}T_s\mathbf{B}u_b \\
    &= \begin{bmatrix}
    1 & qT_s \\ 0 & 1
    \end{bmatrix}\begin{bmatrix}
    \Delta x_k^p \\ \Delta v_k^p
    \end{bmatrix}+\begin{bmatrix}
    q & \frac{q(q+1)}{2}T_s \\ 0 & q
    \end{bmatrix}T_s\mathbf{B}u_b
     \label{equ:prediction error 2}
\end{align}

Expanding equation \ref{equ:prediction error 2}, we have 

\begin{align}
\begin{cases}
\Delta x^p_{k} = \Delta x^p_{k-q} + {qT_s}\Delta {v_x^p}_{k-q} - \frac{q(q+1)}{2}{T_s^2}g{u_b} \\
\Delta v^p_{k} = \Delta {v_x^p}_{k-q} - q{T_s}g{u_b}
\end{cases}
\label{equ:quad model}
\end{align}


Actually, $qT_s = t_k-t_{k-q}$ is the time span of the time window. If we neglect $T_s^2$ term, we can have the prediction error at time $t_k$ 

\begin{align}
    \Delta x^p_k = \Delta x^p_{k-q} + (t_k-t_{k-q})\Delta v^p_{k-q}
    \label{equ:linear_regression_model}
\end{align}

Thus, within a time window, the state estimation problem can be transformed to a linear regression problem
with model equation \ref{equ:linear_regression_model}, where $\mathbf{\hat{\beta}}=[\Delta x^p_{k-q},\Delta v^p_{k-q}]^{\rm T}$ are the parameters to be estimated. From equation \ref{equ:linear_regression_model}, we can see that in a short time window, the AHRS bias does not affect the prediction error. The error is mainly caused by the initial prediction error $\Delta x^p_{k-q}$. Furthermore, velocity error $\Delta v^p_{k-q}$ can cause the prediction error to diverge over time. If the time window is updated frequently, model \ref{equ:linear_regression_model} can remain accurate enough. Hence, in this work, we focus on the main contributors to the prediction error and will not estimate the bias term. The next step is how to efficiently and robustly estimate $\Delta x^p_{k-q}$ and $\Delta v^p_{k-q}$. 

In this simplified linear prediction error model, we use the constant altitude assumption to approximate the thrust $T^B_z$ on the drone, which may lead to inaccuracy of the model. During the flight, this assumption may be violated by aggressive maneuvers in $z$ axis. However, if the maneuver in $z$ axis is not very aggressive and the time window is small (in our case less than $2s$), the prediction error model's inaccuracy level can be kept in an acceptable range. In the simulation and the real-world experiment shown later, we will show that although the altitude of the drone changes $1m$ in $2s$, the proposed filter can still have very high accuracy with this assumption. Another way to improve the model accuracy is to estimate the thrust by fusing the accelerometer readings and rotor speed together, which needs the establishment of the rotors' model. It should also be noted that we neglect $T_s^2$ term in equation \ref{equ:quad model} to have a linear model. To increase the model accuracy, the prediction error model can be a quadratic model. In our case, since the time window is small, the linear model is accurate enough. 

\subsubsection{Parameter Estimation Method}
The classic way for solving the linear regression problem based on equation \ref{equ:linear_regression_model} is to use the Least Square Method (LS Method) with all data within the time window and estimate the parameters $\hat{\mathbf{\beta}}$.

\begin{align}
    \mathbf{\hat{\beta}} = (\mathbf{X}^{\rm T}\mathbf{X})^{-1}\mathbf{X}^{\rm T}\mathbf{Y}
    \label{equ:solution of LS}
\end{align}

where 
\begin{align*}
    \mathbf{\hat{\beta}} &= \begin{bmatrix} \Delta x^p_{k-q} & \Delta v^p_{k-q} \end{bmatrix}, \mathbf{X} = \begin{bmatrix}
    1 & t_{k-q}-t_{k-q} \\
    1 & t_{k-q+1} - t_{k-q} \\
    \vdots & \vdots \\
    1 & t_k -t_{k-q}
    \end{bmatrix}, \mathbf{Y}  = \begin{bmatrix}
    x^p_{k-q} - \bar{x}_{k-q} \\
    x^p_{k-q+1} - \bar{x}_{k-q+1} \\
    \vdots \\
    x^p_{k} - \bar{x}_{k} \\
    \end{bmatrix}
\end{align*}

\begin{figure} [hbt!]
    \centering
    \includegraphics[scale=0.6,trim={0cm 6cm 0cm 5cm},clip]{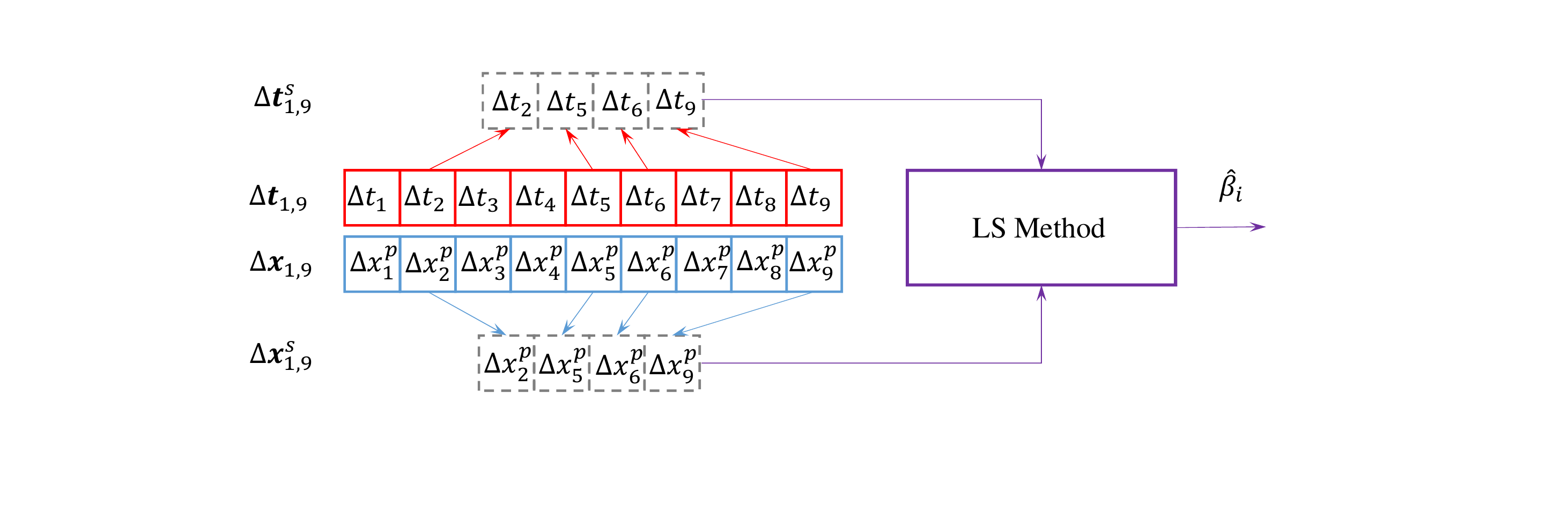}
    \caption{In the $i^{th}$ iteration, the data in the time window $t\in [t_1,t_9]$ will be randomly sampled into $\Delta \mathbf{t}^s_{k-q,k}$ and $\Delta \mathbf{x}^s_{k-q,k}$. Then LS Method (equation \ref{equ:solution of LS}) will be used to estimate the parameters $\hat{\mathbf{\beta}}_i$. In this example, $\sigma_s=0.4$, which means that $n^s=9\times0.4\approx4$ samples should be sampled. }
    \label{fig:ransac}
\end{figure}

The LS Method in equation \ref{equ:solution of LS} can give optimal unbiased estimation. However, if there exist outliers in the time window $t\in [t_{k-q},t_k]$, they will be considered equally during the estimation process. These outliers can significantly affect the estimation result. Thus, to exclude the outliers, we employ random sample consensus (RANSAC) to increase the performance \cite{fischler1981random}. In a time window $t\in [t_{k-q},t_k]$, we first calculate the prediction error $\Delta \mathbf{x}^p_{k-q,k}=\{\Delta x^p_{k-q+i}|\Delta x^p_{k-q+i}=\bar{x}_{k-q+i} - x^p_{k-q+i},0\leq i \leq q\}$ and time difference $\Delta \mathbf{t} = \{\Delta t_i| \Delta t_i = t_i-t_{k-q},0\leq i \leq q\}$. For each iteration $i$, the subsets of $\Delta \mathbf{x}^p_{k-q,k}$ and  $\Delta \mathbf{t}_{k-q,k}$ are randomly selected, which are denoted by $\Delta \mathbf{x}^s_{k-q,k}$ and $\Delta \mathbf{t}^s_{k-q,k}$. The size of the subset $n^s$ can be calculated by $n^s=q\sigma_s$, where $\sigma_s$ is the ratio of sampling. We use subsets $\Delta \mathbf{x}^s_{k-q,k}$ and $\Delta \mathbf{t}^s_{k-q,k}$ to estimate the parameters $\hat{\mathbf{\beta}}_i$ (Figure~\ref{fig:ransac}).

When $\hat{\mathbf{\beta}}_i$ is estimated, it will be used to calculate the total prediction error $\varepsilon_i$ of the all the data in the time window $t_i\in [t_{k-q},t_k]$ by 

\begin{align}
     \varepsilon_i = \sum^k_{j=k-q}\epsilon_j
    \label{equ:residual}
\end{align}

where 

\begin{align}
    \epsilon_j=\left\{
  \begin{array}{@{}ll@{}}
  \norm{\Delta {v^p_{k-q}}_i(\Delta t_j - \Delta t_{k-q})+\Delta {x^p_{k-q}}_i - \Delta x^p_j}_2, & \text{if}\ \epsilon_j <  \sigma_{th} \\
    \sigma_{th}, & \text{otherwise}
  \end{array}\right.
\end{align}


In the process of equation \ref{equ:residual}, if $\epsilon_j$ is larger than a threshold $\sigma_{th}$, it counts the threshold as the error. After all the iterations, the parameters $\hat{\mathbf{\beta}_i}$ which has the least prediction error will be selected to be the estimated parameters for this time window $t_i\in [t_{k-q},t_k]$. The pseudo-code of this Basic RANSAC Fitting (BRF) method can be found in Algorithm $2$.

With the Basic RANSAC Fitting (BRF) method, the influence of the outliers is reduced, but it has no mechanism to handle over-fitting. For example, in time window $t_i\in [t_{k-q},t_k]$, BRF can estimate the optimal parameters $\hat{\beta}$ with the minimal error. However, sometimes it will set $\Delta v^p_{k-q}$ to unrealistically high values. This happens when there are few detections in the time window, which may result in the inaccurate estimation of the parameters. In reality, the drone flies at maximum speed $3m/s$, so the velocity prediction error at the start of time window $t_{k-q}$ should not be too large. To avoid over-fitting, we add a penalty factor/prior matrix $\mathbf{P}$ to limit $\Delta v^p_{k-q}$ in the fitting process. The loss function can be written as 

\begin{align}
    J(\mathbf{\hat{\beta}})= \norm{\mathbf{X}\mathbf{\hat{\beta}}-\mathbf{Y}}_2^2 + \mathbf{\hat{\beta}}^{\rm T}\mathbf{P}\mathbf{\hat{\beta}}
\end{align}

where 
\begin{align}
\mathbf{P} =
\begin{bmatrix}
p_x & 0 \\ 0 & p_v
\end{bmatrix}
\end{align}

is the penalty factor/prior matrix. To minimize the loss function, we take derivatives of $J(\mathbf{\hat{\beta}})$ and let it be $0$

\begin{align}
    \frac{\partial J(\mathbf{\hat{\beta}})}{\partial \hat{\beta}}=2\mathbf{X}^{\rm T}\mathbf{X}\mathbf{\hat{\beta}}-2\mathbf{X}^{\rm T}\mathbf{Y}+\mathbf{P}\mathbf{\hat{\beta}} + \mathbf{P}^{\rm T}\mathbf{\hat{\beta}}=0
    \label{equ:prior}
\end{align}

Then we have the estimated parameters by

\begin{align}
    \mathbf{\hat{\beta}}=(\mathbf{X}^{\rm T}\mathbf{X}+\mathbf{P})^{-1}\mathbf{X}^{\rm T}\mathbf{Y}
\label{equ:solution of prior LS}
\end{align}

We call the use of equation \ref{equ:solution of prior LS} inside the RANSAC fitting the Prior RANSAC fitting (PRF). Compared to equation \ref{equ:solution of LS}, PRF has the penalty factor/prior matrix $\mathbf{P}$ in it. By tuning matrix $\mathbf{P}$ we can add the prior knowledge to the parameter distribution. For example, in our case $\Delta v^p_{k-q}$ should not be high. Thus, we can increase $p_v$ in $\mathbf{P}$ to limit the value of $\Delta v^p_{k-q}$.

To conclude, in this part we propose $3$ methods for estimating the parameters $\mathbf{\hat{\beta}}$. The first one is the LS Method which considers all the data in a time window equally. The second method is Basic RANSAC Fitting method (BRF), which has the mechanism to exclude the outliers. And the third one is Prior RANSAC Fitting method (PRF), which can not only exclude the outliers but also take into account the prior knowledge to avoid over-fitting. In the next section, we will discuss and compare these $3$ methods in simulation to see which one is the most suitable for our drone race scenario. 

\subsubsection{Prediction Compensation}
After the error model (equation \ref{equ:linear_regression_model}) is estimated in time window $k$, the error model can be used to compensate the prediction by
\begin{align}
\begin{bmatrix}
\hat{x}_{k+i} \\ \hat{v}_{k+i}
\end{bmatrix}
=\begin{bmatrix} x^p_{k+i} \\  v^p_{k+i} \end{bmatrix} + 
\begin{bmatrix} 1 & t_{k+i}-t_{k-q} \\  0 & 1 \end{bmatrix}\begin{bmatrix}\Delta x^p_{k-q}\\ \Delta v^p_{k-q}  \end{bmatrix}
\end{align}

Also, at each prediction step, the length $\Delta T = t_k-t_{k-q}$ of the time window will be checked, since the simplified model \ref{equ:linear_regression_model} is based on the assumption that the time span  of the time window $\Delta T$ is small. If $\Delta T$ is larger than the allowed maximum time window size $\Delta T_{max}$, the filter will delete the oldest elements until $\Delta T < \Delta T_{max}$. The pseudo-code of the proposed VML with LS Method can be found in Algorithm $3$ and Algorithm $4$.

\subsubsection{Comparison with Kalman Filter}
When it comes to state estimation or filtering technique, it is inevitable to mention the Kalman filter which is the most commonly used state estimation method. The basic idea of the Kalman filter is that at time $t_{k-1}$, it first predicts the states at time $t_{k}$ with its error covariance $\mathbf{P}_{k|k-1}$ to have prior knowledge of the states at $t_{k}$. 

\begin{align}
\begin{split}
&\hat{\mathbf{X}}_{k|k-1}=\hat{\mathbf{X}}_{k-1}+\mathbf{f}(\hat{\mathbf{X}}_{k-1},\mathbf{u}_{k-1}){\rm T_s} \\
&\mathbf{F}_{k-1}=\frac{\partial}{\partial \mathbf{x}}\mathbf{f}(\mathbf{x}(t),\mathbf{u}(t))|_{{\mathbf{x}(t)=\hat{\mathbf{x}}}_{k-1}} \\
&\Phi_{k|k-1}\approx\mathbf{I}+\mathbf{F}_{k-1}{\rm T} \\
&\mathbf{H}_k=\frac{\partial}{\partial \mathbf{x}}\mathbf{h}(\mathbf{x}(t))|_{{\mathbf{x}(t)=\hat{\mathbf{x}}}_{k}} \\
&\mathbf{P}_{k|k-1}=\mathbf{\Phi}_{k|k-1}\mathbf{P}_{k-1}\mathbf{\Phi}_{k|k-1}^{\rm T}+\mathbf{Q}_{k-1} \\
\end{split}
\end{align}

When an observation arrives, the Kalman filter uses an optimal gain $\mathbf{K}_k$ which is a combination of the prior error covariance $\mathbf{P}_{k+1|k}$ and the observation's covariance $\mathbf{R}_k$ to compensate the prediction, which as a result, leads to the minimum error covariance $\mathbf{P}_{k}$. 

\begin{align}
\begin{split}
&\delta\hat{\mathbf{X}}_k = \mathbf{K}_k\left \{ \mathbf{Z}_k-\mathbf{h}[\hat{\mathbf{X}}_{k|k-1},k]\right \} \\
&\mathbf{K}_k =\mathbf{P}_{k|k-1}\mathbf{H}_k^{\rm T}[\mathbf{H}_k\mathbf{P}_{k|k-1}\mathbf{H}_k^{\rm T}+\mathbf{R}_k]^{-1} \\
&\hat{\mathbf{X}}_k = \hat{\mathbf{X}}_{k|k-1}+\delta\hat{\mathbf{X}}_k \\
&\mathbf{P}_k=(\mathbf{I}-\mathbf{K}_k\mathbf{H}_k)\mathbf{P}_{k/k-1}(\mathbf{I}-\mathbf{K}_k\mathbf{H}_k)^{\rm T}+\mathbf{K}_k\mathbf{R}_k\mathbf{K}_k^{\rm T}
\end{split}
\end{align}

According to \cite{diderrich1985kalman}, a Kalman filter is a least square estimation made into a recursive process by combining prior data with coming measurement data. The most obvious difference between the Kalman filter and the proposed VML is that VML is not a recursive method. It does not estimate the states at $t_k$ only based on the last step states $\mathbf{\hat{x}}_{k-1}$. It estimates the states considering the previous prediction and observations in a time window. 

In the VML approach, we use least square method within a time window, which looks similar to the least square estimation method. However, there are two major differences between the two methods. The first one is that in the proposed VML, the prediction information is fused to the VML. Secondly and most importantly, we estimate the prediction error model $\hat{\beta}$ instead of estimating all the states in the time window as in the least square method. Thus, the VML has its advantages of handling outliers and delay by its time window mechanism and it also has the advantage of computational efficiency to the Least Square Estimation. In Section \ref{lab:simulation result}, we will introduce Kalman filter's different variants for outliers and delay and compare them with VML in estimation accuracy and computation load in detail.

\subsection{Flight Plan and High Level Control}
\begin{figure} [hbt!]
    \centering
  \includegraphics[scale=0.5,trim={0cm 0cm 0cm 0cm},clip]{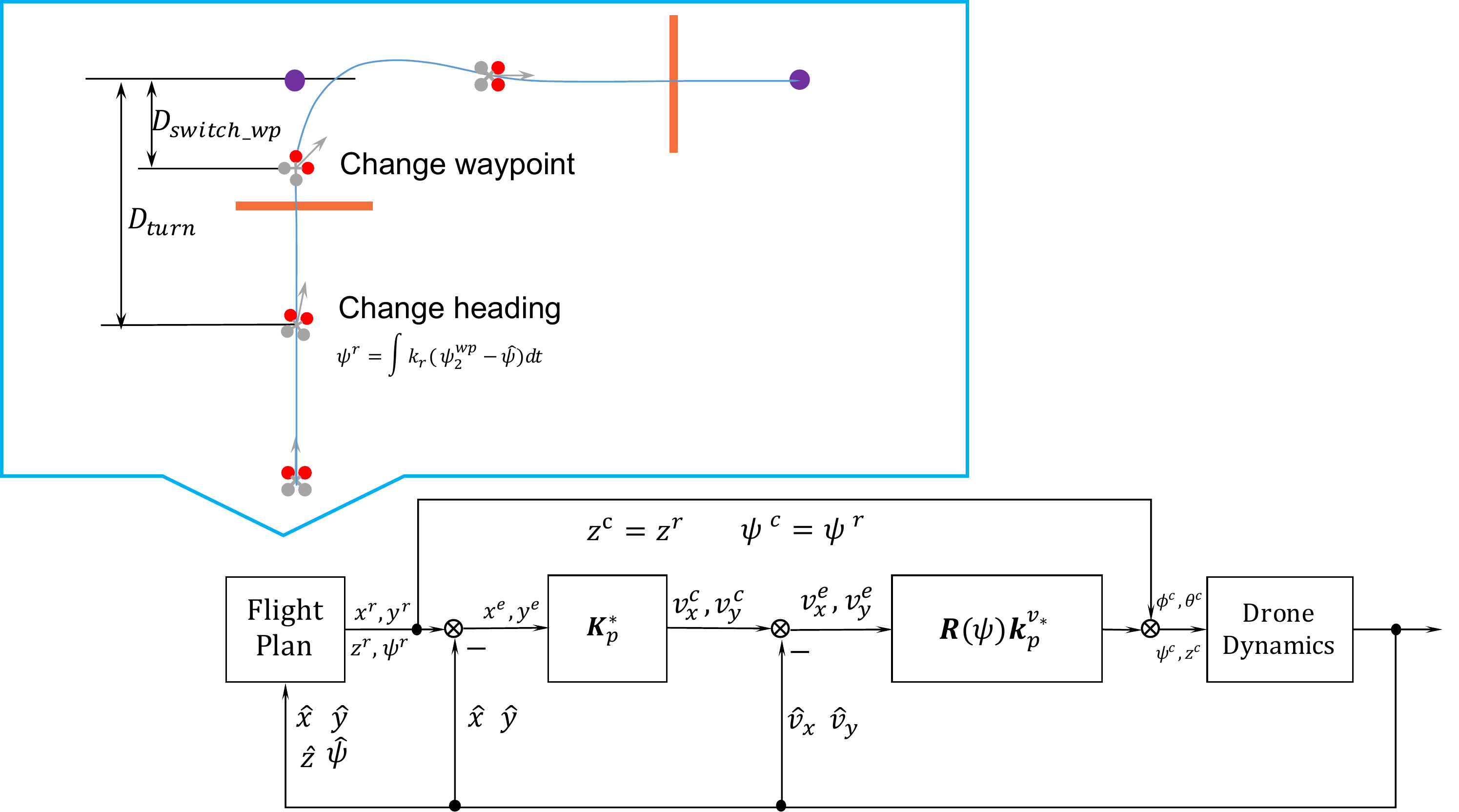}

    \caption{The Flight plan module generates the waypoints for the drone to fly the track. When the distance between the drone and the current waypoint $d < D_{turn}$, the drone starts to turn to the next waypoint while still approaching the current waypoint. When $d < D_{switch\_wp}$, the drone switches the current waypoint to the next one. The cascade P-controller is used for executing the reference trajectory from the flight plan module. The attitude and rate controllers are provided by the Paparazzi autopilot.  $k_r$ is a positive constant to adjust the speed of the drone's yawing to the setpoint. In the real world experiment and simulation, we set $k_r=1$.}
    \label{fig:controller}
\end{figure}

With the state estimation method explained above, to fly a racing track, we employ a flight plan module which sets the waypoints that guide the drone through the track and a two-loop cascade P-controller to execute the reference trajectory (Figure \ref{fig:controller}).

Usually, the waypoint is just behind the gate. When the distance between the drone and the waypoint is less than a threshold $D_{turn}$, the gate can no longer be detected by our method, and we set the heading of the drone to the next waypoint. This way, the drone will start turning towards the next gate before arriving at the waypoint. When the distance between the drone and the waypoint is within another threshold $D_{switch\_wp}$, the waypoint switches to the next point. With this strategy, the drone will not stop at one waypoint but already start accelerating to the next waypoint, which can help to save time. The work flow of flight plan module can be found in Algorithm $5$.

We employ a two-loop cascade P controller (equation \ref{equ:simulation controller}) to control the drone to reach the waypoints and follow the heading reference generated from the flight plan module. The altitude and attitude controllers are provided by the Paparazzi autopilot, and are both two-loop cascade controllers.


\begin{align}
   \mathbf{\Phi}^c(k)=\mathbf{R}_{\psi}\mathbf{K}_v(\mathbf{K}_x(\mathbf{x}^r(k)-\hat{\mathbf{x}}(k))-\hat{\mathbf{v}}(k))
    \label{equ:simulation controller}
\end{align}

where $\mathbf{\Phi}^c(k) = [\phi^c(k), \theta^c(k)]^{\rm T}$, $\mathbf{R}_{\psi}=\begin{bsmallmatrix} \cos(\psi) & -\sin(\psi) \\  \sin(\psi) & \cos(\psi) \end{bsmallmatrix}$, $\mathbf{K}_v=\begin{bsmallmatrix}{k_v}{}_x & 0 \\ 0 & {k_v}_y  \end{bsmallmatrix}$, $\mathbf{K}_x=\begin{bsmallmatrix}k_x & 0 \\ 0 & k_y  \end{bsmallmatrix}$, $\mathbf{x}^r(k) = [x^r(k), y^r(k)]^{\rm T}$, $\hat{\mathbf{x}}(k)=[\hat{x}(k),\hat{y}(k) ]^{\rm T}$, $\hat{\mathbf{v}}(k)=[\hat{v}_x(k),\hat{v}_y(k) ]^{\rm T}$.


\section{Simulation Experiments}
\label{lab:simulation result}
\subsection{Simulation Setup}
To verify the performance of VML in the drone race scenario, we first test it in simulation and then use an Extended Kalman filter as benchmark to compare both filters to see which one is more suitable in different operation points. We first introduce the drone's dynamics model used in the simulation.

\begin{align}
    \begin{split}
        \begin{bmatrix}
        \dot{x}\\\dot{y} \\\dot{z}
        \end{bmatrix} =&
        \begin{bmatrix}
        v_x \\ v_y \\v_z
        \end{bmatrix}\\
        \begin{bmatrix}
        \dot{v}_x  \\ \dot{v}_y \\ \dot{v}_z
        \end{bmatrix} =&
        \begin{bmatrix}
         0 \\ 0 \\ g
        \end{bmatrix}+
        \Re_B^E\begin{bmatrix}
        0 \\ 0 \\ T
        \end{bmatrix}+\Re_B^E\mathbf{K}\Re_E^B\begin{bmatrix} v_x \\ v_y \\ v_z
        \end{bmatrix} \\
        \begin{bmatrix}
        \dot{\phi} \\ \dot{\theta} \\ \dot{\psi} \\\dot{T} 
        \end{bmatrix} =&
        \begin{bmatrix}
        k_{\phi}(\phi^{c}-\phi) \\ k_{\theta}(\theta^{c}-\theta) \\ k_{\psi}(\psi^{c}-\psi) \\ k_{T}(T^{c}-T)
        \end{bmatrix}
    \end{split} 
    \label{equ:drone simulation model}
\end{align}

where $(x, y, z)$ is the position of the drone in the Earth frame. $v_*$ is the velocity of the drone. $g$ is the gravity factor. $T$ is the acceleration caused by the thrust force. $\phi$, $\theta$, $\psi$ are the three Euler angles of the body frame. And $\Re_B^E$ is the rotation matrix from the Body frame to the Earth frame. $\mathbf{K}=diag([-0.5, -0.5, 0])$ is the simplified first order drag matrix, where the values are based on a linear fit of the drag based on real-world data with the Trashcan drone. $\Re_B^E\mathbf{K}\Re_E^B[ v_x  v_y  v_z]^{\rm T}$ is the acceleration caused by other aerodynamics. The last four equations are the simplified first order model of the attitude controller and thrust controllers where the proportional feedback factors are $k_{\phi} = 6$, $k_{\theta} = 6$,$k_{\psi} = 5$,$k_{T} = 3$. Thus, the model \ref{equ:drone simulation model} in the simulation is a $10$ states $\mathbf{x} = [x, y, z, v_x, v_x, v_x, \phi, \theta, \psi, T]^{\rm T}$ and $4$ inputs $\mathbf{u}=[\phi^{c}, \theta^{c}, \psi^{c}, T^{c}]^{\rm T}$ nonlinear system. In this simulation, we use the same flight plan module and high-level controllers discussed in Section \ref{lab:MHE} (Figure \ref{fig:controller}) to generate a ground truth trajectory through a 4-gate square racing track. In this track, we use different height to test if the altitude change affects the accuracy of the VML.

\begin{table}[H]
\caption{The map of the simulated racing track}
\centering
\begin{tabular}{|c|c|c|c|c|}
\hline
\centering
Gate ID & $x[m]$ & $y[m]$ & $z[m]$ & $\psi[^{\circ}]$   \\
 \hline\hline
$1$ & $4$ & $0$  & $-1.5$ & $0$    \\ \hline
$2$ & $4$ & $4$ & $-2.5$ & $90$  \\ \hline
$3$ & $0$ & $4$ & $-1.0$ & $180$   \\ \hline
$4$ & $0$ & $0$ & $-1.5$ & $270$   \\ \hline
\end{tabular}
\end{table}

With the ground truth states, next step is to generate the sensor reading. In the real world, AHRS estimation outputs biased attitude estimation because of the accelerator's bias. To model AHRS bias, we have a simplified AHRS bias model   
\begin{align}
\begin{bmatrix} \phi_b \\ \theta_b \end{bmatrix} = \begin{bmatrix} \cos{\psi} & \sin{\psi} \\ -\sin{\psi} & \cos{\psi} \end{bmatrix}\begin{bmatrix} B_N \\ B_E \end{bmatrix}
\label{equ:AHRS bias model}
\end{align}

where $\phi_b$ and $\theta_b$ are the AHRS biases on $\phi$ and $\theta$. $B_N$ and $B_E$ are the north and east bias caused by the accelerometer bias, which can be considered as constants in short time. From real-world experiments, they are less than $3^{\circ}$. Thus, the AHRS reading can be modelled by

\begin{align}
    \begin{bmatrix} \bar{\phi}_k \\ \bar{\theta}_k \end{bmatrix} = 
    \begin{bmatrix} \phi_k \\ \theta_k \end{bmatrix} + 
    \begin{bmatrix} \cos{\psi} & \sin{\psi} \\ -\sin{\psi} & \cos{\psi} \end{bmatrix}
    \begin{bmatrix} B_N \\ B_E \end{bmatrix} + 
    \begin{bmatrix} \epsilon_{\phi} \\ \epsilon_{\theta} \end{bmatrix}
\end{align}

where $\epsilon_{*} \sim N(0,\sigma_{*})$ is the AHRS noise and in our simulation we will set $\sigma_{*} = 0.5^{\circ}$, $B_N = -2^{\circ}$, $B_E = 1^{\circ}$. For vision measurements generation, we first determine the segment $[u,v]$ of the trajectory where the drone can detect the gate. Then, we calculate the number of the detection by $n_v = \frac {t_u-t_v}{f_{v}}$, where $f_v$ is the detection frequency. Next, we randomly select $n_v$ points between $u$ and $v$ to be vision points. For these points, we generate detection measurement by

\begin{align}
\begin{bmatrix} \bar{x}_k \\ \bar{y}_k \end{bmatrix} = 
    \begin{bmatrix} x_k \\ y_k \end{bmatrix} + 
    \begin{bmatrix} \epsilon_{x} \\ \epsilon_{y} \end{bmatrix}
    \label{equ:vision measurement model}
\end{align}

\begin{figure} [hbt!]
    \centering
    \subfigure[Generated ground truth states and vision measurements in $x-y$ plane]{
\includegraphics[scale=0.4,trim={1cm 6cm 1cm 6cm},clip]{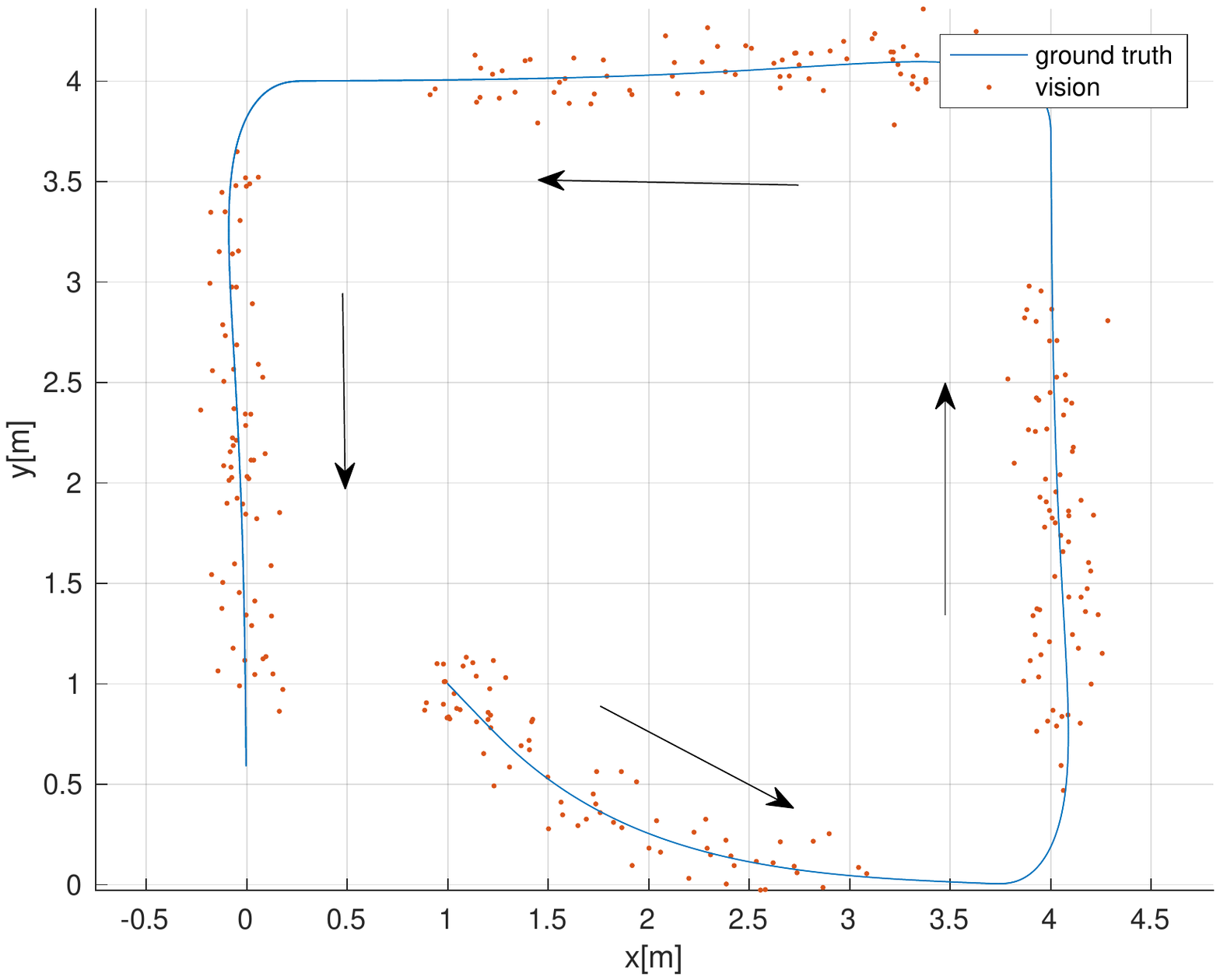}}
\subfigure[Generated ground truth position and vision measurements]{
\includegraphics[scale=0.4,trim={1cm 6cm 1cm 6cm},clip]{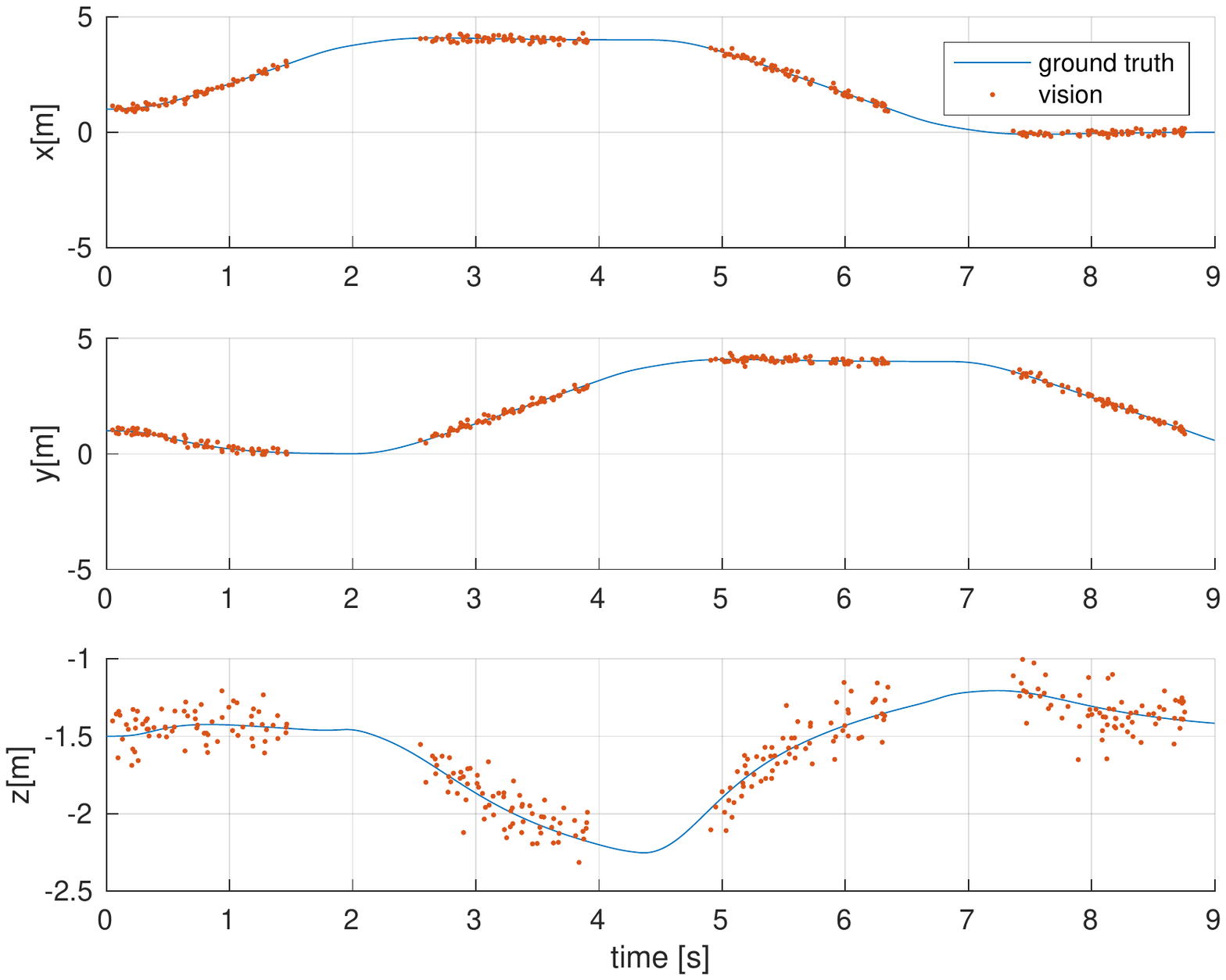}
}
\caption{In the simulation, the ground truth states are first generated (blue curve). Then, vision measurements and AHRS readings are generated. It can be seen clearly that the bias of AHRS readings changes with the heading, as on a real drone. Namely, the offset in $\phi$ and $\theta$ changes when the $\psi$ changes. This phenomenon is modeled by equation \ref{equ:AHRS bias model}. In this simulation $f_v=30$HZ,$\sigma_x = \sigma_y = \sigma_z=0.1m$}.
    \label{fig:simulated gt and sensor}
\end{figure}

In equation \ref{equ:vision measurement model}, $\epsilon_{*}  \sim N(0,\sigma_{*})$ is the detection noise and $\sigma_*=0.1m$ In these $n_v$ vision points, we also randomly select a few points as outlier points, which have the same model with equation \ref{equ:vision measurement model} but $\sigma_{*}=3m$. In the following simulations, the parameters are the same with the value mentioned in this section if there is no statement. The simulated ground truth states and sensor measurements are shown in Figure \ref{fig:simulated gt and sensor}.

\subsection{Simulation result and analysis}
\subsubsection{Comparison between EKF, BRF and PRF without outliers}
\begin{figure} [hbt!]
    \centering
    \subfigure[Position estimation of EKF, BRF and PRF]{\includegraphics[scale=0.4,trim={2cm 8cm 2cm 8cm},clip]{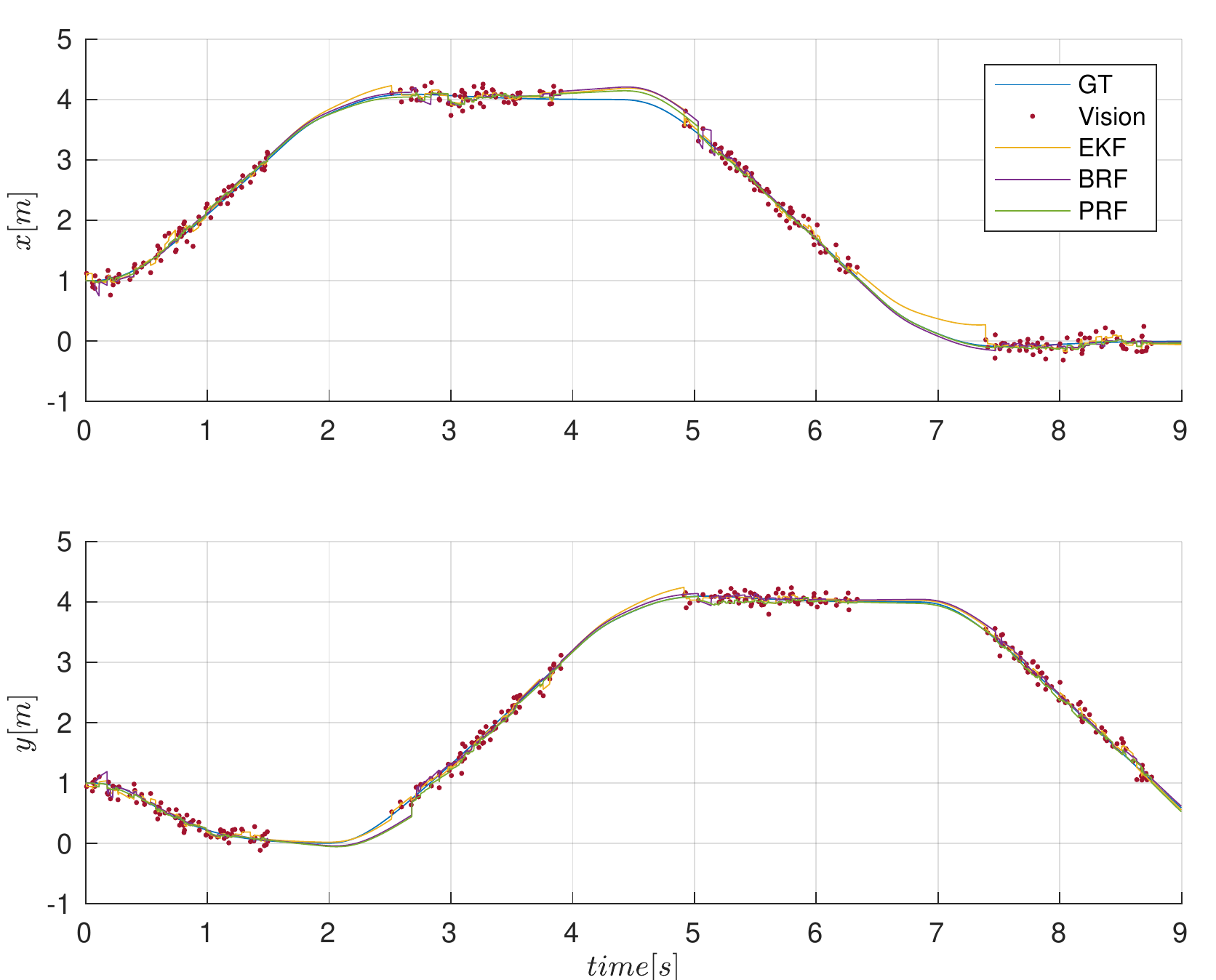}} 
    \subfigure[Velocity estimation of EKF, BRF and PRF]{\includegraphics[scale=0.4,trim={2cm 8cm 2cm 8cm},clip]{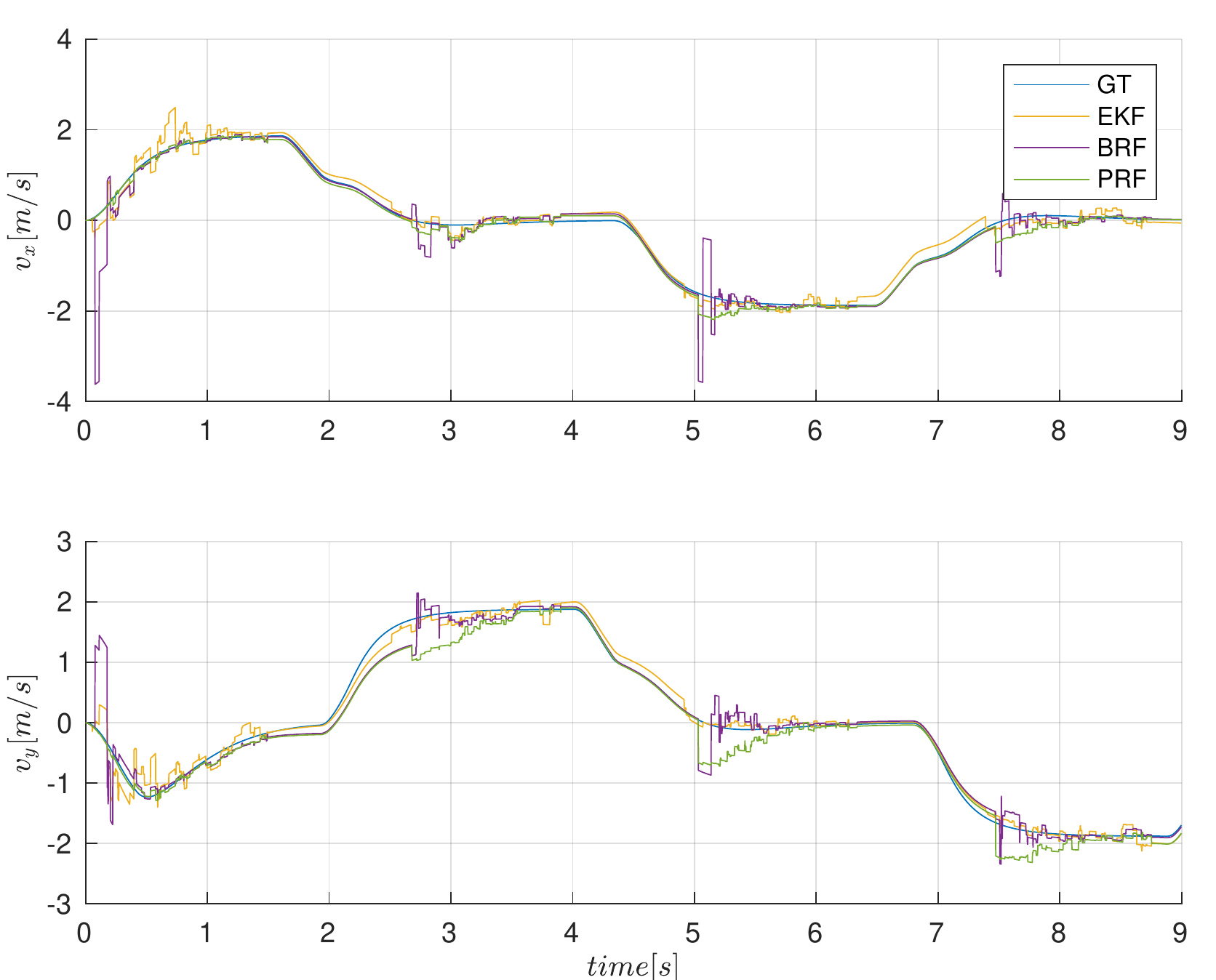}} 
    \caption{The filtering result of EKF, BRF and PRF. $f_v=50HZ$ and $\sigma_x=\sigma_y = 0.1$. When there are no outliers, EKF, BRF and PRF's estimating result all converge to ground truth value. In velocity estimation, however, EKF has longer startup period than VML and BRF shows peaks, which is caused by the over-fitting. To limit this overfitting, in PFR, we add a prior matrix $\mathbf{P} = \begin{bmatrix} 0 & 0 \\ 0 & 0.3 \end{bmatrix}$  and the velocity's peak is significantly smoothed and is closer to the ground truth velocity.}
    \label{fig:filter comparision with no outliers}
\end{figure}

We employ an EKF as benchmarks to compare the performance of our proposed filter. The details of the EKF can be found in the Appendix. We first do the simulation in only one operation point, where $f_v=30$HZ, $\sigma_*= 0.1m$ and the probability of outliers $P_{out}=N_{outliers}/N_{detection}=0$. At this operation point, three filters are run separately. The result is shown in Figure \ref{fig:filter comparision with no outliers}.

When there are no outliers, all three filters can converge to the ground truth value. However, the EKF has a longer startup period and BRF overfits after turning, leading to unlikely high velocity offsets (the peaks in Figure \ref{fig:filter comparision with no outliers}b)). This is because, after the turn, the RANSAC buffer is empty. When the first few detections come into the buffer, the RANSAC has a larger chance to estimate inaccurate parameters. In PRF, however, we add a prior matrix $\mathbf{P} = \begin{bmatrix} 0 & 0 \\ 0 & 0.3 \end{bmatrix}$ to limit the value of $\Delta v$ and the number of the peaks in the velocity estimation is significantly decreased. At the same time, the velocity estimation is closer to the ground truth value.

\begin{figure}[hbt!]
    \centering
    \centering
    \subfigure[Estimation error of the filters with different detection frequencies. ]{\includegraphics[scale=0.5,trim={4cm 8cm 4cm 8cm},clip]{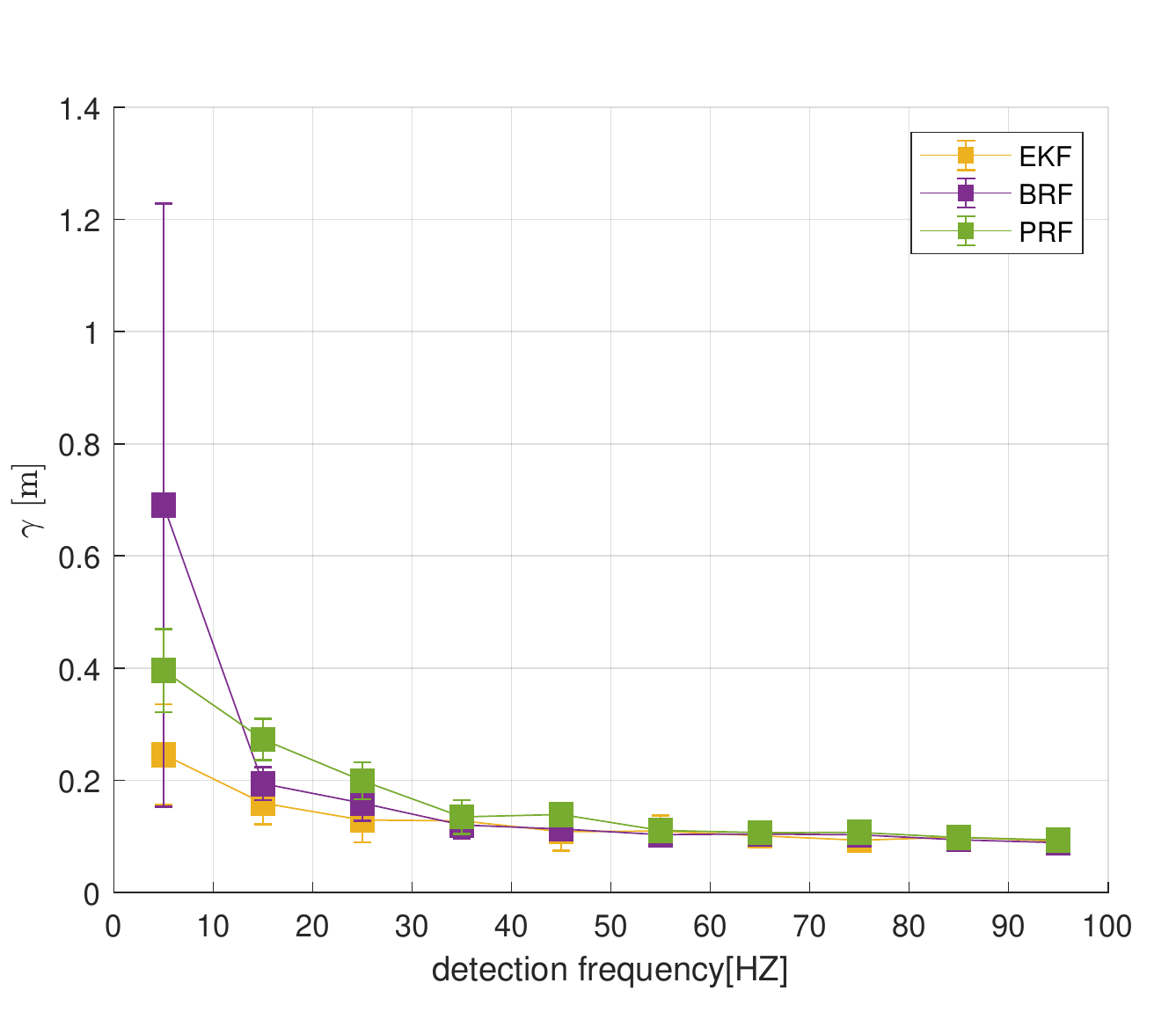}} 
    \subfigure[Calculation time of the filters.]{\includegraphics[scale=0.5,trim={4cm 8cm 4cm 8cm},clip]{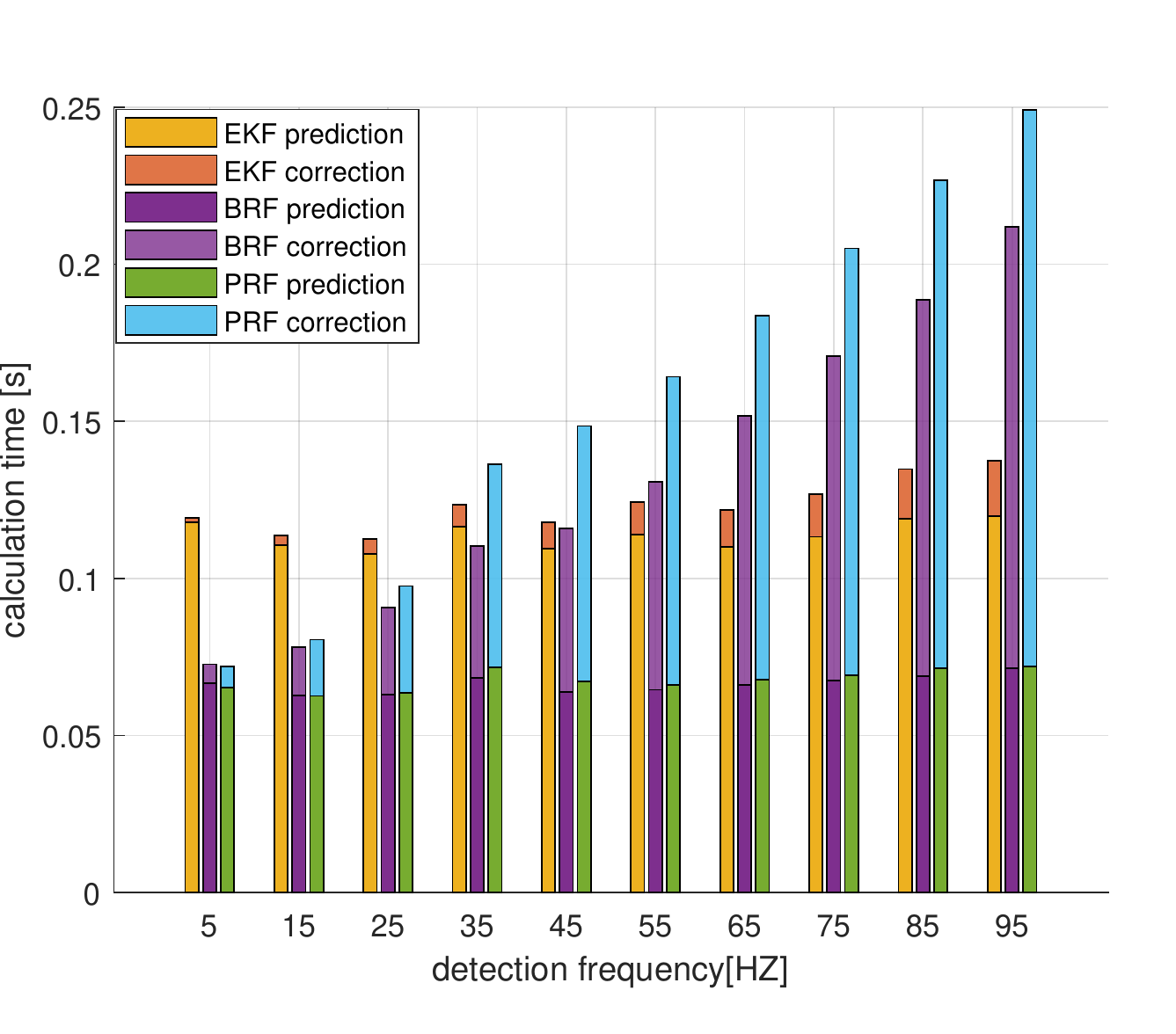}} 
    \caption{The simulation result of the filters. It can be seen that when the detection frequencies are below $20HZ$, the EKF performs better than BRF and PRF. However, when the detection frequencies are higher than $20HZ$, BRF and PRF start performing better than the EKF. In terms of computation time, the EKF is affected by the detection frequency slightly while the computation load of BRF and PRF increase significantly higher detection frequencies}
    \label{fig:performance of the filters without outliers}
\end{figure}

To evaluate the estimation accuracy of each filter, we first introduce a variable, average estimation error $\gamma$, to be an index of the filter's performance:

\begin{align}
    \gamma = \sqrt{\frac{\sum^N_{i=1}(\hat{x}_i-x_i)^2+(\hat{y}_i-y_i)^2}{N}}
    \label{equ:filter index}
\end{align}

where $N$ is the number of the sample points on the whole trajectory. $\hat{x}$ and $\hat{y}$ are the estimated states by the filter. $x$ and $y$ are the ground truth positions generated by the simulation. $\gamma$ captures how much the estimated states deviate from the ground truth states. A smaller $\gamma$ indicates a better filtering result.

We use running time to evaluate the computation efficiency of each filter. It should be noted that since we need to store all the simulation data for visualization and MATLAB has no mechanism of passing pointers, data accessing can take much computation time. Thus, we only count the running time of the core parts of the filters, which are the prediction and the correction. 

The results are shown in Figure \ref{fig:performance of the filters without outliers}. In the simulation, the time-window in BRF and PRF is set to be $1s$ and $5$ iterations are performed in the RANSAC procedure. For each frequency, the filters are run $10$ times separately and their average $\gamma$ and running time are calculated. It can be seen in Figure \ref{fig:performance of the filters without outliers}(a) that when the detection frequency is larger than $30$ HZ, BRF and PRF perform close to the EKF. In terms of calculation time, the EKF is heavier than BRF and PRF when the frequency is lower than $40HZ$. It is because that during the prediction phase, the EKF not only predicts the states but also calculates the Jacobian matrix and the prior error covariance $\mathbf{P}_{k|k-1}$ by high frequency while BRF and PRF only do the state prediction. However, when the detection comes, the EKF does the correction by several matrix operations while BRF and PRF do the RANSAC which is much heavier. This explains why the EKF's computation load is only slightly affected by the detection frequency but BRF and PRF's computation load increases significantly with higher detection frequency.

\subsubsection{Comparison between EKF, BRF and PRF with outliers}
When outliers appear, the regular EKF can be affected significantly. Thus, outlier rejection strategies are always used within an EKF to increase its robustness. A commonly used method is using Mahalanobis distance between the observation and its mean as an index to determine whether an observation is an outlier. \cite{chang2014robust,li2016gps} Thus, in this section, we implement an EKF with outlier rejection (EKF-OR) as a benchmark to compare the outlier rejection performance of BRF and PRF. The basic idea for the EKF-OR is that the square of the observation's Mahalanobis distance is Chi-square distributed. Hence, when the observation arrives, its Mahalanobis distance will be calculated and checked whether it is within a threshold $\chi_{\alpha}$. If it is not, this observation will be rejected. 

\begin{figure} [!hbt]
    \centering
    \subfigure[When outliers appear, EKF-OR, BRF and PRF can reject them.]{\includegraphics[scale=0.45,trim={2cm 8cm 2cm 8cm},clip]{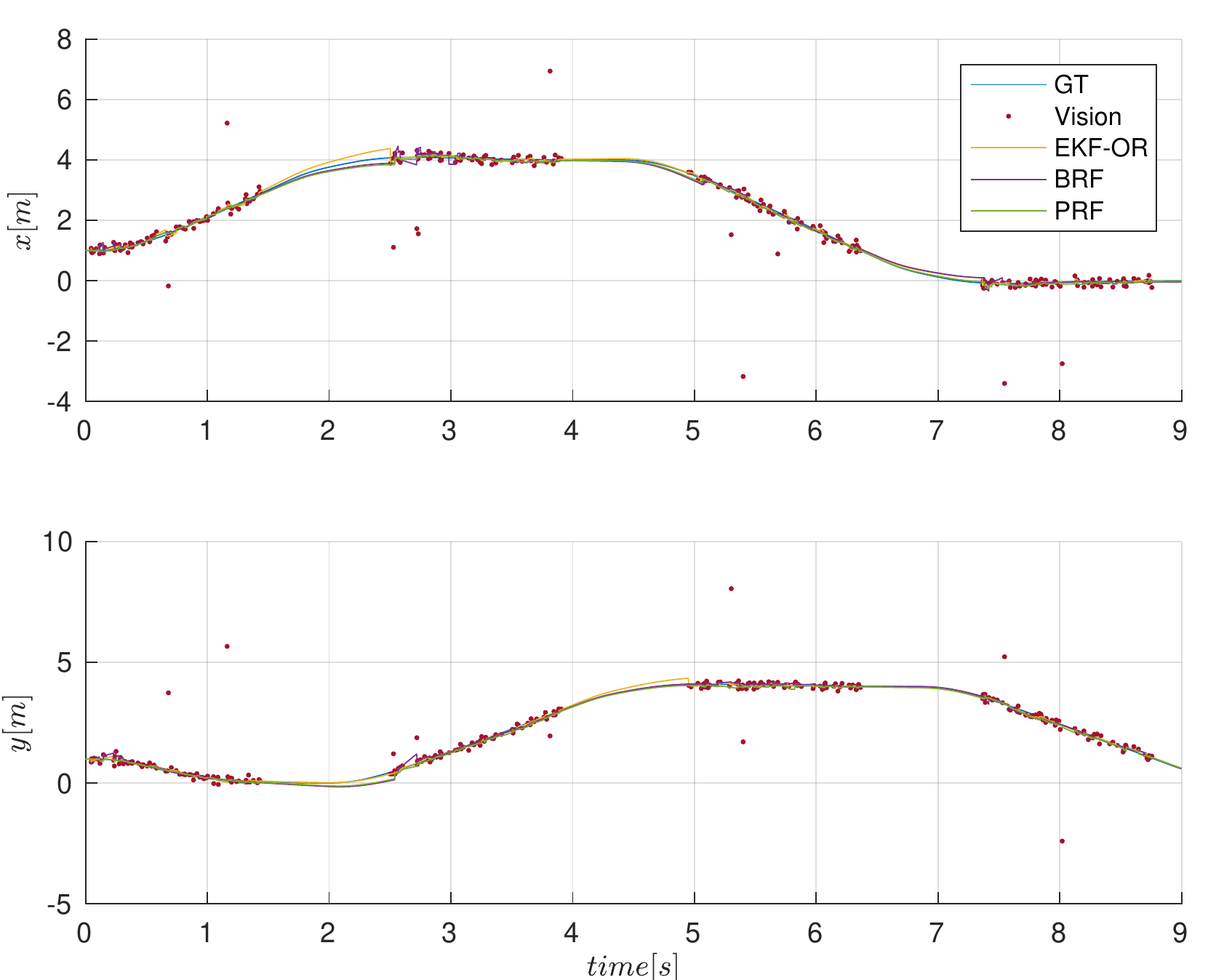}} 
    \subfigure[After a long time of pure prediction, EKF-OR has large error covariance. Once it meets an outlier, it has a high chance to jump to it. As a consequence, the later true positive detections are beyond the threshold $\chi_{\alpha}$ and EKF-OR will treat them as outliers]{\includegraphics[scale=0.45,trim={2cm 8cm 2cm 8cm},clip]{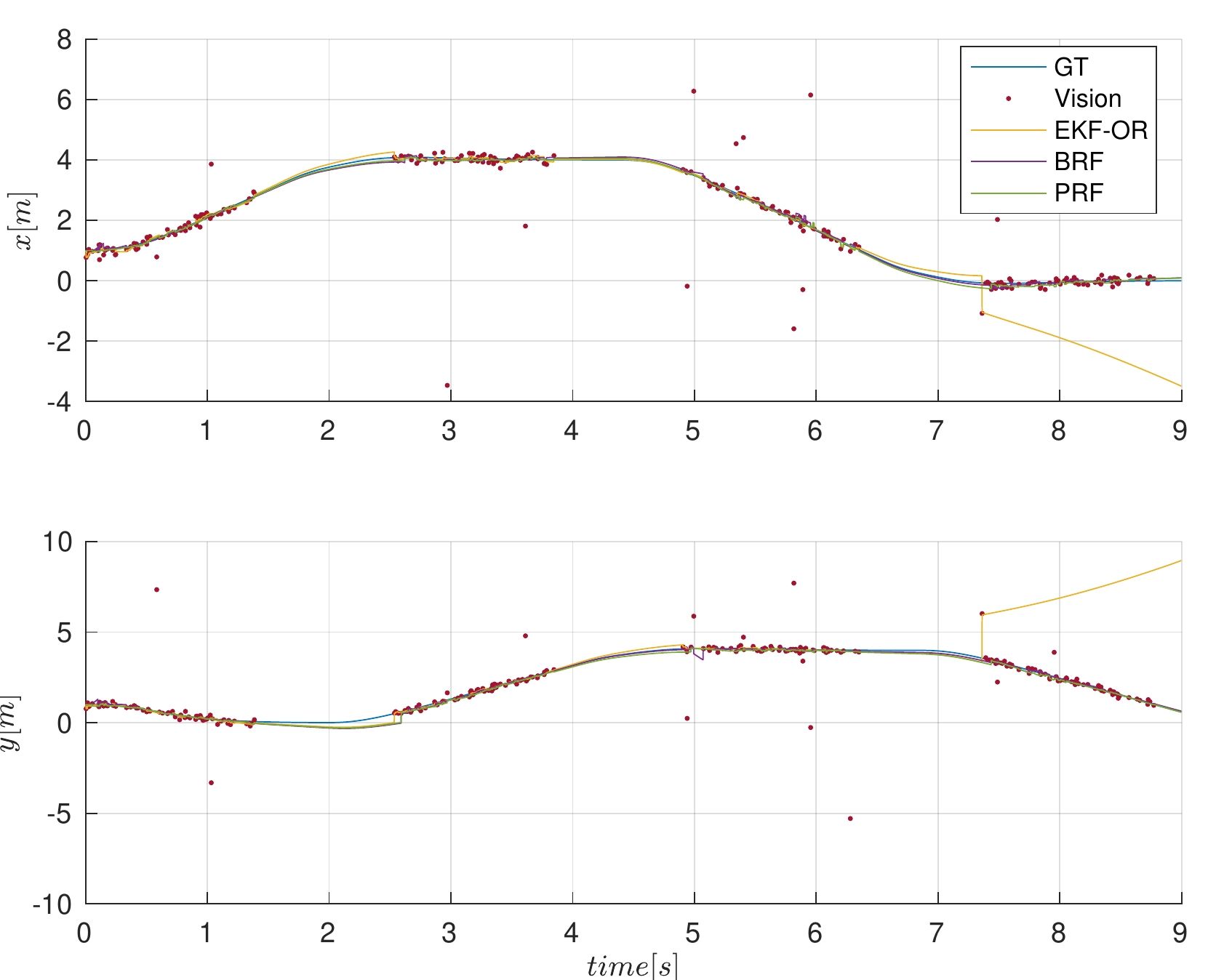}} 
    \caption{In most cases, EKF-OR, BRF and PRF can reject the outliers. But after a long time of pure prediction, EKF-OR is very vulnerable to the outliers while BRF and PRF still perform well.}
    \label{fig:EKF-OR normal and diverge}
\end{figure}

Two examples of the filters' rejecting outliers are shown in Figure \ref{fig:EKF-OR normal and diverge}. The first figure shows a common case that the three filters can reject the outliers successfully. However, in some special cases, EKF-OR is vulnerable to the outliers. In Figure \ref{fig:EKF-OR normal and diverge}(b), for instance, after a long time of pure prediction, the error covariance $\mathbf{P}_{k|k-1}$ becomes large. Once EKF-OR meets an outlier, it has a high chance to jump to it. The subsequent true positive detections will be treated as outliers and EKF-OR starts diverging. At the same time, BRF and PRF are more robust to the outliers. The essential reason is that for EKF-OR, it depends on its current state estimation (mean and error covariance) to identify the outliers. When the current state estimation is not accurate enough, like the long-time prediction in our case, EKF-OR loses its ability to identify outliers. In other words, it tends to trust whatever it meets. The worse situation is that after jumping to the outlier, its error covariance become smaller which, as a consequence, leads to the rejection of the coming true positive detections. However, for BRF and PRF, outliers are determined in a time window including history. Thus, after long time of prediction, when BRF and PRF meet an outlier, they will judge it considering the detections in the past. If there is no other detection in the time window, they will wait for enough detections to make a decision. With this mechanism, BRF and PRF become more robust than EKF-OR especially when EKF-OR's estimation is not accurate.

\begin{figure} [!hbt]
    \centering
    \subfigure[Estimation error of the EKF-OR, BRF and PRF with different detection frequencies]{\includegraphics[scale=0.35,trim={4cm 8cm 4cm 8cm},clip]{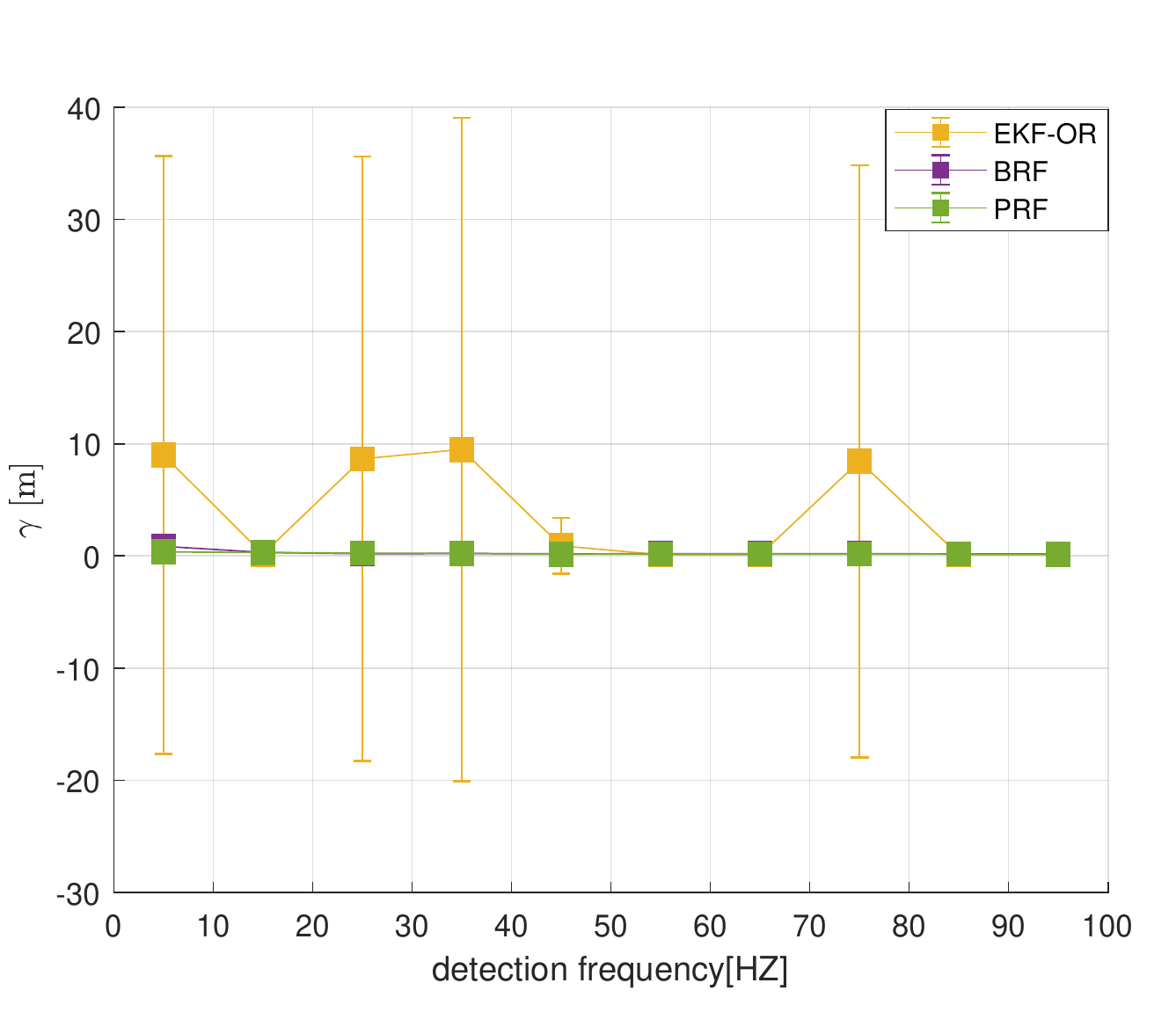}} 
      \subfigure[Partial enlarged drawing of (a)]{\includegraphics[scale=0.35,trim={4cm 8cm 4cm 8cm},clip]{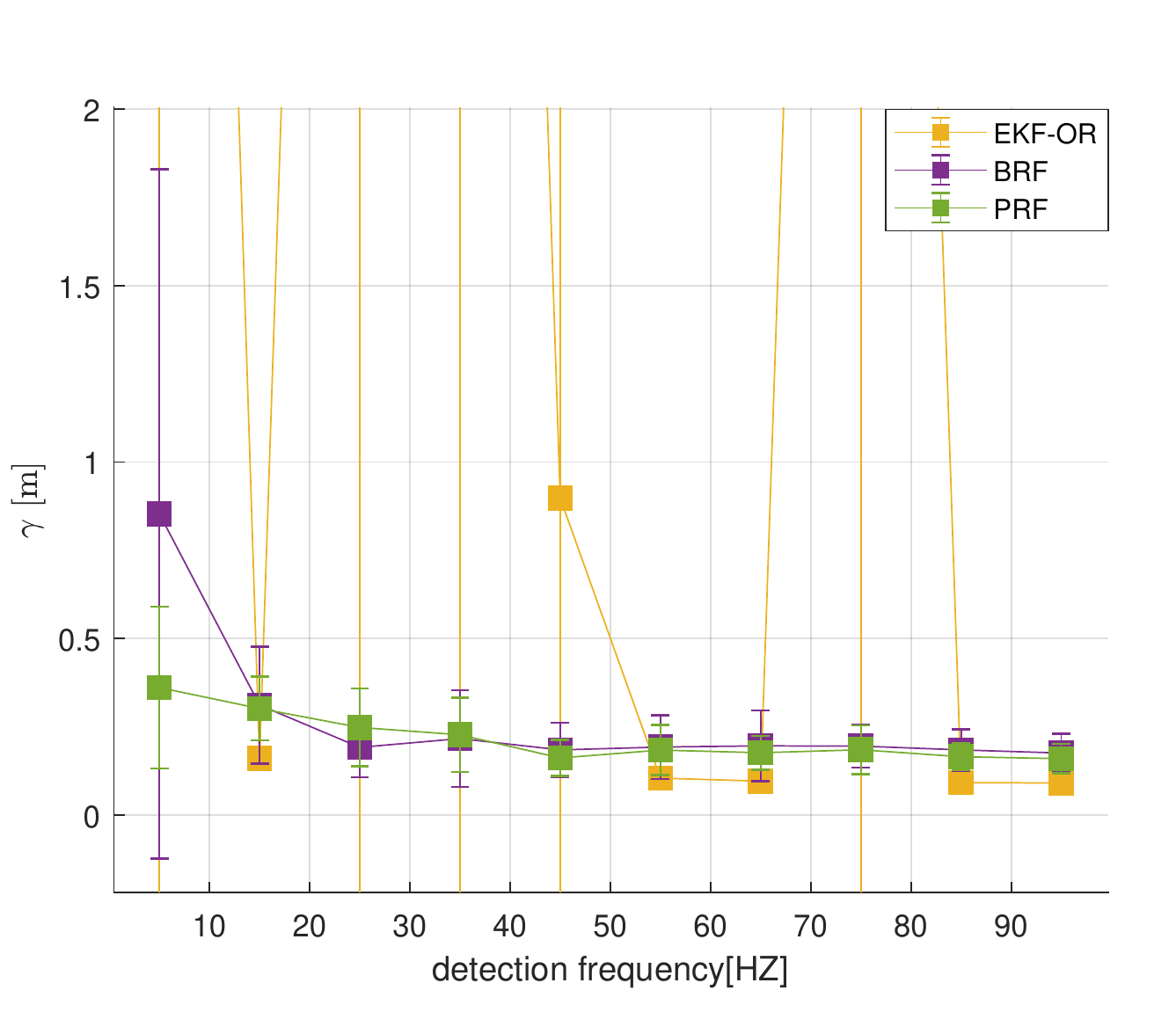}} 
    \subfigure[ Calculation time of the filters]{\includegraphics[scale=0.35,trim={4cm 8cm 4cm 8cm},clip]{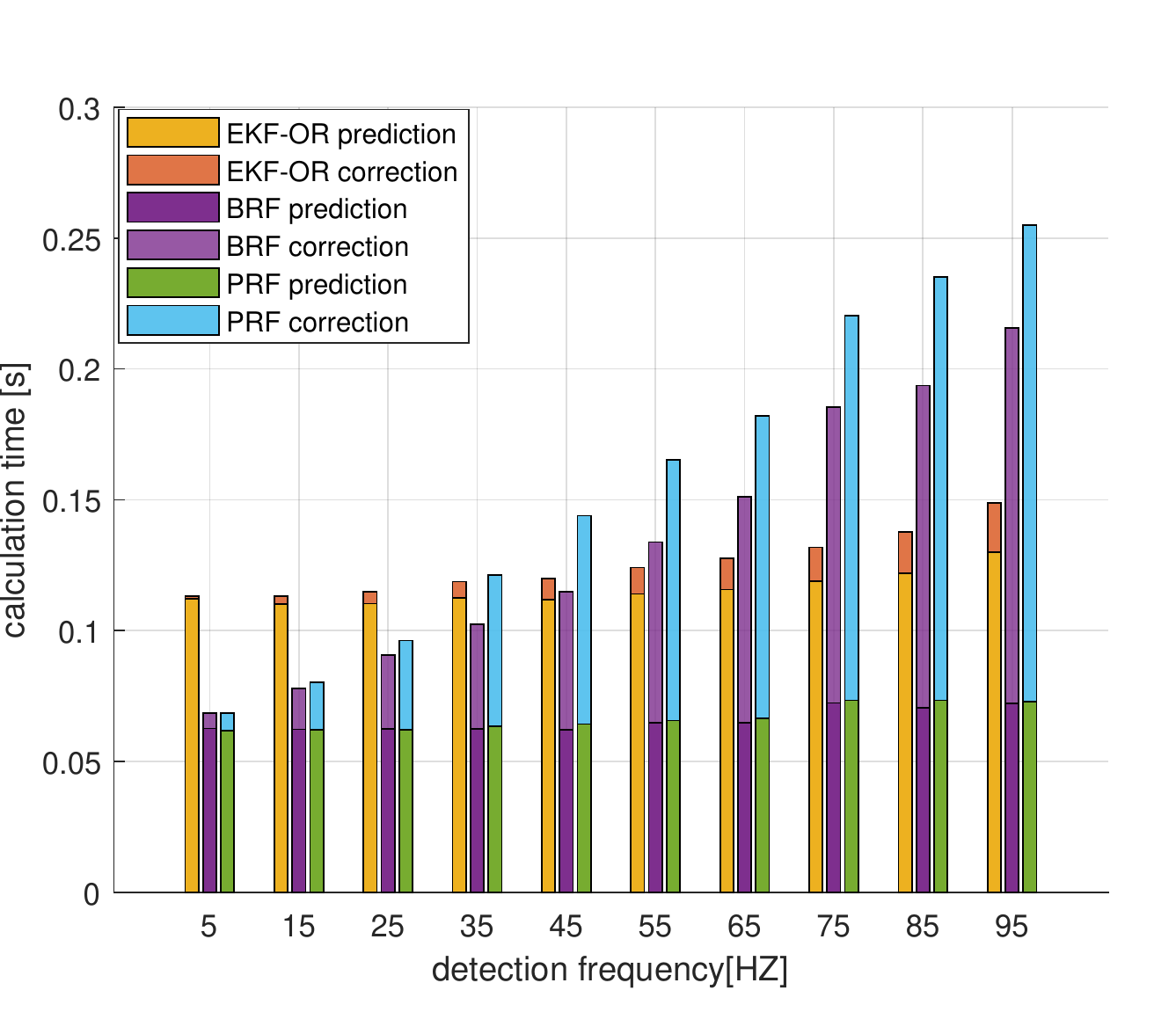}} 
    \caption{The estimation error of EKF-OR, BRF and PRF and their calculation time with outliers. EKF-OR has some chance ($15\%$) to diverge, which leads to the high estimation error.}
    \label{fig:error_and_time_outliers}
\end{figure}

Figure \ref{fig:error_and_time_outliers} shows the estimation error and the calculation time of the three filters. As we stated before, although EKF-OR has the mechanism of dealing with the outliers, it still can diverge due to the outliers in some special cases. Thus, in Figure \ref{fig:error_and_time_outliers}(a) EKF-OR has large estimation error when the detection frequency is both low and high. In terms of calculation time, it can be seen that it has no significant difference with the non-outlier case.

\subsubsection{Filtering result with delayed detection}
\begin{figure} [!hbt]
    \centering
    \subfigure[The sketch of the EKF-DH proposed in \cite{weiss2012versatile}. When the measurement arrives at $t_k$, EKF-DH first corrects the corresponding states at $t_s$ and then updates the states until $t_k$.]{\includegraphics[scale=0.35,trim={0cm 0cm 0cm 0cm},clip]{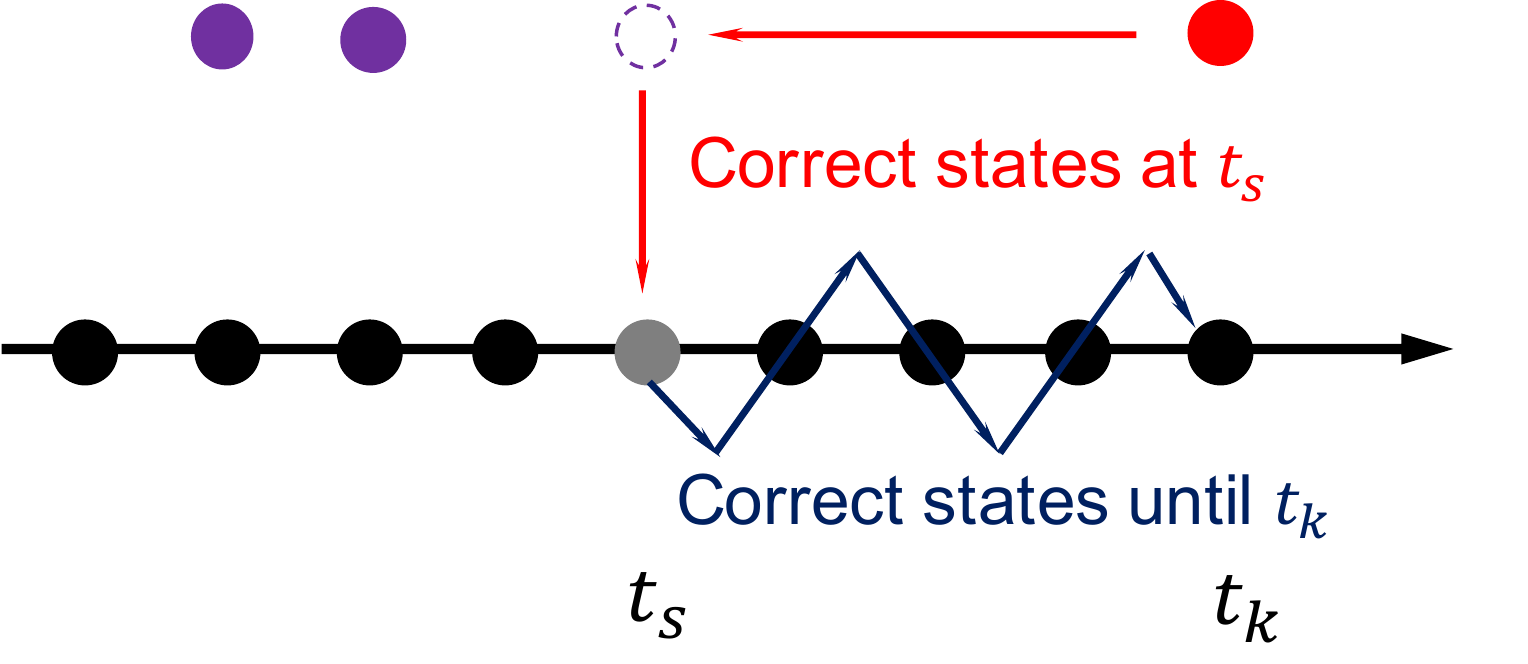}} \hspace{1cm}
    \subfigure[The sketch of VML's mechanism of handling delay. When the measurement arrives, it will be pushed to the buffer with the corresponding states. Then, the error model will be estimated by the RANSAC approach. At last, the estimated model will be used to compensate the prediction at $t_k$. There is no need to update all the states between $t_s$ and $t_k$]{\includegraphics[scale=0.35,trim={0cm 0cm 0cm 0cm},clip]{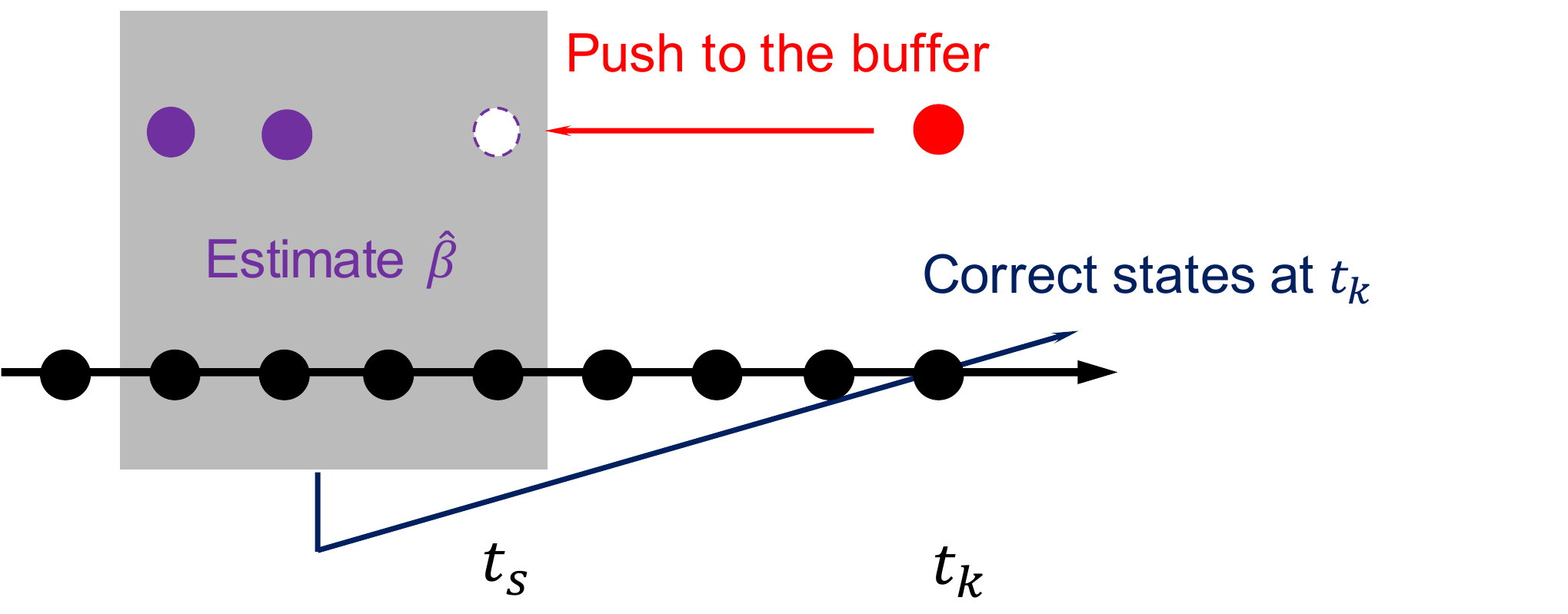}} 
    \caption{The sketches of EKF-DH and VML's handling delay mechanism.}
    \label{fig:sketches delay}
\end{figure}

Image processing and visual algorithms can be very computationally expensive for running onboard a drone, which can lead to significant delay. \cite{van2019event,weiss2012versatile} Many visual navigation approaches ignore this delay and directly fuse the visual measurements with the onboard sensors, which sacrifices the accuracy of the state estimation. A commonly used approach for compensating this vision delay is a modified Kalman filter proposed by Weiss et al. \cite{weiss2012versatile}. The main idea of this approach, called EKF delay handler (EKF-DH), is having a buffer to store all sensor measurements within a certain time. At time $t_k$, a vision measurement corresponding to the states at earlier time $t_s$ arrives. It will be used to correct the states at time $t_s$. Then, the states will be propagated again from $t_s$ to $t_k$. (Figure \ref{fig:sketches delay}(a)) Although updating the covariance matrix is not needed according to \cite{weiss2012versatile}, this approach still requires updating history states whenever a measurement arrives, which can be computationally expensive especially when the delay and the measurement frequency get larger. In our case, we need to use the error covariance for outlier rejections, it is necessary to update the history error covariance matrices, which in turn increases the computation load further. At the same time, for VML, when the measurement arrives, it will first be pushed into the buffer. Then, the error model will be estimated within the buffer/time window. With the estimated parameter $\hat{\beta}$, the prediction at $t_k$ can be corrected directly without the need of correcting all the states between $t_s$ and $t_k$. (Figure \ref{fig:sketches delay}(b)) Thus, the computational burden will not increase when the delay exists.

\begin{figure} [!hbt]
    \centering
    \subfigure[Position estimation of the three filters with outliers and delay]{\includegraphics[scale=0.35,trim={1cm 8cm 1cm 8cm},clip]{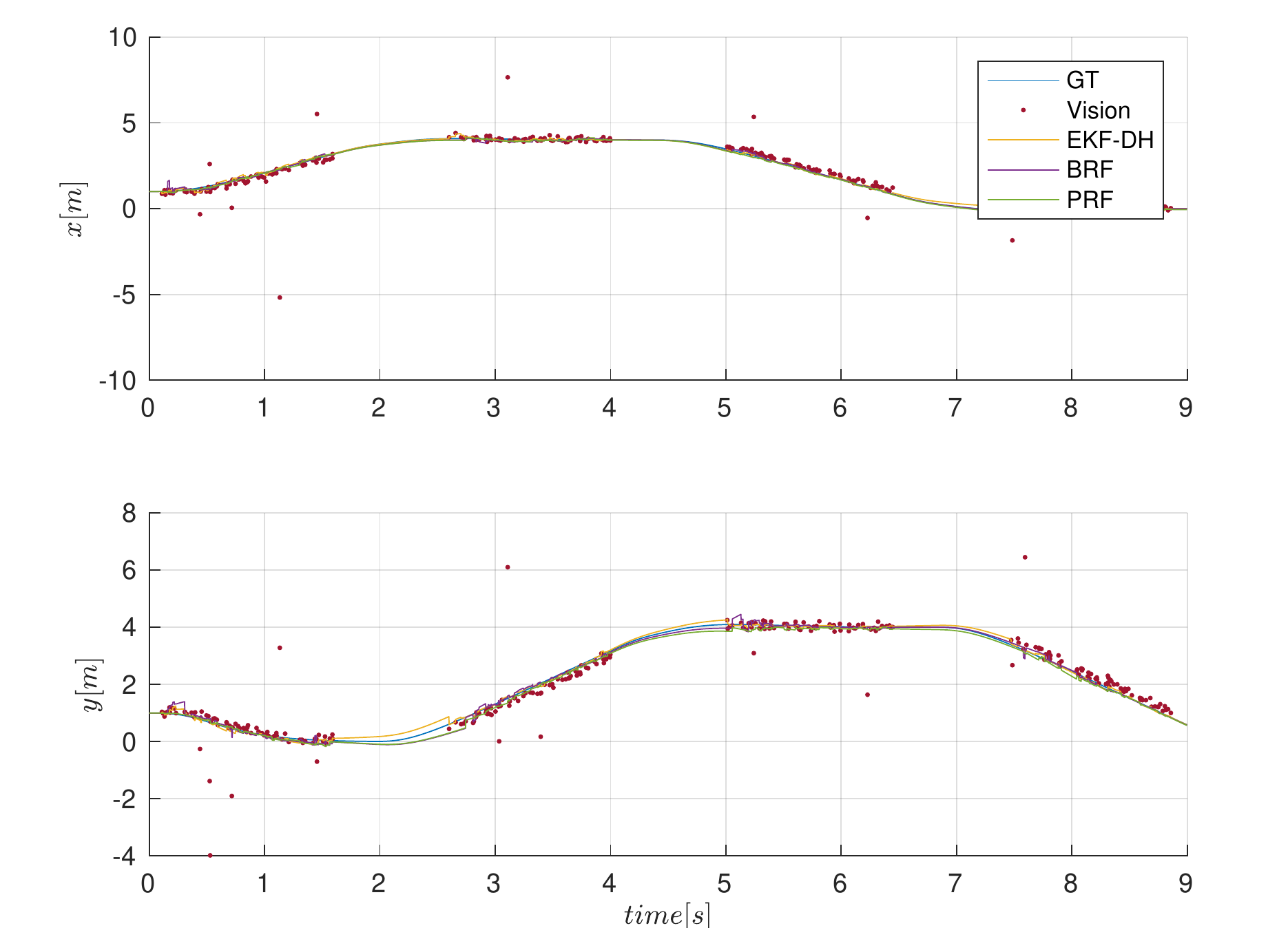}} \hspace{2mm}
    \subfigure[Velocity estimation of the three filters with outliers and delay]{\includegraphics[scale=0.35,trim={1cm 8cm 1cm 8cm},clip]{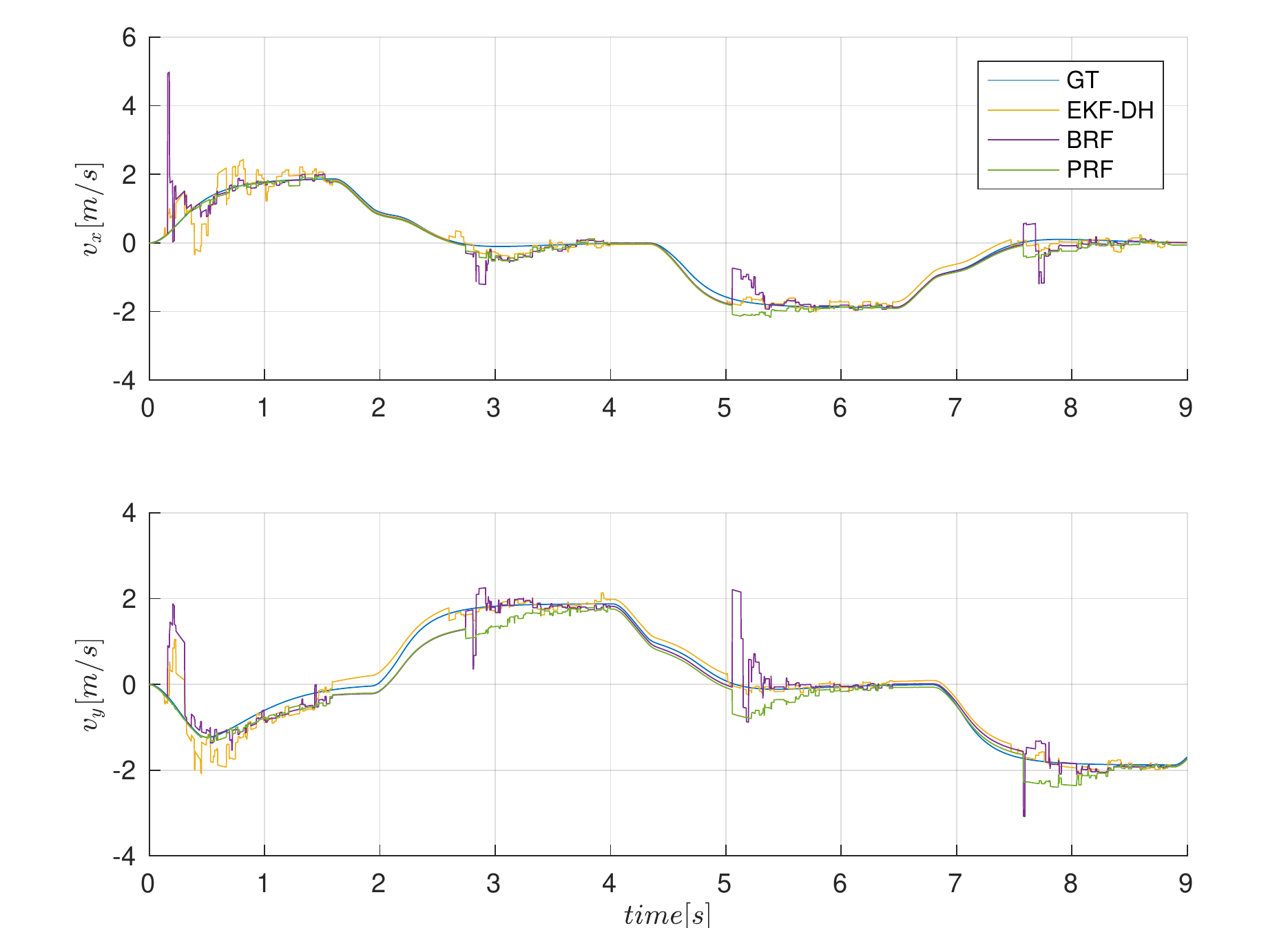}} 
    \caption{An example of the performance of the three filters when outliers and delay exist.}
    \label{fig:example delay}
\end{figure}

\begin{figure} [!hbt]
    \centering
    \subfigure[Estimation error of the EKF-DH, BRF and PRF with different detection frequencies]{\includegraphics[scale=0.3,trim={2cm 8cm 2cm 8cm},clip]{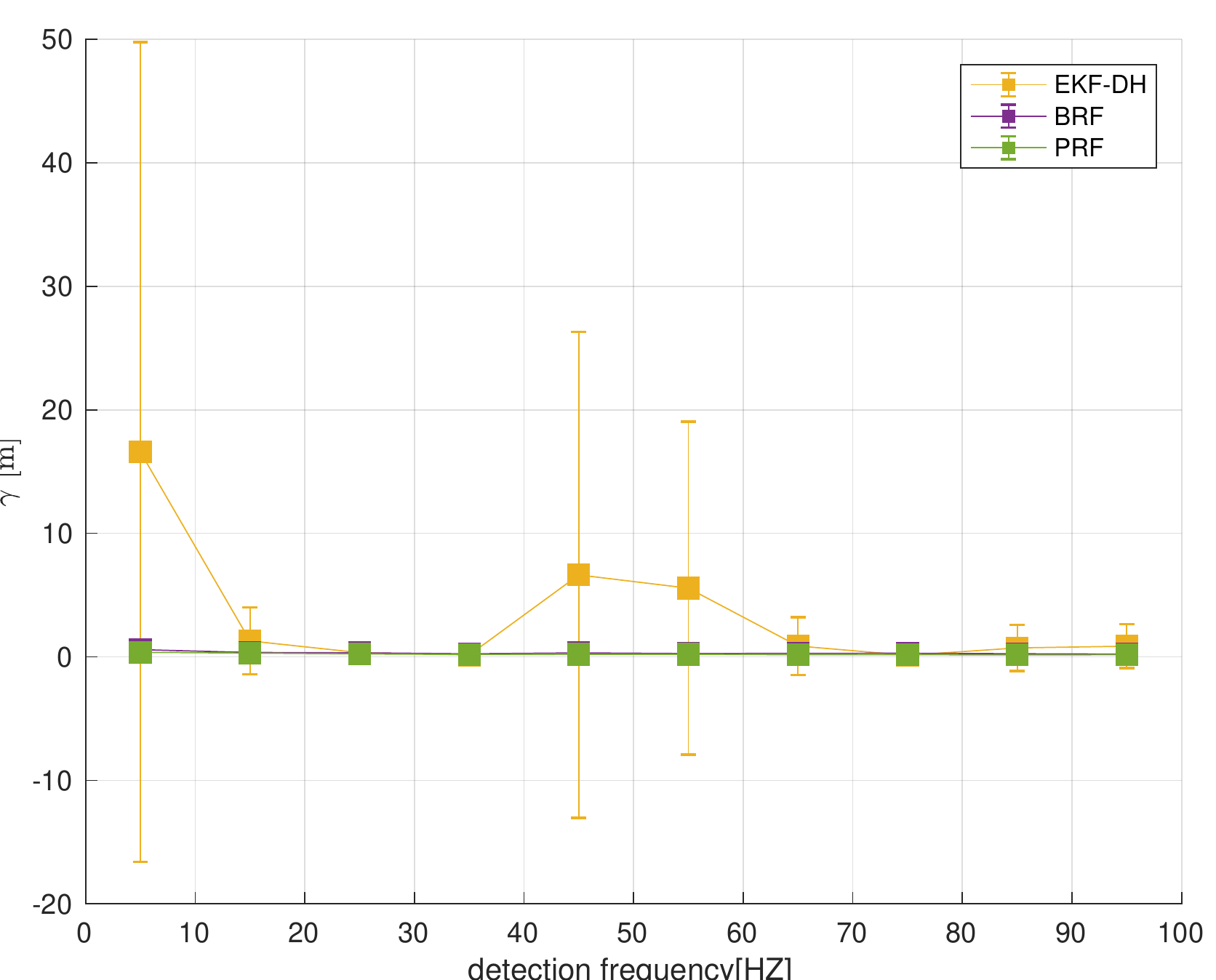}} 
    \subfigure[Partial enlarged drawing of (a)]{\includegraphics[scale=0.3,trim={2cm 8cm 2cm 8cm},clip]{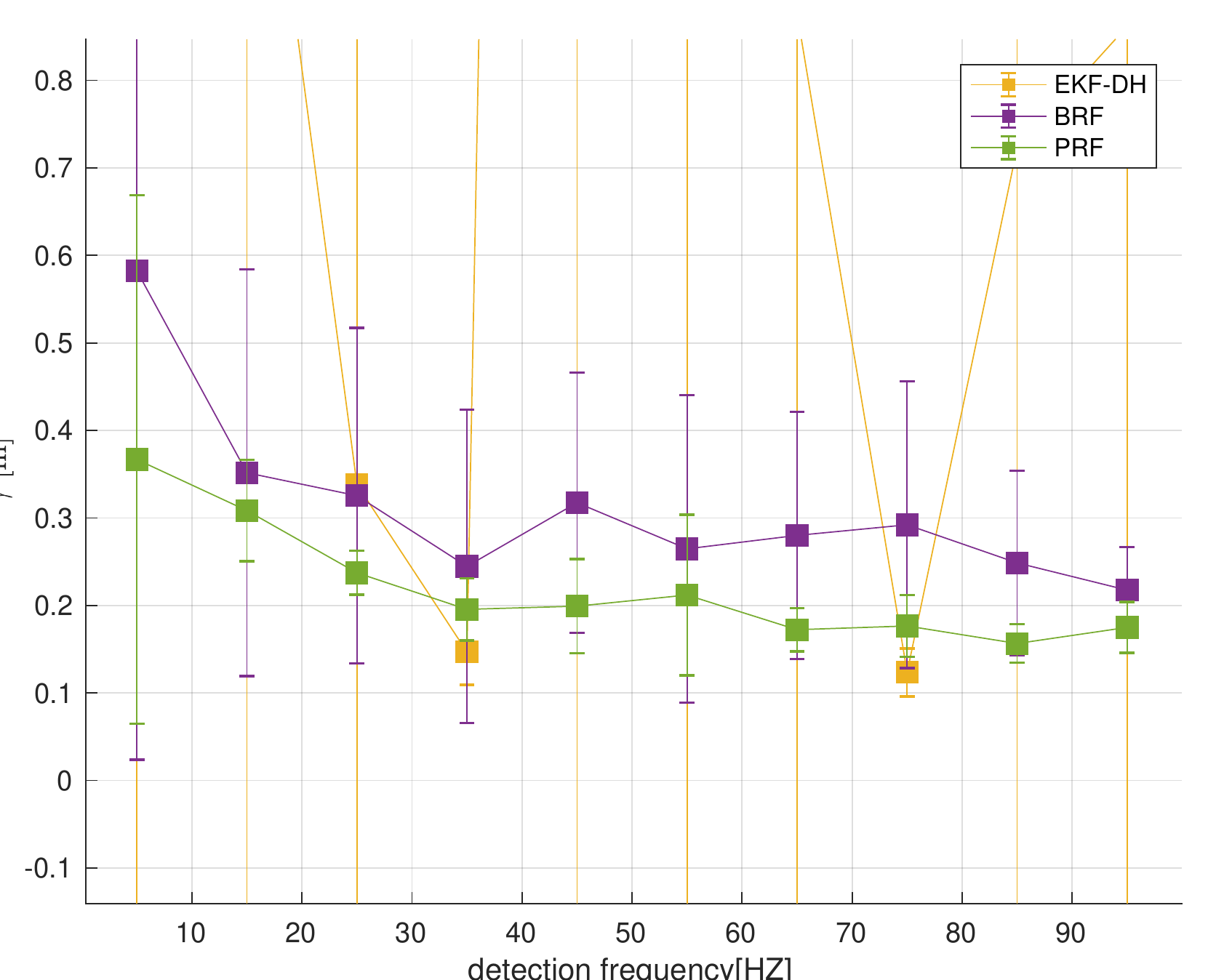}} 
    \subfigure[ Calculation time of the filters]{\includegraphics[scale=0.3,trim={2cm 8cm 2cm 8cm},clip]{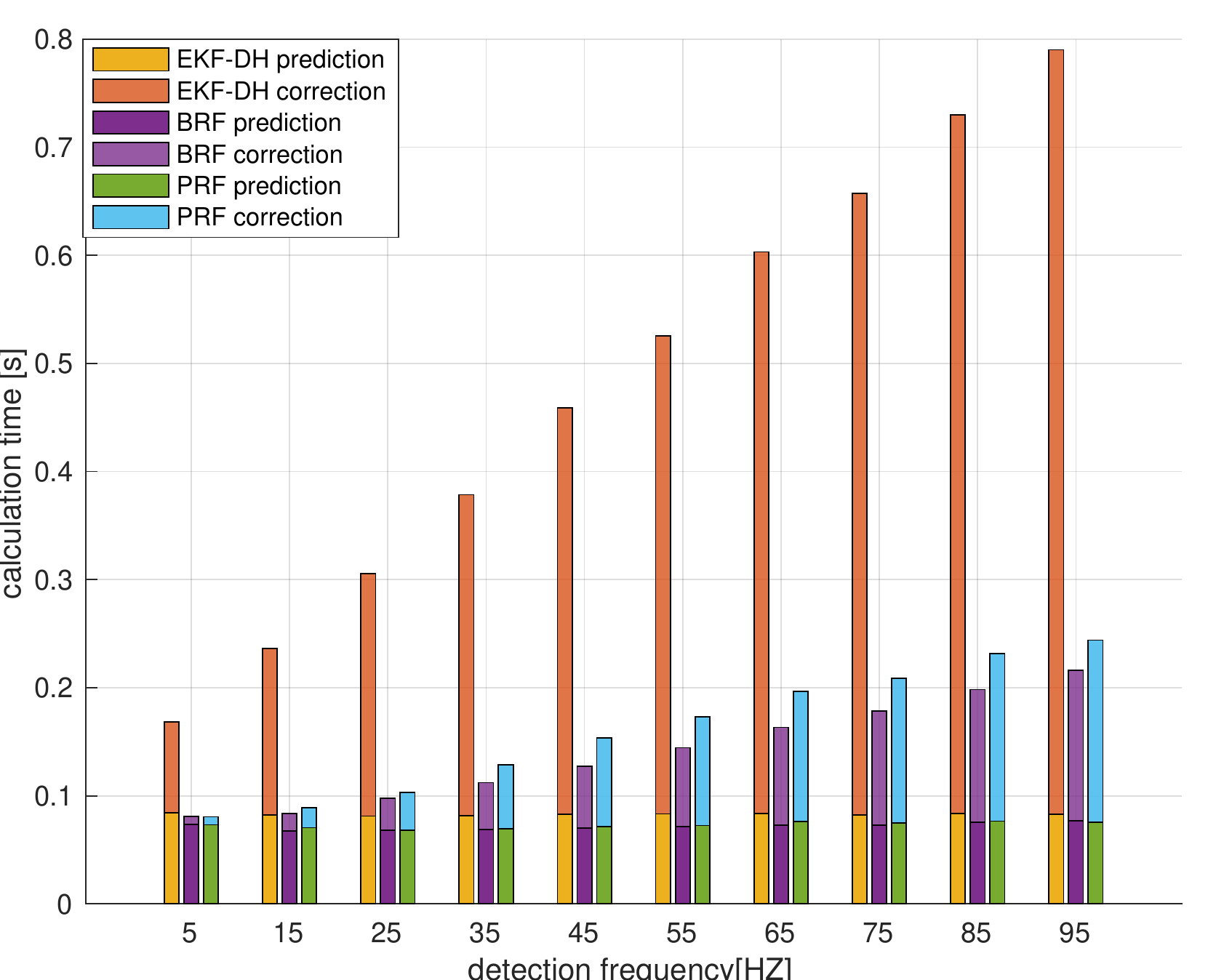}} 
    \caption{The estimation error of EKF-DH, BRF and PRF and their calculation time with outliers and delay.}
    \label{fig:error_and_time_delay}
\end{figure}

Figure \ref{fig:example delay} shows an example of the simulation result of the three filters when both outliers and delay exist. In this simulation, the visual delay is set to be $0.1s$. It can be seen that although there is a lag between the vision measurements and the ground-truth, all the filters can estimate accurate states. However, EKF-DH requires much more computation effort. Figure \ref{fig:error_and_time_delay} shows the estimation error and the computation time of the three filters.

In Figure \ref{fig:error_and_time_delay}, we can see that the computation load of EKF-DH increases significantly due to its mechanism of handling delay. Unsurprisingly, EKF-DH is still sensitive to some outliers while BRF and PRF can handle the outliers.  

\section{Real-world Experiments}
\label{sec:Experiment Result}

\subsection{Processing time of each component}
Before testing the whole system, we first test on the ground how much time the Snake gate detection, the VML and the controller take when running on a Jevois smart camera. On the ground, we set an orange gate in front of a Jevois camera and calculate the time that each component takes. For each image, we start timing when a new image arrives and the Snake gate detection is run. Then, we stop timing when the snake gate finishes. For VML, in each loop, the timing includes both prediction and correction no matter if there are enough detections for correction. We start counting when the Jevois is powered on. In this test, the vision detection frequency is $15HZ$ and the number of RANSAC iterations in VML is set to $5$. Table \ref{tab:processing time} shows the statistical results of the time each component takes on the Jevois.

\begin{table}[H]
\caption{The processing time of each component of the approach running on the Jevois}
\centering
\begin{tabular}{|c|c|c|}
\hline
\centering
Snake gate detection (each image) & VML (each loop) & Controller (each loop) \\ \hline \hline
$17\pm 2.5 ms$ & $0.02\pm 0.15ms$ & $0.01\pm 0.1ms$ \\ \hline 
\end{tabular}
\label{tab:processing time}
\end{table}

From Table \ref{tab:processing time}, it can be seen that vision takes much more time than the other two parts. Please note though that the snake gate computer vision detection algorithm is already a very efficient gate detection algorithm. In fact, it has tunable parameters, i.e., the number of samples taken per image for the detection (3000 in the current setup), which allow the algorithm to run even much faster at the cost of having less accuracy (see \cite{li2018autonomous} for more details). The main gain in time in the approach presented in this article is that we do not employ VIO and SLAM, which would take substantially more processing. However, as the Snake gate detection provides relatively low-frequency and noisy position measurements, the VML needs to run in high frequency and cope with the detection noise to still provide accurate estimation for the controller.

\subsection{Flying experiment without gate displacement}
\begin{figure} [hbt!]
    \centering
    \includegraphics[width=0.6\columnwidth,trim={0cm 0cm 0 0cm},clip]{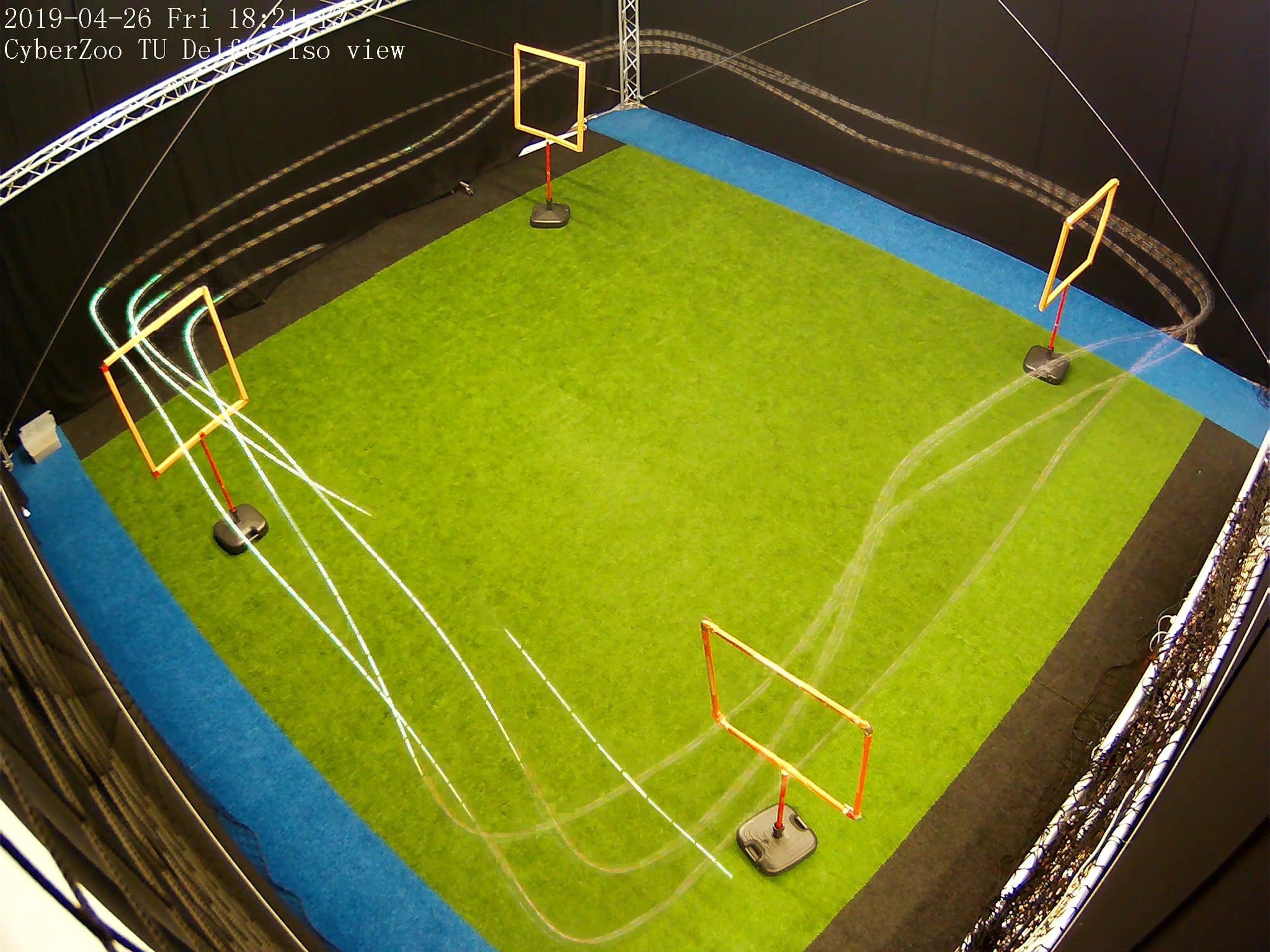}
\caption{The picture of the Trashcan flying the track where the gates are displaced. The average speed is $2m/s$ and the maximum speed is $2.6m/s$.}
\label{fig:flying picture}
\end{figure}

\begin{figure} [!hbt]
    \centering
    \includegraphics[width=0.7\columnwidth,trim={0cm 7cm 0cm 8cm},clip]{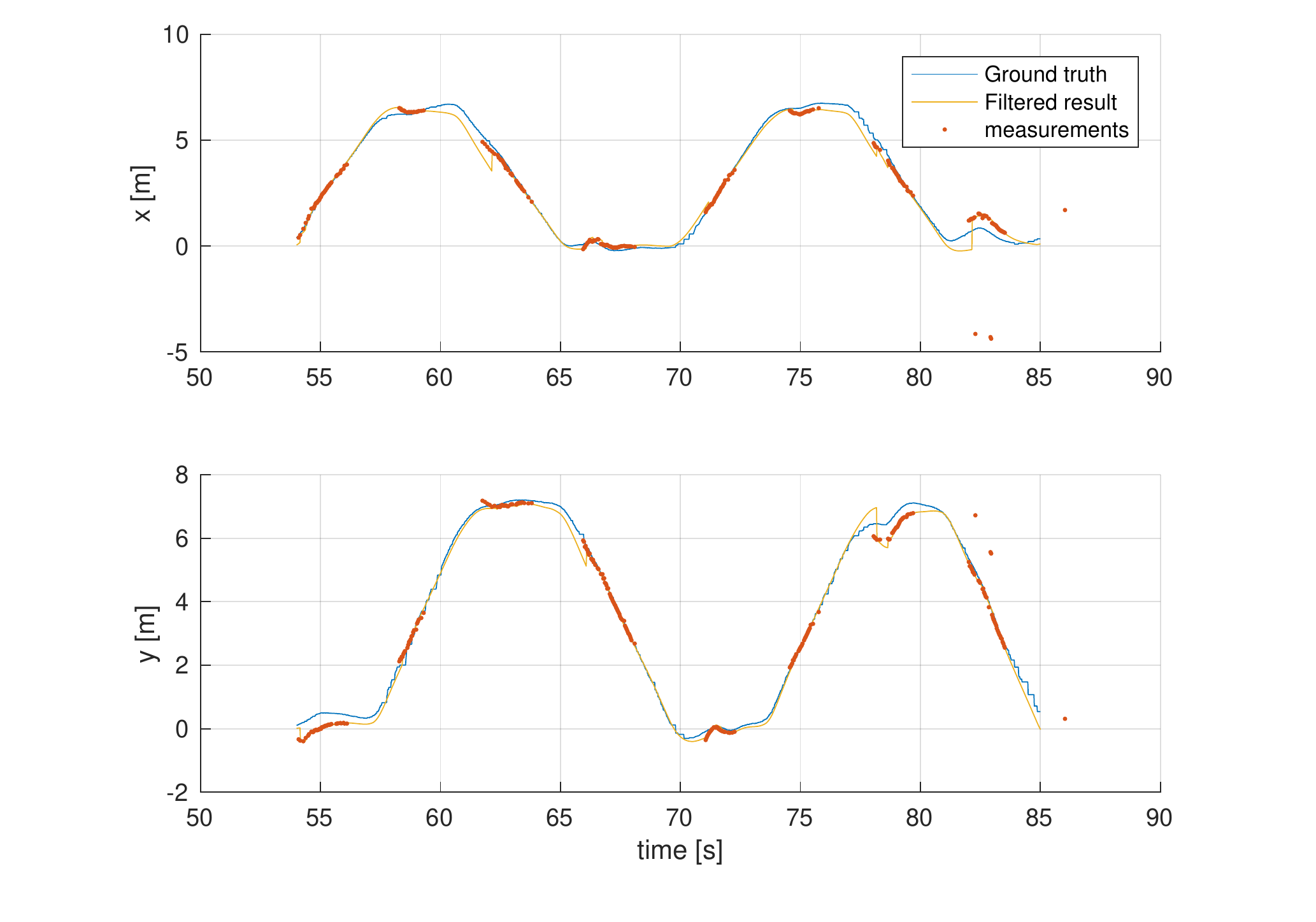}
\caption{The flying result of the drone flying the track without the gate displacement.}
\label{fig:2_laps_log}
\end{figure}

Figure \ref{fig:2_laps_log} shows the flying result of the drone flying the track without gate displacement. The position of the $4$ gates is listed in Table \ref{tab:race track no diaplacement}. In Table \ref{tab:race track no diaplacement}, $x_g$ and $y_g$ are the position of the gates in the real world and $\Tilde{x}_g$ and $\Tilde{y}_g$ are their position on the map. In this situation, they are the same. The aim of this experiment is to test the filter's performance with sufficient detections. Thus, the velocity is set to be $1.5m/s$ to give the drone more time to detect the gate. In Figure \ref{fig:2_laps_log}, the blue curve is the ground truth data from Optitrack motion capture system and the yellow curves are the filtering results. From the flying result, it can be seen that the filtered results are smooth and coincide with the ground truth position well. During the period when the detections are not available, the state prediction is still accurate enough to navigate the drone to the next gate. When the drone detects the next gate, the filter will correct the prediction. In this situation, the divergence of the states is only caused by the prediction drift. It should also be noted that when the outliers appears at $84s$, the filter is not affected by them because of the RANSAC technique in the filter. The processing time of the visual detection, the filter and the controller are listed in Table \ref{tab:processing time}. It can be seen that the VML proposed in this article is extremely efficient.

 \begin{table}[H]
\caption{The position of the gates without displacement}
\centering
\begin{tabular}{|c|c|c|c|c|}
\hline
\centering
gate ID & $x_g[m]$ & $y_g[m]$ & $\Tilde{x}_g[m]$ &  $\Tilde{y}_g[m]$ \\ \hline \hline
1 & 5 & 0 & 5 & 0    \\ \hline
2 & 6.5 & 5 & 6.5 & 5    \\ \hline
3 & 1 & 7 & 1 & 7    \\ \hline
4 & 0 & 1 & 0 & 1    \\ \hline
\end{tabular}
\label{tab:race track no diaplacement}
\end{table}

\subsection{Flying experiment with gate displacement}
\begin{figure} [hbt!]
    \centering
    \subfigure[The map of the race track where the gates in the real world are displaced.]{ \includegraphics[width=0.3\columnwidth,trim={0cm 0cm 0cm 0cm},clip]{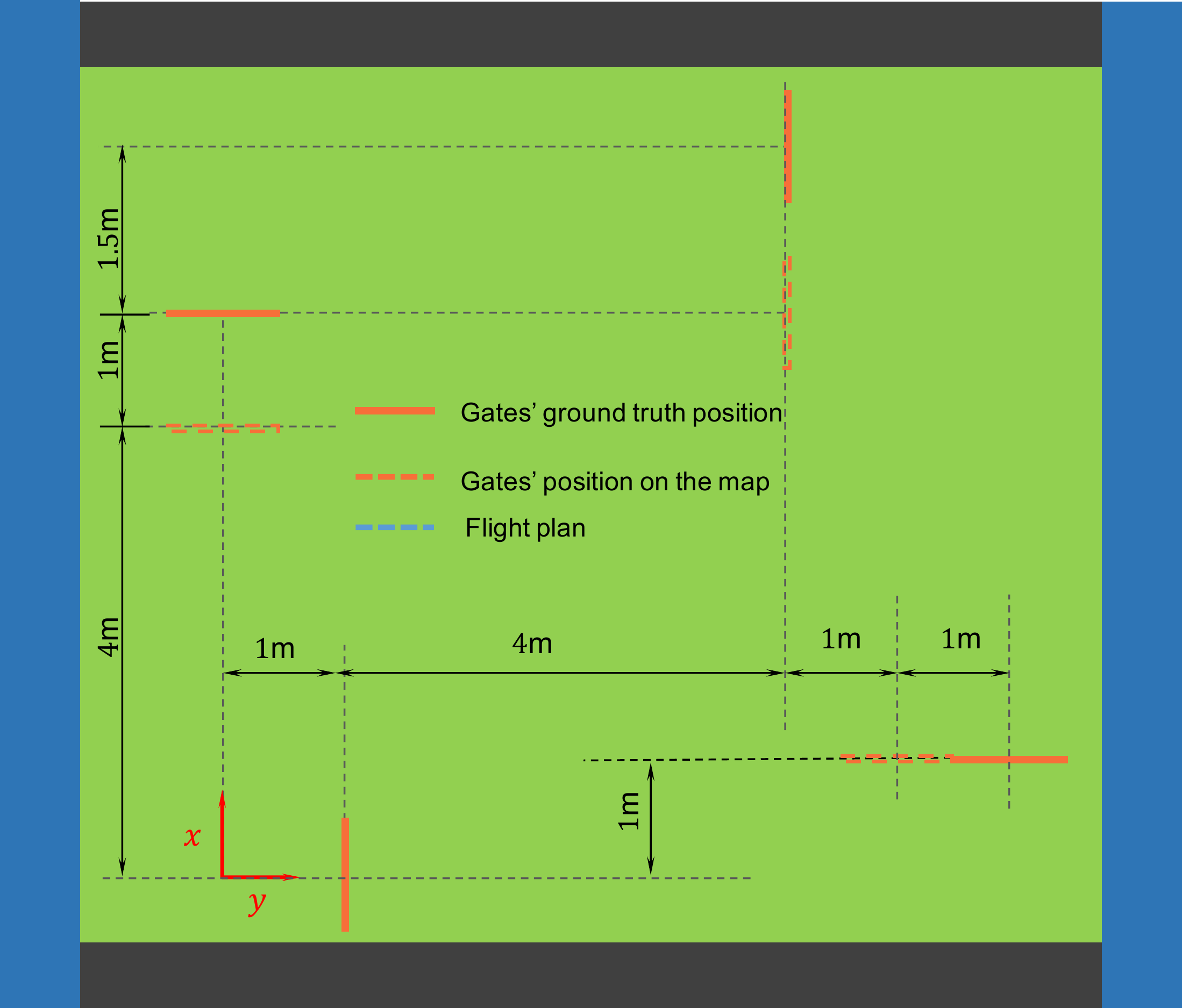}} 
    \subfigure[The flying data of the first lap.]{\includegraphics[scale=0.5,trim={3cm 9cm 3cm 8cm},clip]{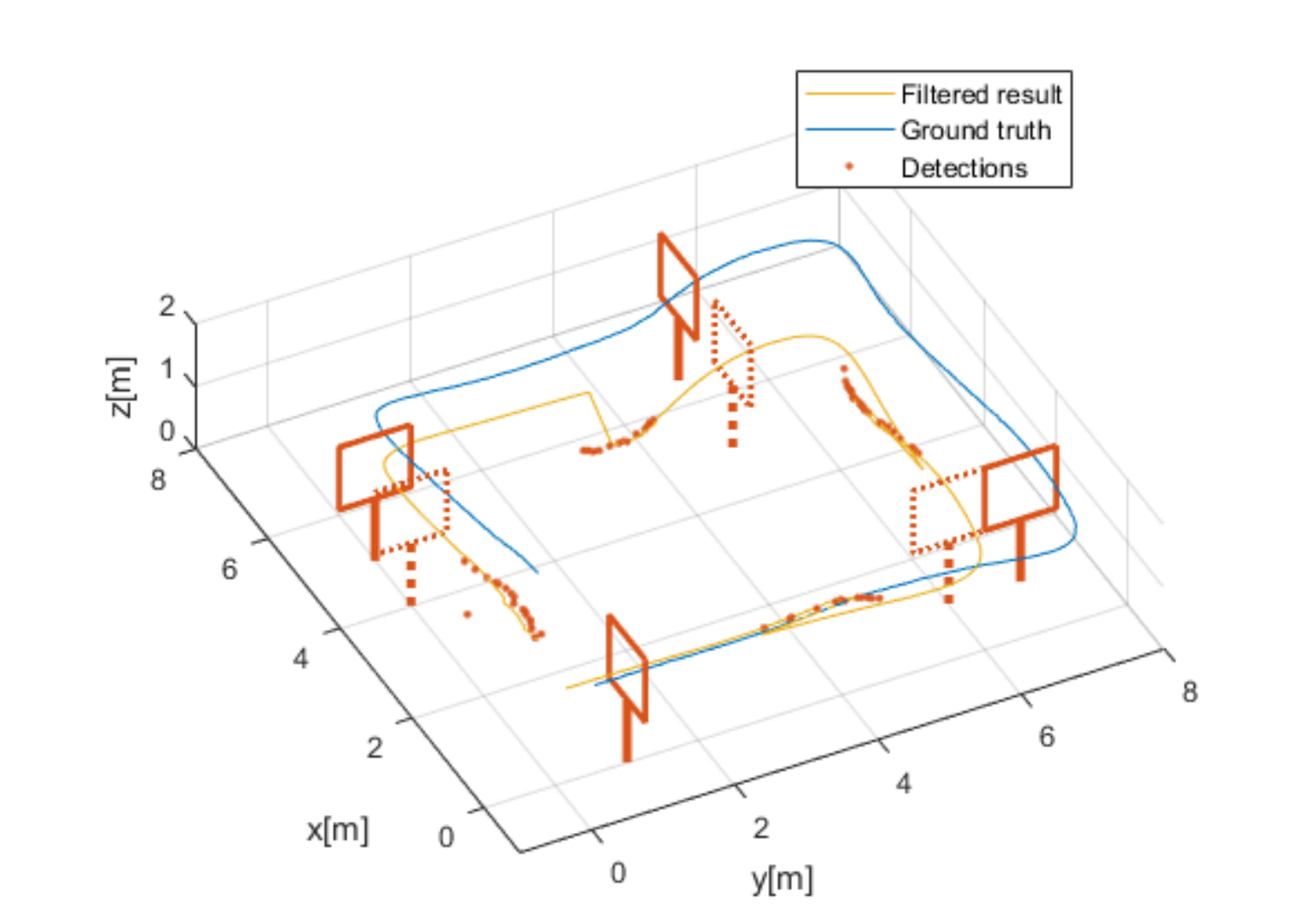}}
   
\caption{The experiment where the gates are displaced. When the drone sees the next gate after passing through one gate, the filter will start correcting the error caused by the prediction drift and the gate's displacement. Thus, there is a jump in the filtering result.}
\label{fig:track_map}
\end{figure}

In this section, we test our strategy under a difficult condition where the drone flies faster, the gates are displaced and the detection frequency is low. The real gate positions and their position on the map are listed in Table \ref{tab:race track} and shown in Figure \ref{fig:track_map}(a). Gates are displaced between 0 and 1.5m from their supposed positions. The dashed orange lines in Figure \ref{fig:track_map}(a) denote the gate positions on the map while the solid orange lines denote the real gate positions which are displaced from the map. Figure \ref{fig:track_map}(b) shows the flight data of the first lap. The orange solid gates are the ground truth positions of the gates. The yellow curve is the filtered position based on the gates' positions on the map (orange dashed gates). In other words, the yellow curve is where the drone thinks it is based on the knowledge of the map. After passing through one gate, when the drone detects the next gate, the filter will start correcting the filtering error from the prediction error and the gate displacement.
 
\begin{figure} [!htb]
    \centering
\subfigure[The position estimation result. It should be noted that the position estimation curve does not coincide with the ground truth curve coming from our motion capture system because the gate displacements.]{\includegraphics[width=0.45\columnwidth,trim={2cm 8cm 2cm 8cm},clip]{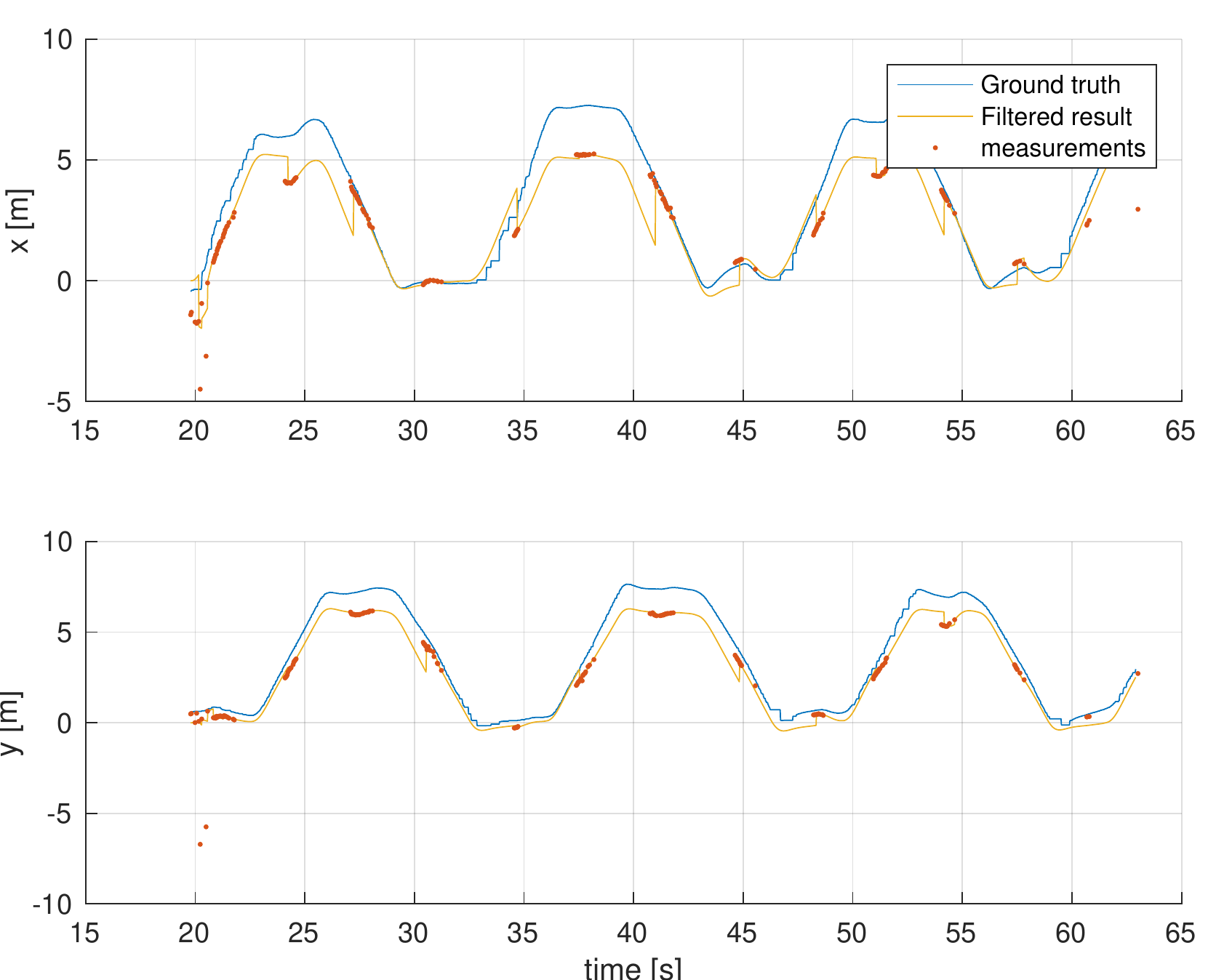}} 
\subfigure[The velocity estimation result of VML]{\includegraphics[width=0.45\columnwidth,trim={2cm 8cm 2cm 8cm}]{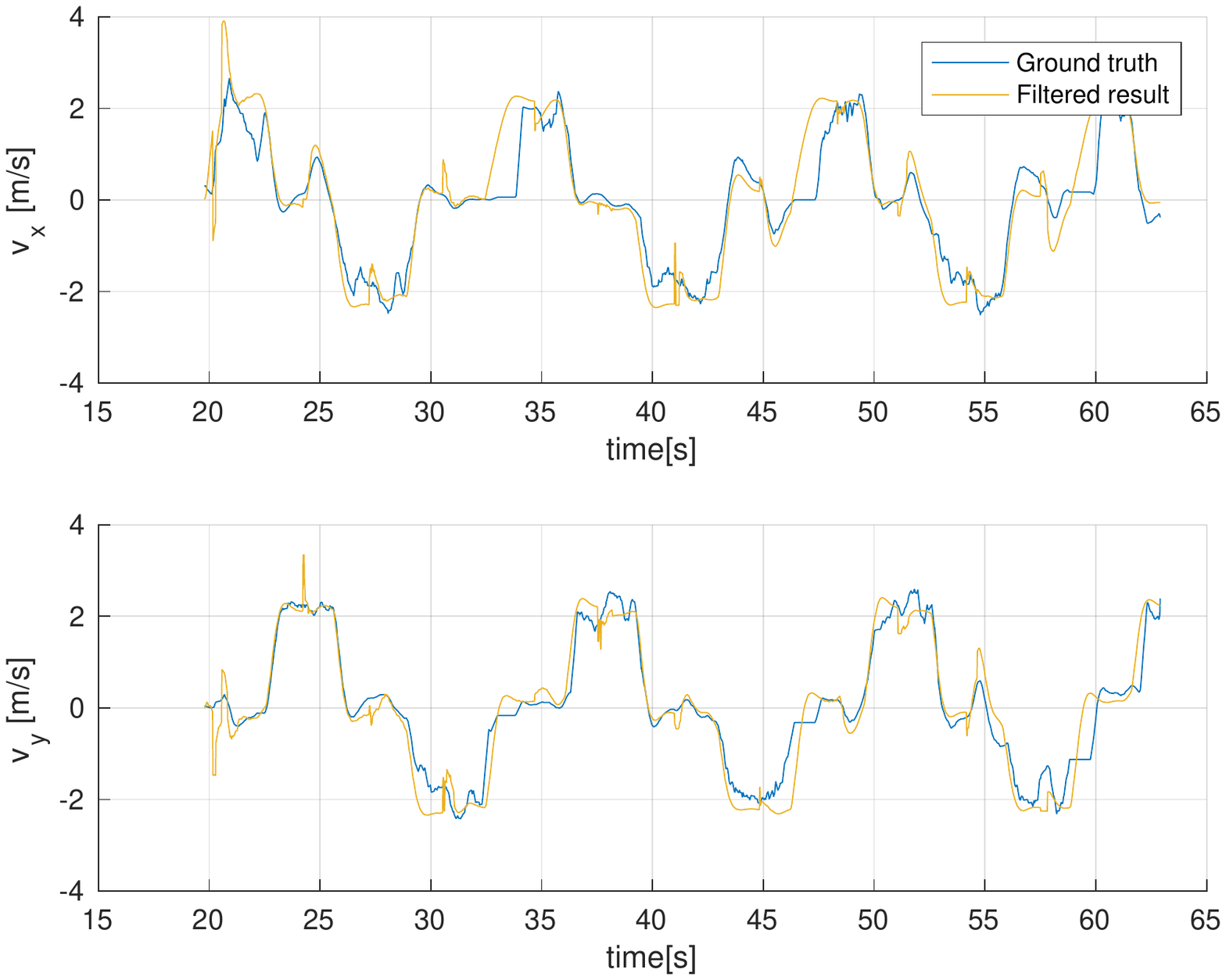}} %
    \caption{The result of flying the track with the gate displacement.}%
    \label{fig:flying log}
\end{figure}

 The whole flight result is shown in Figure \ref{fig:flying log}. From the result, it can be seen that the drone can fly the track for $3$ laps with an average speed of $2m/s$ and a maximum speed of $2.6m/s$ while an experienced pilot flies the same drone in the same track with an average speed of $2.7m/s$ after several runs of training. Figure \ref{fig:flying log}(a) is the filtering result of the position. It should be noted that the filtering result does not coincide with the ground truth curve because of the displacement of the gates. The pose estimation is based on the gates' position on the map. When the gates are displaced, the drone still thinks they are at the position which the map indicates. After the turn, when the drone sees the next gate, which is displaced, it will attribute the misalignment to the prediction error and correct the prediction by means of new detections. With this strategy, our algorithm is robust to the displacement of the gates.
 
 \begin{table}[!htb]
\caption{The position of the gates with displacement}
\centering
\begin{tabular}{|c|c|c|c|c|}
\hline
\centering
gate ID & $x_g[m]$ & $y_g[m]$ & $\Tilde{x}_g[m]$ &  $\Tilde{y}_g[m]$ \\ \hline \hline
1 & 5 & 0 & 4 & 0    \\ \hline
2 & 6.5 & 5 & 5 & 5    \\ \hline
3 & 1 & 7 & 1 & 6    \\ \hline
4 & 0 & 1 & 0 & 1    \\ \hline
\end{tabular}
\label{tab:race track}
\end{table}

\subsection{Flying experiment with different altitude and moving gate}
\begin{figure} [!hbt]
    \centering
\subfigure[After the drone passes through the second gate, the gate is moved.]{\includegraphics[scale = 0.08,trim={2cm 0cm 2cm 0cm},clip]{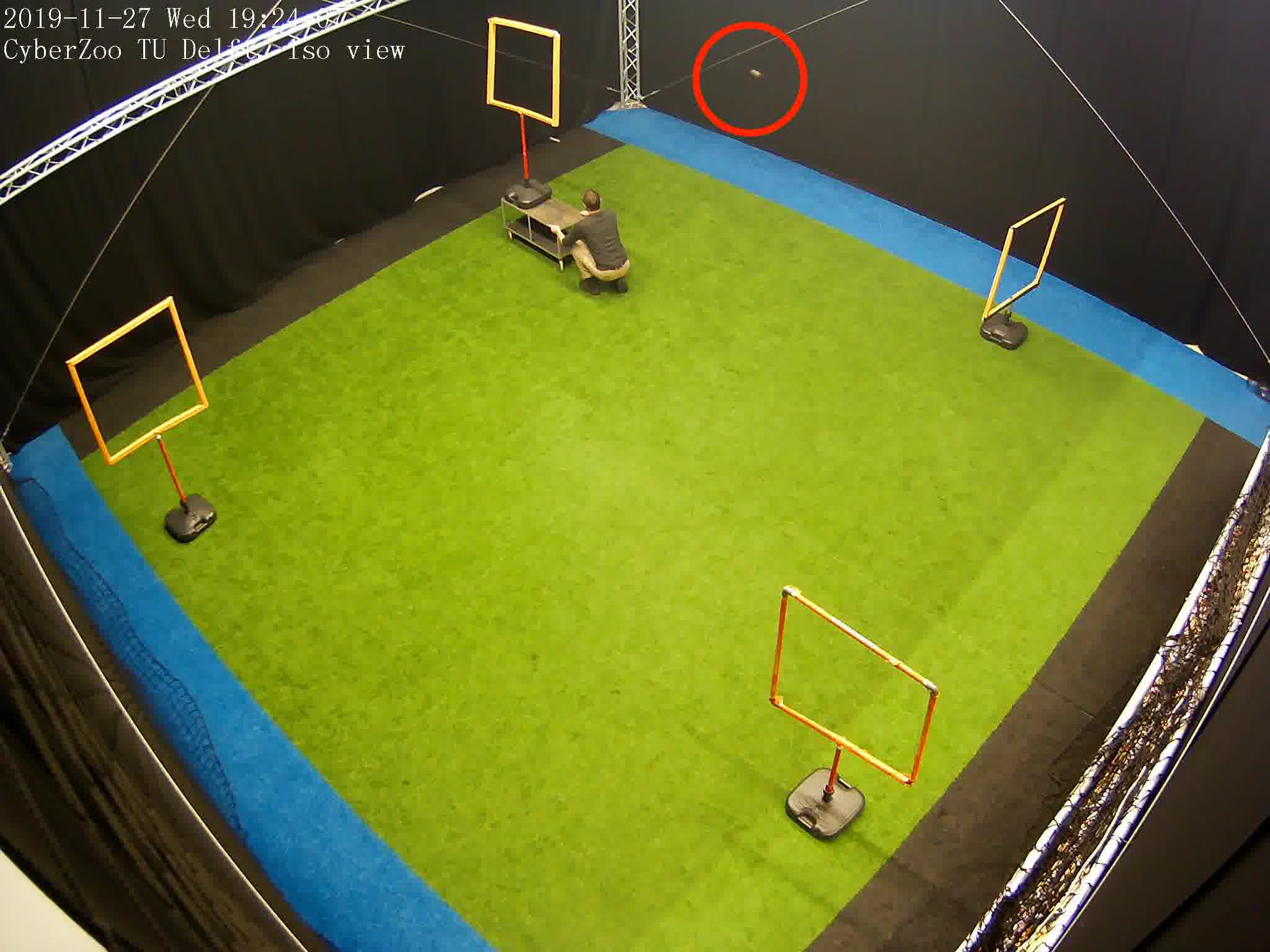}} \hspace{1cm}
\subfigure[In the next lap, the drone can adapt to the changing position of the gate and fly through it.]{\includegraphics[scale = 0.08,trim={2cm 0cm 2cm 0cm}]{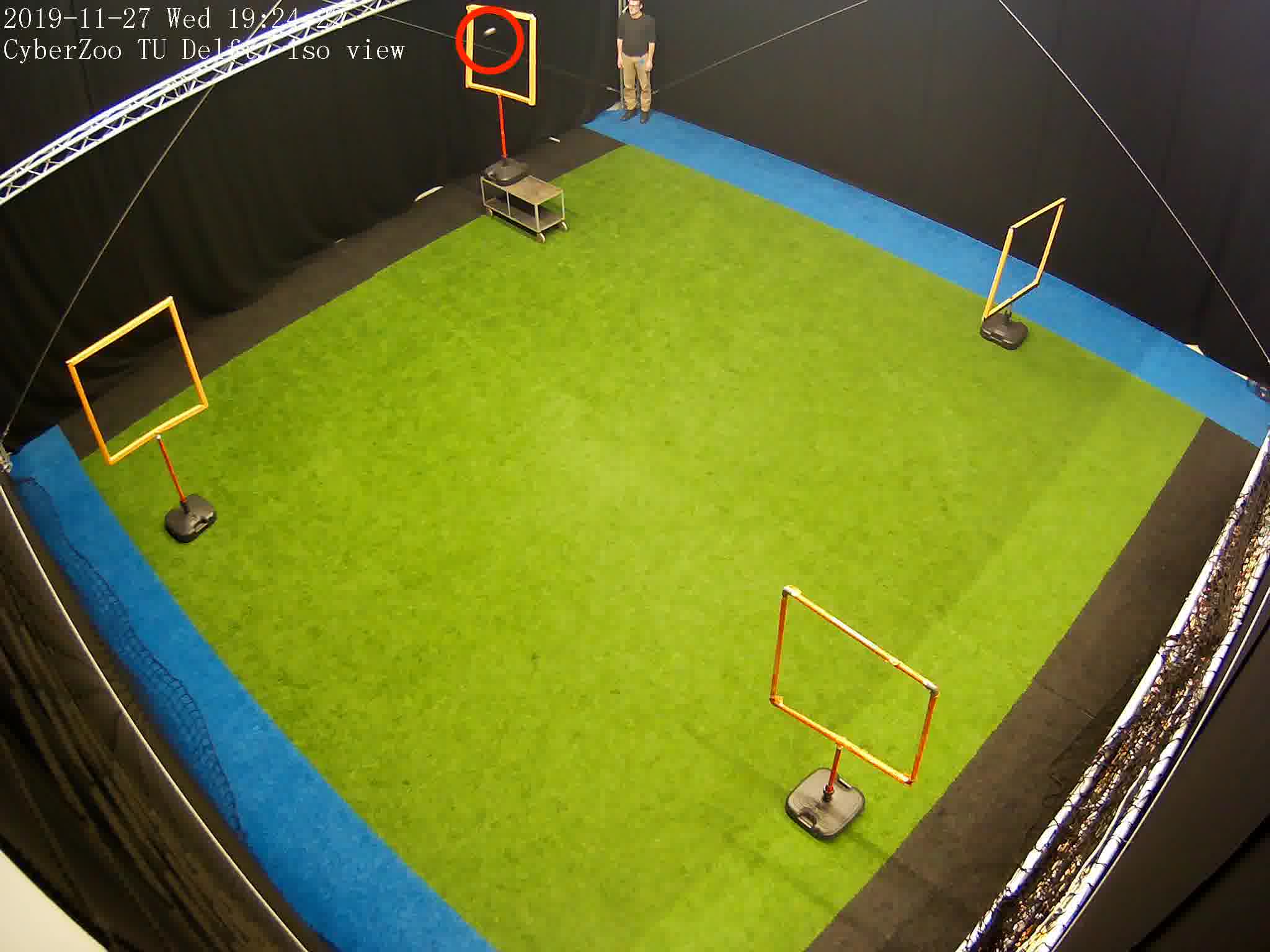}} %
    \caption{The flying experiment where the heights of the gates vary from $0.5m$ to $2.5m$. During the flight, the position of the second gate is changed.}%
    \label{fig:photos_moving_gate}
\end{figure}

We also show a more challenging trace track where the height of the gates varies from $0.5m$ to $2.5m$. Also, during the flight, the position of the second gate ($2.5m$) is changed after the drone passes through it. In the next lap, the drone can adapt to the changing position of the gate. (Figure \ref{fig:photos_moving_gate})

\begin{figure} [!hbt]
    \centering
    \includegraphics[width=0.7\columnwidth,trim={0cm 7cm 0cm 8cm},clip]{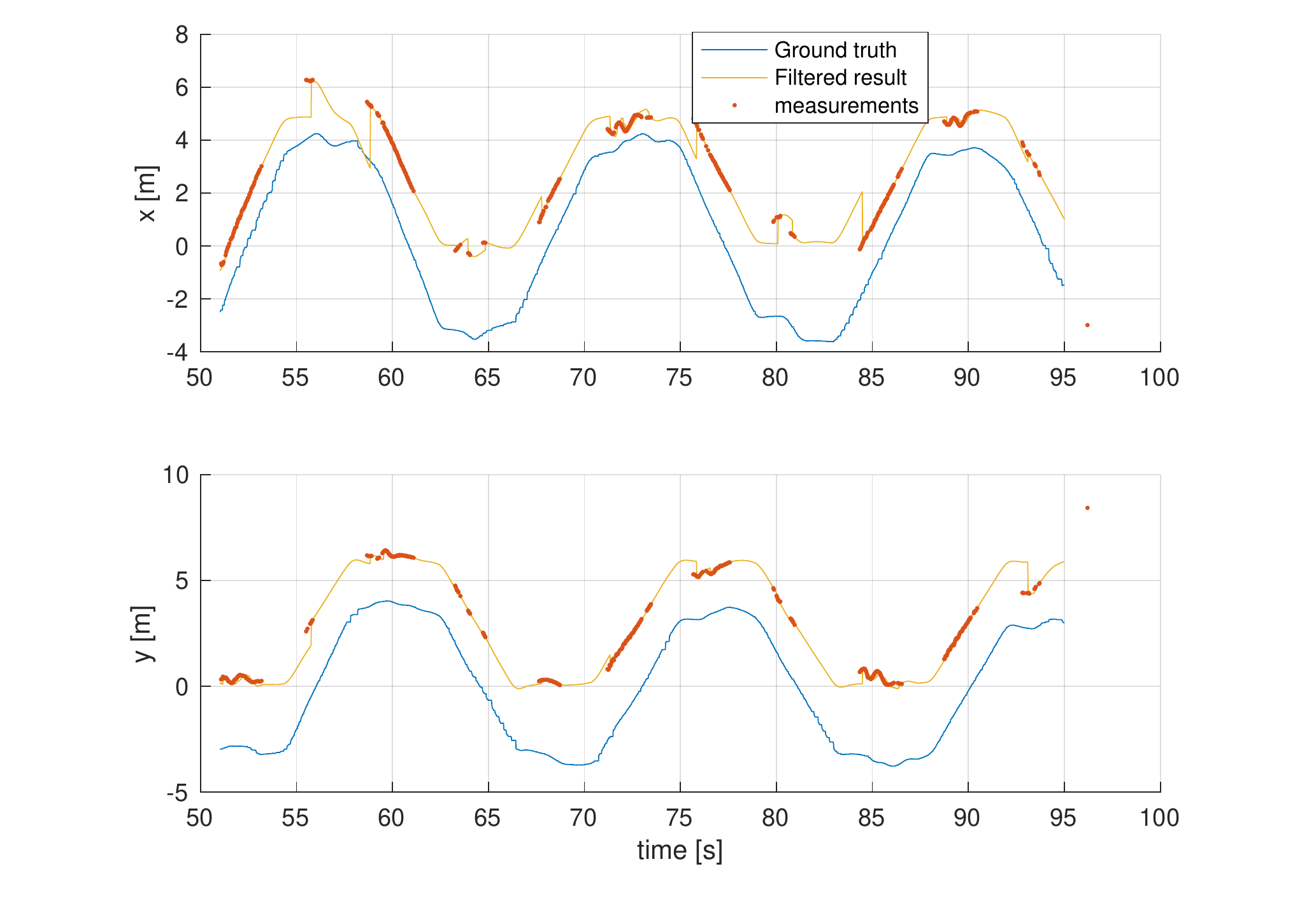}
\caption{The flying result of the drone flying the track with different height and the gate's position changing.}
\label{fig:log_chaning_height}
\end{figure}

The flight result is shown in Figure \ref{fig:log_chaning_height}. In this flight, the waypoints are not changed and the gates are deployed without any ground truth measurement. Thus, the estimated position does not coincide with the ground-truth position. It should be noted that the height difference between the second gate and the third gate is $2m$. With this altitude change which violates the constant altitude assumption for the prediction error model, the proposed VML is still accurate enough to navigate the drone through the gate.  

From the real flight result, we can see that the VML performs well and can navigate the drone through the racing track with high speed even though the gates are displaced. Also, this strategy does not need computationally expensive methods like generic VIO and SLAM. This allows it to be run on a very light-weight flying platform.

\section{Discussion}
\label{sec:Discussion}
In this paper, we proposed a novel state estimation method called Visual Model-predictive Localization which provides navigation information for a 72 gram autonomous racing drone. The algorithm's properties were thoroughly studied in simulation and the feasibility of real-world implementation was shown in challenging real world experiments. Although in this paper VML is used for a specific drone race scenario, this method can be directly used for navigation in other more general scenarios where the sensors have low frequency, temporary failure, outliers and delays. For example, our approach can be directly adopted into an outdoor environment where position measurements are provided by a GPS signal that has a delay, temporary failures and outliers. Just as in our drone race experiments, the proposed approach should be more reliable than a Kalman filter. For indoor flight, we used a common linear drag model for state prediction which does not need a lot of effort and precise equipment to identify. Outdoor flight would require adaptations to this model, for instance such as the ones explained in, e.g., \cite{sikkel2016novel}.

We implemented our approach by adding a cheap smart camera Jevois to a tiny racing drone Trashcan. With very limited carrying capacity and more complex aerodynamics property, it is still demonstrated that this light-weight flying platform has the ability to finish the drone race task autonomously. Compared to a regular size racing drone, the Trashcan has more complex aerodynamics and is more sensitive to disturbances. On the other hand, it has faster dynamics which can make maneuvers more agile. More importantly, it is much safer than a regular size racing drone, which may even allow for flying at home. In any case, the present approach represents another direction of the autonomous drone race, which does not need high performance and heavy onboard computers. Also, without computationally expensive navigation methods such as SLAM and VIO, the proposed approach is still able to make the drone navigate autonomously with relatively high speed.

However, the proposed approach still has its limitations. First of all, in this approach, we don't estimate the thrust. Instead, we use a non-changing altitude assumption to approximate the thrust to derive the prediction error model. The simulation and real world experiments have shown that violating this assumption can still have accurate estimation. Still, when the racing track will contain more considerable height changes, it will become desirable  to estimate the thrust with a model, in order to have a more accurate error model and increase the estimation accuracy, especially in more aggressive flight.

Secondly, the current detection method is sensitive to light conditions. Most failures are caused by the non-detection of the gate. This is a major bottleneck of increasing the speed of the flight. In the future, we will design a gate detection method using deep learning methods to detect the gate in a more complex environment. This deep net can then run on the GPU of the Jevois. Also, higher speeds could be attainable. 

Thirdly, in this paper, we mainly focus on the navigation part of the drone. The guidance is only a way-point based method and the controller is a PID controller. To make the drone fly faster, optimal guidance and control methods are needed. Another direction is to explore joint estimation for navigation. This will become very useful when one assumes that gates are mostly not displaced. Then, over multiple laps, the drone can get a better idea of where the gates are. 

In the future, with the high speed development of computational capacity, when the more reliable gate detection and online optimal control are implemented onboard, the speed of this autonomous racing drone should certainly be increased significantly. Compared to regularly sized drones, this tiny flying platform should perform faster and more agile flight. At that time, the proposed VML approach will still be suitable for providing stable state estimation for the drone.

\section{Conclusion}
\label{lab:conclusion}
In this paper, we presented an efficient Visual Model-predictive Localization (VML) approach to autonomous drone racing. The approach employs a velocity-stable model that predicts lateral accelerations based on attitude estimates from the AHRS. Vision is used for detecting gates in the image, and - by means of their supposed location in the map - for localizing the drone in the coarse global map. Simulation and real-world flight experiments show that VML can provide robust estimates with sparse visual measurements and large outliers. This robust and computationally very efficient approach was tested on an extremely lightweight flying platform, i.e., a Trashcan racing drone with a Jevois camera. In the flight experiments, the Trashcan flew a track of $3$ laps with an average speed of $2m/s$ and a maximum speed of $2.6m/s$. To the best of our knowledge, it is the world's smallest autonomous racing drone with a weight $6$ times lighter than the currently lightest autonomous racing drone setup, while its velocity is on a par with the currently fastest autonomously flying racing drones seen at the latest IROS autonomous drone race.

\bibliographystyle{apalike}
\bibliography{jfrExampleRefs}

\begin{thebibliography}{}

\bibitem[Chang, 2014]{chang2014robust}
Chang, G. (2014).
\newblock Robust kalman filtering based on mahalanobis distance as outlier
  judging criterion.
\newblock {\em Journal of Geodesy}, 88(4):391--401.

\bibitem[Diderrich, 1985]{diderrich1985kalman}
Diderrich, G.~T. (1985).
\newblock The kalman filter from the perspective of goldberger—theil
  estimators.
\newblock {\em The American Statistician}, 39(3):193--198.

\bibitem[Faessler et~al., 2017]{faessler2017differential}
Faessler, M., Franchi, A., and Scaramuzza, D. (2017).
\newblock Differential flatness of quadrotor dynamics subject to rotor drag for
  accurate tracking of high-speed trajectories.
\newblock {\em IEEE Robotics and Automation Letters}, 3(2):620--626.

\bibitem[Falanga et~al., 2017]{falanga2017aggressive}
Falanga, D., Mueggler, E., Faessler, M., and Scaramuzza, D. (2017).
\newblock Aggressive quadrotor flight through narrow gaps with onboard sensing
  and computing using active vision.
\newblock In {\em 2017 IEEE International Conference on Robotics and Automation
  (ICRA)}, pages 5774--5781. IEEE.

\bibitem[Fischler and Bolles, 1981]{fischler1981random}
Fischler, M.~A. and Bolles, R.~C. (1981).
\newblock Random sample consensus: a paradigm for model fitting with
  applications to image analysis and automated cartography.
\newblock {\em Communications of the ACM}, 24(6):381--395.

\bibitem[Gao et~al., 2019]{gao2019optimal}
Gao, F., Wang, L., Wang, K., Wu, W., Zhou, B., Han, L., and Shen, S. (2019).
\newblock Optimal trajectory generation for quadrotor teach-and-repeat.
\newblock {\em IEEE Robotics and Automation Letters}.

\bibitem[Gati, 2013]{gati2013open}
Gati, B. (2013).
\newblock Open source autopilot for academic research-the paparazzi system.
\newblock In {\em 2013 American Control Conference}, pages 1478--1481. IEEE.

\bibitem[Gross et~al., 2012]{gross2012flight}
Gross, J.~N., Gu, Y., Rhudy, M.~B., Gururajan, S., and Napolitano, M.~R.
  (2012).
\newblock Flight-test evaluation of sensor fusion algorithms for attitude
  estimation.
\newblock {\em IEEE Transactions on Aerospace and Electronic Systems},
  48(3):2128--2139.

\bibitem[Hattenberger et~al., 2014]{hattenberger2014using}
Hattenberger, G., Bronz, M., and Gorraz, M. (2014).
\newblock Using the paparazzi uav system for scientific research.
\newblock In {\em IMAV 2014, International Micro Air Vehicle Conference and
  Competition 2014}, pages pp--247.

\bibitem[Jung et~al., 2018]{jung2018direct}
Jung, S., Cho, S., Lee, D., Lee, H., and Shim, D.~H. (2018).
\newblock A direct visual servoing-based framework for the 2016 iros autonomous
  drone racing challenge.
\newblock {\em Journal of Field Robotics}, 35(1):146--166.

\bibitem[Kaufmann et~al., 2018a]{kaufmann2018beauty}
Kaufmann, E., Gehrig, M., Foehn, P., Ranftl, R., Dosovitskiy, A., Koltun, V.,
  and Scaramuzza, D. (2018a).
\newblock Beauty and the beast: Optimal methods meet learning for drone racing.
\newblock {\em arXiv preprint arXiv:1810.06224}.

\bibitem[Kaufmann et~al., 2018b]{kaufmann2018deep}
Kaufmann, E., Loquercio, A., Ranftl, R., Dosovitskiy, A., Koltun, V., and
  Scaramuzza, D. (2018b).
\newblock Deep drone racing: Learning agile flight in dynamic environments.
\newblock {\em arXiv preprint arXiv:1806.08548}.

\bibitem[Li et~al., 2018]{li2018autonomous}
Li, S., Ozo, M., De~Wagter, C., and de~Croon, G. (2018).
\newblock Autonomous drone race: A computationally efficient vision-based
  navigation and control strategy.
\newblock {\em arXiv preprint arXiv:1809.05958}.

\bibitem[Li et~al., 2016]{li2016gps}
Li, Z., Chang, G., Gao, J., Wang, J., and Hernandez, A. (2016).
\newblock Gps/uwb/mems-imu tightly coupled navigation with improved robust
  kalman filter.
\newblock {\em Advances in Space Research}, 58(11):2424--2434.

\bibitem[Loianno et~al., 2017]{loianno2017estimation}
Loianno, G., Brunner, C., McGrath, G., and Kumar, V. (2017).
\newblock Estimation, control, and planning for aggressive flight with a small
  quadrotor with a single camera and imu.
\newblock {\em IEEE Robotics and Automation Letters}, 2(2):404--411.

\bibitem[Lupashin et~al., 2014]{lupashin2014platform}
Lupashin, S., Hehn, M., Mueller, M.~W., Schoellig, A.~P., Sherback, M., and
  D’Andrea, R. (2014).
\newblock A platform for aerial robotics research and demonstration: The flying
  machine arena.
\newblock {\em Mechatronics}, 24(1):41--54.

\bibitem[McGuire et~al., 2017]{mcguire2017efficient}
McGuire, K., De~Croon, G., De~Wagter, C., Tuyls, K., and Kappen, H. (2017).
\newblock Efficient optical flow and stereo vision for velocity estimation and
  obstacle avoidance on an autonomous pocket drone.
\newblock {\em IEEE Robotics and Automation Letters}, 2(2):1070--1076.

\bibitem[Mellinger and Kumar, 2011]{mellinger2011minimum}
Mellinger, D. and Kumar, V. (2011).
\newblock Minimum snap trajectory generation and control for quadrotors.
\newblock In {\em 2011 IEEE International Conference on Robotics and
  Automation}, pages 2520--2525. IEEE.

\bibitem[Mellinger et~al., 2012]{mellinger2012trajectory}
Mellinger, D., Michael, N., and Kumar, V. (2012).
\newblock Trajectory generation and control for precise aggressive maneuvers
  with quadrotors.
\newblock {\em The International Journal of Robotics Research}, 31(5):664--674.

\bibitem[Moon et~al., 2019]{moon2019challenges}
Moon, H., Martinez-Carranza, J., Cieslewski, T., Faessler, M., Falanga, D.,
  Simovic, A., Scaramuzza, D., Li, S., Ozo, M., De~Wagter, C., et~al. (2019).
\newblock Challenges and implemented technologies used in autonomous drone
  racing.
\newblock {\em Intelligent Service Robotics}, pages 1--12.

\bibitem[Moon et~al., 2017]{moon2017iros}
Moon, H., Sun, Y., Baltes, J., and Kim, S.~J. (2017).
\newblock The iros 2016 competitions [competitions].
\newblock {\em IEEE Robotics \& Automation Magazine}, 24(1):20--29.

\bibitem[Morrell et~al., 2018]{morrell2018differential}
Morrell, B., Rigter, M., Merewether, G., Reid, R., Thakker, R., Tzanetos, T.,
  Rajur, V., and Chamitoff, G. (2018).
\newblock Differential flatness transformations for aggressive quadrotor
  flight.
\newblock In {\em 2018 IEEE International Conference on Robotics and Automation
  (ICRA)}, pages 1--7. IEEE.

\bibitem[Mueller et~al., 2015]{mueller2015fusing}
Mueller, M.~W., Hamer, M., and D'Andrea, R. (2015).
\newblock Fusing ultra-wideband range measurements with accelerometers and rate
  gyroscopes for quadrocopter state estimation.
\newblock In {\em 2015 IEEE International Conference on Robotics and Automation
  (ICRA)}, pages 1730--1736. IEEE.

\bibitem[Sanket et~al., 2018]{sanket2018gapflyt}
Sanket, N.~J., Singh, C.~D., Ganguly, K., Ferm{\"u}ller, C., and Aloimonos, Y.
  (2018).
\newblock Gapflyt: Active vision based minimalist structure-less gap detection
  for quadrotor flight.
\newblock {\em IEEE Robotics and Automation Letters}, 3(4):2799--2806.

\bibitem[Santamaria-Navarro et~al., 2018]{santamaria2018autonomous}
Santamaria-Navarro, A., Loianno, G., Sol{\`a}, J., Kumar, V., and
  Andrade-Cetto, J. (2018).
\newblock Autonomous navigation of micro aerial vehicles using high-rate and
  low-cost sensors.
\newblock {\em Autonomous robots}, pages 1--18.

\bibitem[Santana et~al., 2015]{santana2015outdoor}
Santana, L.~V., Brandao, A.~S., and Sarcinelli-Filho, M. (2015).
\newblock Outdoor waypoint navigation with the ar. drone quadrotor.
\newblock In {\em 2015 International Conference on Unmanned Aircraft Systems
  (ICUAS)}, pages 303--311. IEEE.

\bibitem[Sikkel et~al., 2016]{sikkel2016novel}
Sikkel, L., de~Croon, G., De~Wagter, C., and Chu, Q. (2016).
\newblock A novel online model-based wind estimation approach for quadrotor
  micro air vehicles using low cost mems imus.
\newblock In {\em 2016 IEEE/RSJ International Conference on Intelligent Robots
  and Systems (IROS)}, pages 2141--2146. IEEE.

\bibitem[van Horssen et~al., 2019]{van2019event}
van Horssen, E., van Hooijdonk, J., Antunes, D., and Heemels, W. (2019).
\newblock Event-and deadline-driven control of a self-localizing robot with
  vision-induced delays.
\newblock {\em IEEE Transactions on Industrial Electronics}.

\bibitem[Weiss et~al., 2012]{weiss2012versatile}
Weiss, S., Achtelik, M.~W., Chli, M., and Siegwart, R. (2012).
\newblock Versatile distributed pose estimation and sensor self-calibration for
  an autonomous mav.
\newblock In {\em 2012 IEEE International Conference on Robotics and
  Automation}, pages 31--38. IEEE.

\end{thebibliography}

\section*{Appendex}

\subsection*{Kalman filter's prediction model}
\begin{align}
\begin{cases}
     \begin{bmatrix} \dot{x} \\ \dot{y} \end{bmatrix} &= \begin{bmatrix} v_x \\ v_y \end{bmatrix} \\
    \begin{bmatrix} \dot{v}_x \\ \dot{v}_y \end{bmatrix} &= \begin{bmatrix} \cos{\psi^m} & -\sin{\psi^m}  \\\sin{\psi^m} & \cos{\psi^m} \end{bmatrix}\{
    \begin{bmatrix} -g\sin{\theta} \\ g\cos\phi \end{bmatrix} +
    \begin{bmatrix} k_x & 0 \\ 0 & k_y \end{bmatrix} \begin{bmatrix} \cos{\psi^m} & \sin{\psi^m}  \\-\sin{\psi^m} & \cos{\psi^m} \end{bmatrix}\begin{bmatrix} v_x \\ v_y \end{bmatrix}\} \\
    \begin{bmatrix} \dot{B}_N \\ \dot{B}_E \end{bmatrix} &= \begin{bmatrix} 0 \\ 0 \end{bmatrix} \\
    \begin{bmatrix} \phi \\ \theta \end{bmatrix} &= \begin{bmatrix} \phi^{m} \\ \theta^{m} \end{bmatrix} + \begin{bmatrix} \cos{\psi^m} & \sin{\psi^m} \\ -\sin{\psi^m} & \cos{\psi^m} \end{bmatrix}\begin{bmatrix} B_N \\ B_E \end{bmatrix}
\end{cases}
\label{equ:kalman filter prediction}
\end{align}

The inputs of the system \ref{equ:kalman filter prediction} is the AHRS reading $\mathbf{u}= [\phi^m,\theta^m,\psi^m]^{\rm T}$. The states of the Extended Kalman filter are $\mathbf{X}= [x,y,v_x,v_y,B_N,B_E]^{\rm T}$. With the standard Extended Kalman filter procedure list below, the states of the system can be estimated.

\subsection*{Pseudocodes}
\begin{algorithm}[hbt!]
    \caption{$gate\_assignment$}
    \label{alg:mhe_run}
    \KwIn{$\Delta \bar{x}_k,\Delta \bar{y}_k$}
    \KwOut{$\bar{x}_k,\bar{y}_k$}
    \For{$i=1;i<=gate\_numbers;i++$}
    {
      $\bar{x}^i_k = \cos\psi^i_g\Delta \bar{x}_k +\sin\psi^i_g\Delta \bar{y}_k + x_g^i$\\
      $\bar{y}^i_k = -\sin\psi^i_g\Delta \bar{x}_k +\cos\psi^i_g\Delta \bar{y}_k + y_g^i$\\
      $\Delta d_k^i = (\bar{x}^i_k - \hat{x}_{k})^2 +(\bar{y}^i_k - \hat{y}_{k})^2$ \\
    }
    $j = \operatorname*{argmin}_i\Delta d_k^i$ \\
    $\bar{x}_k = \bar{x}^j_k$ \\
    $\bar{y}_k = \bar{y}^j_k$
\end{algorithm}

\begin{algorithm}[hbt!]
    \caption{Basic RANSAC Fitting}
    \label{alg:BRF}
    \KwIn{$\Delta \mathbf{x}^p_{k-q,k}$, $\Delta \mathbf{t}$}
    \KwOut{$\mathbf{\hat{\beta}} = \begin{bmatrix} \Delta x^p_{k-q} \Delta v^p_{k-q} \end{bmatrix}$}
    \For{$i=1;i<=iterations;i++$}
    {
      $\mathbf{sample\_id} = random\_integers(k-q, k, n^s$)  \\
      $\Delta \mathbf{t}^s = \Delta \mathbf{t}[\mathbf{sample\_id}]$ \\
      $\Delta \mathbf{x}^s_{k-q,k} =\Delta \mathbf{x}^p_{k-q,k}[\mathbf{sample\_id}]$ \\
      $[\Delta {x^p_{k-q}}_i, \Delta {v^p_{k-q}}_i] = linear\_regression(\Delta \mathbf{t}^s,\Delta \mathbf{x}^s_{k-q,k})$ \\
      \For{$j=k-q;j<k;j++$}
      {
         $\epsilon_j = \norm{\Delta {v^p_{k-q}}_i(\Delta t_j - \Delta t_{k-q})+\Delta {x^p_{k-q}}_i - \Delta x^p_j}_2$ \\
         \If{$\epsilon_j > \sigma_{th}$}{$\varepsilon_i +=\sigma_{th}$}
         \Else {$\varepsilon_i += \epsilon_j$}
      }
      \If{$\varepsilon_i < \varepsilon_{min}$}
      {
      $\varepsilon_{min} = \varepsilon$ \\
      $\Delta x^p_{k-q} = \Delta {x^p_{k-q}}_i$ \\
      $\Delta v^p_{k-q} = \Delta {v^p_{k-q}}_i$ \\
      }
    }
\label{alg:BRF}
\end{algorithm}

\begin{algorithm}
    \caption{Visual Model-predictive Localization}
    \label{alg:mhe_run}
    \While{true}
    {
      $t_{k+i} = current\_time$ \\
      $x^p_k += {v_x^p}_k{T_s}$ \\
      $y^p_k += {v_y^p}_k{T_s}$ \\
      ${v^p_x}_k += (-g\tan{\theta}-c{v^p_x}_k)T_s$\\
      ${v^p_y}_k += (g\tan{\phi}-c{v^p_y}_k)T_s$\\
      $clear\_old\_elements\_in\_queue()$ \\
      \If{flagNewPoseEstimation}
      {
          $queue.front++$ \\
          $queue.\mathbf{time}[queue.front] = t_{k+i}$ \\
          $queue.\mathbf{x}^p_{k}[queue.front] = x^p_{k}$ \\
          $queue.\mathbf{y}^p_{k}[queue.front] = y^p_{k}$ \\
          $queue.\bar{\mathbf{x}}[queue.front] = \bar{x}_k$ \\
          $queue.\bar{\mathbf{y}}[queue.front] = \bar{y}_k$ \\
          $queue.size++$ \\
          \If{$queue.size > N_{fit}$}
          {
             $[\Delta x^p_{k-q},\Delta {v_x}^p_{k-q},\Delta y^p_{k-q},\Delta {v_y}^p_{k-q}] = filter\_correct()$\\
             $t_{k} = t_{k+i}$\\
          }
      }
      $\hat{x}_{k} = x^p_k + \Delta x^p_{k-q} + (t_{k+i} - t_{k})\Delta {v_x}^p_{k-q}$ \\
      $\hat{y}_{k} = y^p_k + \Delta y^p_{k-q} + (t_{k+i} - t_{k})\Delta {v_y}^p_{k-q}$ \\
       ${\hat{v}_x}{}_{k} = {v^p_x}_k + \Delta {v_x}^p_{k-q}$ \\
       ${\hat{v}_y}{}_{k} = {v^p_y}_k + \Delta {v_y}^p_{k-q}$ \\
    }
\end{algorithm}

\begin{algorithm}[hbt!]
    \caption{filter\_correct}
    \label{alg:mhe_run}
    \KwOut{$\Delta x^p_{k-q},\Delta {v_x}^p_{k-q},\Delta y^p_{k-q},\Delta {v_y}^p_{k-q}$}
    \For{$i=1;i<=queue.size;i++$}
    {
    $\Delta \mathbf{t}_i = queue.\mathbf{time}.[newest\_item\_id] - queue.\mathbf{time}[i]$ \\
    $\Delta \mathbf{x}^p_i = queue.\bar{\mathbf{x}}[i] - queue.{\mathbf{x}}^p[i]$ \\
    $\Delta \mathbf{y}^p_i = queue.\bar{\mathbf{y}}[i] - queue.\hat{\mathbf{y}}[i]$ \\
    }
    $[\Delta x^p_{k-q},\Delta {v_x}^p_{k-q}] = linear\_regression(\Delta \mathbf{t},\Delta \mathbf{x}^p)$ \\
    $[\Delta y^p_{k-q},\Delta {v_y}^p_{k-q}] = linear\_regression(\Delta \mathbf{t},\Delta \mathbf{y}^p)$
\end{algorithm}

\begin{algorithm}[hbt!]
    \caption{$flight\_plan$}
    \label{alg:mhe_run}
    \KwOut{$x^r,y^r,z^r,\psi^r$}
    \If{$(\hat{x}_{k} - \mathbf{wp}_x[waypoint\_id])^2+(\hat{y}_{k} - \mathbf{wp}_y[waypoint\_id])^2<D_{switch\_wp}$}
    {
    $waypoint\_id ++ $
    }
    \If{$(\hat{x}_{k} - \mathbf{wp}_x[waypoint\_id])^2+(\hat{y}_{k} - \mathbf{wp}_y[waypoint\_id])^2<D_{turn}$}
    {
     $\psi_{sp} = \mathbf{wp}_{\psi}[waypoint\_id+1])$ \\
     $r_{cmd} = k_r(\psi_{sp}-\psi^r)$ \\
     $\psi^r += r_{cmd}$
    }
    $x^r =   \mathbf{wp}_x[waypoint\_id])$ \\
    $y^r =   \mathbf{wp}_y[waypoint\_id])$ \\
    $z^r =   \mathbf{wp}_z[waypoint\_id])$ \\
    $\psi^r = \psi_{ref}$
    \label{alg:flight plan}
\end{algorithm}

\end{document}